\documentclass{article} 
\usepackage[final]{colm2025_xllmworkshop}

\usepackage{microtype}
\usepackage{hyperref}
\usepackage{url}
\usepackage{booktabs}

\usepackage{lineno}
\usepackage{graphicx}
\usepackage{float}
\usepackage{subcaption}

\definecolor{darkblue}{rgb}{0, 0, 0.5}
\hypersetup{colorlinks=true, citecolor=darkblue, linkcolor=darkblue, urlcolor=darkblue}

\title{Rethinking Human Preference Evaluation of LLM Rationales}

\author{
  Ziang Li$^{1*}$, 
  Manasi Ganti$^{1*}$, 
  Zixian Ma$^{1*}$, 
  Helena Vasconcelos$^{2}$, 
  Qijia He$^{1}$, 
  Ranjay Krishna$^{1}$ \\
  $^{1}$University of Washington \quad
  $^{2}$Stanford University \\
  \texttt{\{ziangl4, mganti, zixianma, ranjay\}@cs.washington.edu}, \\
  \texttt{heqj3@uw.edu}, \\
  \texttt{helenav@stanford.edu} \\
}

\begin{document}

\ifcolmsubmission
\linenumbers
\fi

\maketitle

\begin{abstract}
Large language models (LLMs) often generate natural language rationales—free-form explanations that help improve performance on complex reasoning tasks and enhance interpretability for human users. However, evaluating these rationales remains challenging. While recent work has relied on binary preference judgments from humans or LLM judges, such evaluations are often opaque and coarse-grained, offering limited insight into what makes one rationale better than another. In this work, we rethink preference evaluation for LLM-generated rationales by asking: (1) What attributes define good rationales? (2) Can human preferences be explained by these attributes? (3) Can attribute-based evaluation overcome the limitations of binary comparisons? We identify a set of key rationale attributes from prior literature and assess them using automatic metrics, LLM judgments, and human annotations. We then analyze two standard human preference datasets MT Bench and Chatbot Arena using SHAP to identify which attributes best explain human preference outcomes. Finally, we re-evaluate model-generated rationales using attribute-specific ELO scores, revealing more nuanced model comparisons and insights. Our findings suggest that fine-grained attribute evaluations can better characterize rationale quality and guide future research toward more interpretable and reliable evaluation practices.

\end{abstract}
\section{Introduction}
Recent advances in large language models (LLMs) have enabled them to solve complex tasks, including logical and mathematical questions that require multi-step reasoning~\citep{openai_learning_to_reason_2024,guo2025deepseek}. Prior work has shown that LLM-generated free-form textual rationales can improve model performance on such reasoning-intensive tasks~\citep{wei2022chain,yao2023tree}. For instance, chain-of-thought prompting has demonstrated enhanced LLM accuracy across various benchmarks~\citep{wei2022chain,openai_learning_to_reason_2024}. Beyond performance, rationales offer an interpretable window into the model’s reasoning process, helping human users better understand how conclusions are reached.

Despite their utility, free-form rationales remain challenging to evaluate. Several studies have attempted to characterize the desirable qualities of rationales, identifying attributes such as consistency, faithfulness, clarity, and length~\citep{golovneva2023roscoesuitemetricsscoring, chen2023revinformationtheoreticevaluationfreetext, joshi2023machinerationalesnotuseful,prasad2023recevalevaluatingreasoningchains, ramnath2024tailoringselfrationalizersmultirewarddistillation}. Recently, preference evaluations—in which human annotators or models compare two free-form responses and select the preferred one—have become the predominant approach for evaluating free-form responses, including rationales~\citep{zheng2023judgingllmasajudgemtbenchchatbot}. While preference evaluation is straightforward and intuitive, it has key limitations:
(1) Human preferences can be ambiguous and difficult to interpret, raising the question of whether they truly reflect rationale quality;
(2) The commonly used binary (win/loss) preference format obscures finer-grained insights into what makes one rationale better than another.

In this paper, we rethink the practice of human preference evaluation for LLM-generated rationales. Specifically, we explore the following research questions: \textbf{Q1}: What are the key attributes of a good rationale? \textbf{Q2}: Can human preferences over rationales be explained by these attributes, and if so, which attributes are most predictive? \textbf{Q3}: Can we use these attributes to offer more informative evaluations than existing binary preference evaluations?

To address Q1, we conduct a survey of recent literature on rationale evaluation and synthesize a set of core attributes that define high-quality rationales. 
These include diversity, faithfulness, hallucination, repetition, informativeness, perplexity, plausibility, self-consistency, and source consistency. We then operationalize these attributes using three evaluation methods: (1) existing automated metrics~\citep{golovneva2023roscoesuitemetricsscoring}, (2) LLM judges with both open- and closed-source LLMs, and (3) human annotations.

For Q2, we analyze whether these attributes explain human preferences by applying SHAP (SHapley Additive exPlanations) analysis \citep{NIPS2017_7062, lundberg2020local2global} to a LightGBM \citep{NIPS2016_10a5ab2d, NIPS2017_6449f44a} model trained to predict human preference using the rationale attributes as input features. 
While a previous paper conducts a similar analysis, it relies on one single model GPT4 for its analysis~\citep{hu2023decipherpref}.
We use two widely-used human preference datasets MT-Bench and Chatbot Arena \citep{zheng2023judgingllmasajudgemtbenchchatbot}, and treat their human annotations as the gold standard for this analysis.

To investigate Q3, we use these fine-grained attribute scores to re-evaluate LLM-generated rationales in the same two datasets. We compute ELO ratings per attribute and compare them against conventional ELO scores based on binary preference judgments, uncovering how fine-grained evaluations shift model rankings and offer more detailed insights.

While prior work has identified many of these attributes individually, our primary contribution lies in the holistic methodology we propose to bridge the gap between them and the coarse-grained human preferences that currently dominate LLM evaluation. We move beyond simply listing attributes by (1) empirically quantifying which attributes are most predictive of human judgment using SHAP analysis, thereby deconstructing the ambiguous signal of 'preference,' and (2) introducing attribute-specific ELO scores as a novel, fine-grained evaluation framework. This new framework allows for a more nuanced comparison of model capabilities, revealing specific trade-offs that are obscured by a single preference score.

Our study yields several key findings: first, there are a few common attributes that are most predictive of human preference across datasets and LLM judges: Correctness, Plausibility, and Completeness. Second, our re-evaluations with finegrained attributes reveal that while per-attribute ratings generally align with the generic ELO ratings based on binary human preference, they also uncover novel findings about models' strengths and weaknesses. We find that although GPT-4, GPT-3.5-turbo, and Claude-v1 emerge as victors, Claude-v1 struggles to avoid repetition and GPT-4 has lower arithmetic accuracy and self-consistency than GPT-3.5-turbo. 

Based on our results, we offer practical recommendations for future work on rationale evaluation:
First, researchers should move beyond binary preference evaluations and adopt fine-grained, attribute-level assessments to gain a more nuanced understanding of LLM-generated rationales.
Second, fine-grained evaluations can focus more on attributes that are more predictive of human preference, including  Correctness, Plausibility, Completeness, followed by Informativeness and Conciseness. Third, while LLM judges can be a scalable solution for fine-grained rationale evaluation, they should be used with caution: different models can produce diverging results, and we recommend using multiple LLM judges and reporting their outputs transparently to mitigate biases.

\section{Related Work} \label{relatedwork}

\paragraph{Human preference evaluation}
Human preference data have proven valuable for aligning LLMs with human values~\citep{christiano2023deepreinforcementlearninghuman, christiano2017deep, ouyang2022training, rafailov2023direct}. Beyond fine-tuning LLMs using this data, human preferences are also used to evaluate model performance. For instance, Chatbot Arena—a platform for evaluating LLMs based on human preferences—publishes human preference annotations on benchmark questions as well as real user queries~\citep{bai2024mt, zheng2023judgingllmasajudgemtbenchchatbot}.
\vspace{-3mm}
\paragraph{Rationale evaluation}
Multiple works have explored the evaluation of natural language free-text rationales. 
Works such as \cite{joshi2023machinerationalesnotuseful} have investigated the utility of such rationales to humans, emphasizing the gap between preference and utility and necessitating measures for rationale quality outside of human preference.  We derive our overall attribute definitions from \cite{golovneva2023roscoesuitemetricsscoring, ramnath2024tailoringselfrationalizersmultirewarddistillation, chen2023revinformationtheoreticevaluationfreetext, wiegreffe2022reframinghumanaicollaborationgenerating, rajani2019explainyourselfleveraginglanguage, atanasova2023faithfulnesstestsnaturallanguage, prasad2023recevalevaluatingreasoningchains, hase-etal-2020-leakage, wang2023pintofaithfullanguagereasoning}. \cite{wiegreffe2022reframinghumanaicollaborationgenerating} evaluates free-text explanations from GPT-3 on attributes such as "providing new information", factuality, and grammaticality. \cite{chen2023revinformationtheoreticevaluationfreetext} proposes a metric to grade rationales on novelty of information as well.  \cite{ramnath2024tailoringselfrationalizersmultirewarddistillation} evaluates rationales on the properties of plausibility, diversity, and consistency, highlighting their usefulness to humans.
Fewer works have explored evaluation of step-by-step reasoning, or chain-of-thought explanations. ROSCOE~\citep{golovneva2023roscoesuitemetricsscoring} provides a set of metrics specifically for evaluating step-by-step rationales, i.e., chain-of-thought explanations; hence we use ROSCOE for our automated metrics.

\section{Methods}
To define a "good" model-generated rationale and move beyond binary "chosen" and "rejected" preference labels, we identify a set of 12 key attributes from prior work on rationale evaluation. 
In this section, we first define each of these attributes. Then, we describe three approaches to measure these attributes. 
\vspace{-3mm}
\subsection{Attributes}
\vspace{-2mm}
We evaluate LLM-generated rationales using a set of 12 fine-grained attributes (Table~\ref{tab:attribute-definitions}) which capture key aspects of rationale quality as identified in recent literature (Section \ref{relatedwork}). 
\begin{table}[ht]
\centering
\begin{tabular}{p{3.5cm} p{10cm}}
\toprule
\textbf{Attribute} & \textbf{Definition} \\
\midrule
Faithfulness & Is the rationale supported by the model’s actual computation or the provided evidence? \\ 
Hallucination & Does the rationale introduce information not present in the source/context? \\
Repetition & Does the rationale unnecessarily repeat points or phrases? \\
Informativeness & Does the rationale add meaningful, relevant details? \\
Plausibility & Does the rationale ``sound right'' or seem believable, regardless of truth? \\
Self-Consistency & Does the rationale avoid contradictions within itself, with all reasoning steps logically aligned? \\
Source Consistency & Does the rationale avoid contradicting the given context or information in the problem statement? \\
Grammar & Is the rationale well-written, clear, and free of grammatical mistakes? \\
Arithmetic Accuracy & Are any calculations in the rationale correct? \\
Conciseness & Is the rationale as short as possible, without losing information? Especially if length is a concern. \\
Completeness & Does the rationale explain all necessary steps/evidence? \\
Correctness & Are all steps and answers in the rationale objectively correct? \\
\bottomrule
\end{tabular}
\caption{Definitions of rationale quality attributes used in our analysis.}
\label{tab:attribute-definitions}
\end{table}
\vspace{-3mm}
\subsection{Attribute measurements}
\vspace{-2mm}
We measure these attributes of rationales using three approaches: (1) automated heuristics; (2) LLM judges; and (3) human annotations. 

For automated scoring, we use ROSCOE metrics, which quantify aspects of rationale quality using interpretable heuristics and alignment scores. Table~\ref{tab:roscoe-metrics} in Appendix~\ref{roscoe-appendix} summarizes the specific ROSCOE metrics and their descriptions. While ROSCOE metrics provide a baseline for automated rationale scoring, we observe several limitations: their scores can be noisy, as they require step-by-step formats and are highly sensitive to rationale style. A detailed analysis of ROSCOE metric performance is in Appendix~\ref{roscoe-results-appendix}.

To address these issues, we instead focus on attribute scoring with LLM judges, which allow for greater flexibility. Unlike formulaic ROSCOE metrics, LLM judges can interpret a variety of response formats much like human annotators. We prompt multiple LLMs to evaluate each attribute, including the closed-source GPT-4o and Gemini 2.5-Flash models (scoring on a 0–1 scale), as well as the open-source OLMo 32B model (\cite{olmo20252olmo2furious}) (scored on a 0–10 scale, as OLMo produces more reliable and calibrated scores with integer-valued prompts). Using both closed- and open-source models helps mitigate concerns that closed models may have been trained on our evaluation data. Exact prompt templates for each model can be found in Appendix~\ref{prompts-appendix}.

Lastly, for human annotations, the three co-first-authors annotate a randomly sampled set of rationales from the two human preference datasets on all the attributes. Human scores also range from 0 to 1. Human annotations results can be found in Appendix~\ref{human-results-appendix}.

\begin{figure}[t]
    \centering
    \includegraphics[width=0.7\linewidth]{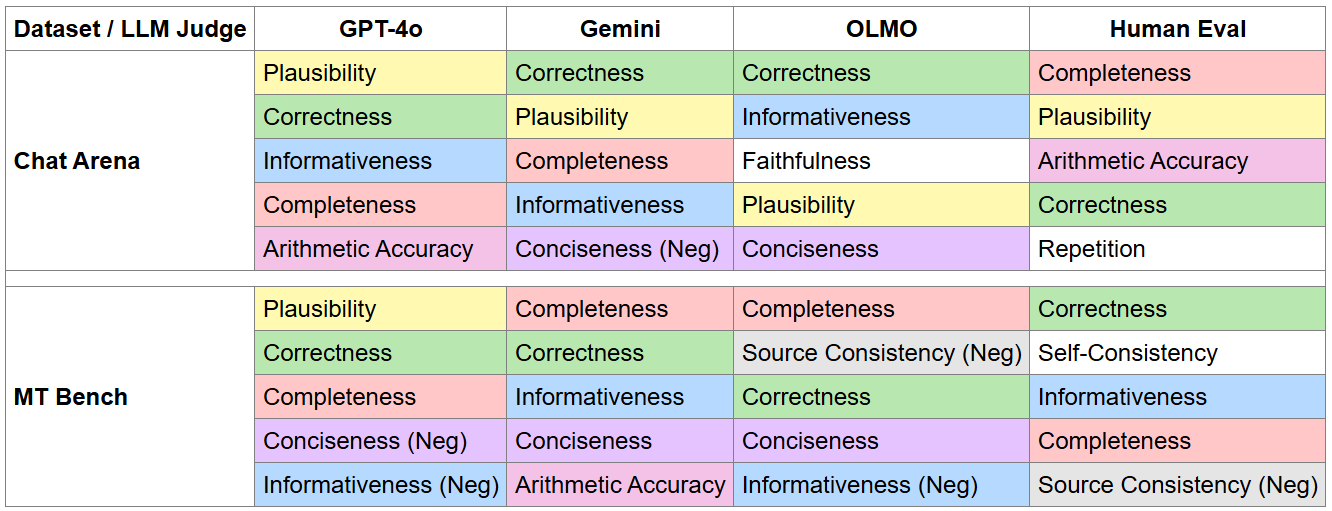}
    \caption{
       Most influential attributes as identified by SHAP value analysis on Chatbot Arena and MT Bench across LLM judges and human annotators. “Neg” indicates negative influence on predicted preference.
    }
    \label{fig:top-attributes-human-pref}
    \vspace{-5mm}
\end{figure}

\vspace{-5mm}
\begin{figure}[ht]
    \centering
    \begin{subfigure}[t]{0.45\textwidth}
        \centering
        \includegraphics[width=\linewidth]{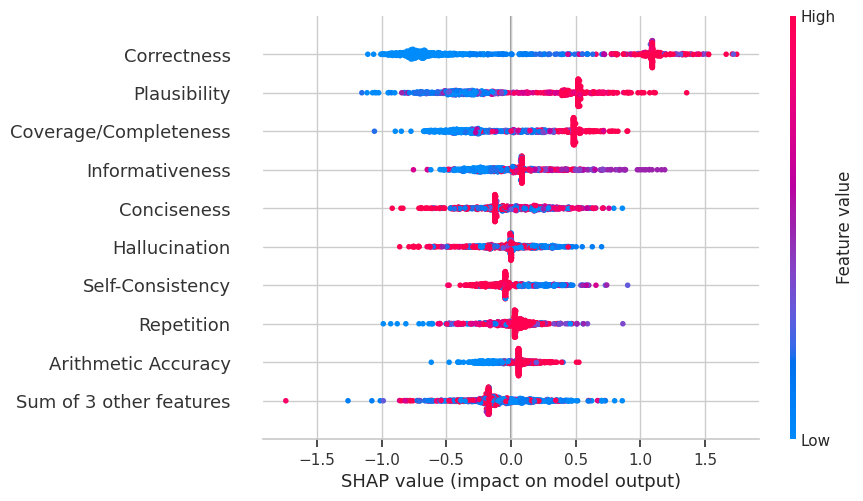}
        \caption{SHAP beeswarm plot}
    \end{subfigure}
    \hfill
    \begin{subfigure}[t]{0.45\textwidth}
        \centering
        \includegraphics[width=\linewidth]{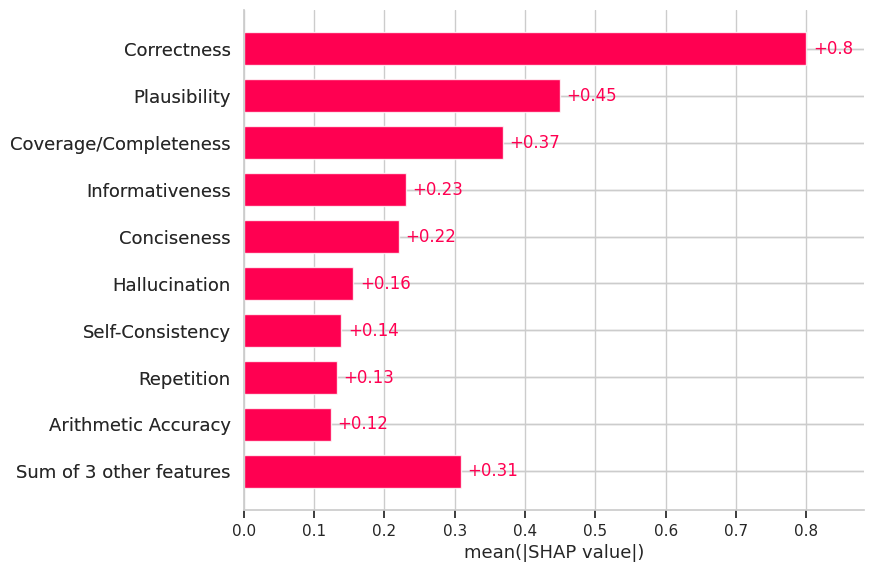}
        \caption{Mean absolute SHAP value plot}
    \end{subfigure}
    \caption{
        SHAP analysis of Gemini-2.5-Flash on Chatbot Arena. (a) Beeswarm plot shows SHAP value distribution and direction per attribute. (b) Bar plot shows overall attribute impact via mean absolute SHAP values.
    }
    \label{fig:shap_side_by_side}
    \vspace{-3mm}
\end{figure}

\section{Experiments}
To answer Q2 and Q3, we conduct two sets of experiments with the 12 fine-grained attributes on two human preference datasets: MT-Bench and Chatbot Arena~\citep{zheng2023judgingllmasajudgemtbenchchatbot}. 
\vspace{-4mm}
\paragraph{Datasets.}
Chatbot Arena is a dataset of model responses and human preference annotations collected in a tournament-style setup~\citep{chiang2024chatbotarenaopenplatform}, where pairwise model responses are judged by humans. 
It also includes human preference annotations on MT-Bench, a benchmark for fine-LLM evaluation in multi-turn dialogues~\citep{bai2024mt}. Because Chatbot Arena consists of user queries in the wild where model responses might not be rationales, we use GPT-4o to filter for mathematical and logical questions, resulting in \textbf{1,367 questions} with step-by-step or adjacent responses. For MT Bench, we retain general and mathematical reasoning questions, yielding a total \textbf{80 unique questions}.


\vspace{-3mm}
\paragraph{Analysis of human preference.}
To address Q2 and assess the relative importance of fine-grained attributes in shaping human preferences, we employ SHAP analysis on predictions from a LightGBM model. We prefer SHAP over simple correlation analysis because it captures complex, nonlinear interactions among attributes and provides instance-level interpretability.
For the predictive model, we choose LightGBM, a gradient-boosted decision tree, due to its efficiency, scalability to large datasets, and interpretable feature interactions. In this analysis, the attributes serve as input features (X), and human preference scores constitute the target variable (y). SHAP values offer insights into which attributes significantly enhance or diminish the likelihood of a rationale being preferred by human annotators.

\vspace{-3mm}
\paragraph{Re-evaluations of LLM rationales.}
Chatbot Arena along with other previous works \citep{bai2022traininghelpfulharmlessassistant} have used ELO rankings to rank models based on binary preference outcomes. 
We propose a more fine-grained metric: we compute attribute-specific ELO ratings. For each attribute (e.g., faithfulness, informativeness, etc.), we use its LLM judge score instead of the binary human preference label to determine the winning model. We then re-compute ELO rankings per attribute for each model, averaging the score of all three LLM judges. This enables us to evaluate models on each dimension of rationale quality.

\begin{figure}[t]
    \centering
    \begin{subfigure}[t]{0.45\textwidth}
        \centering
        \includegraphics[width=\linewidth]{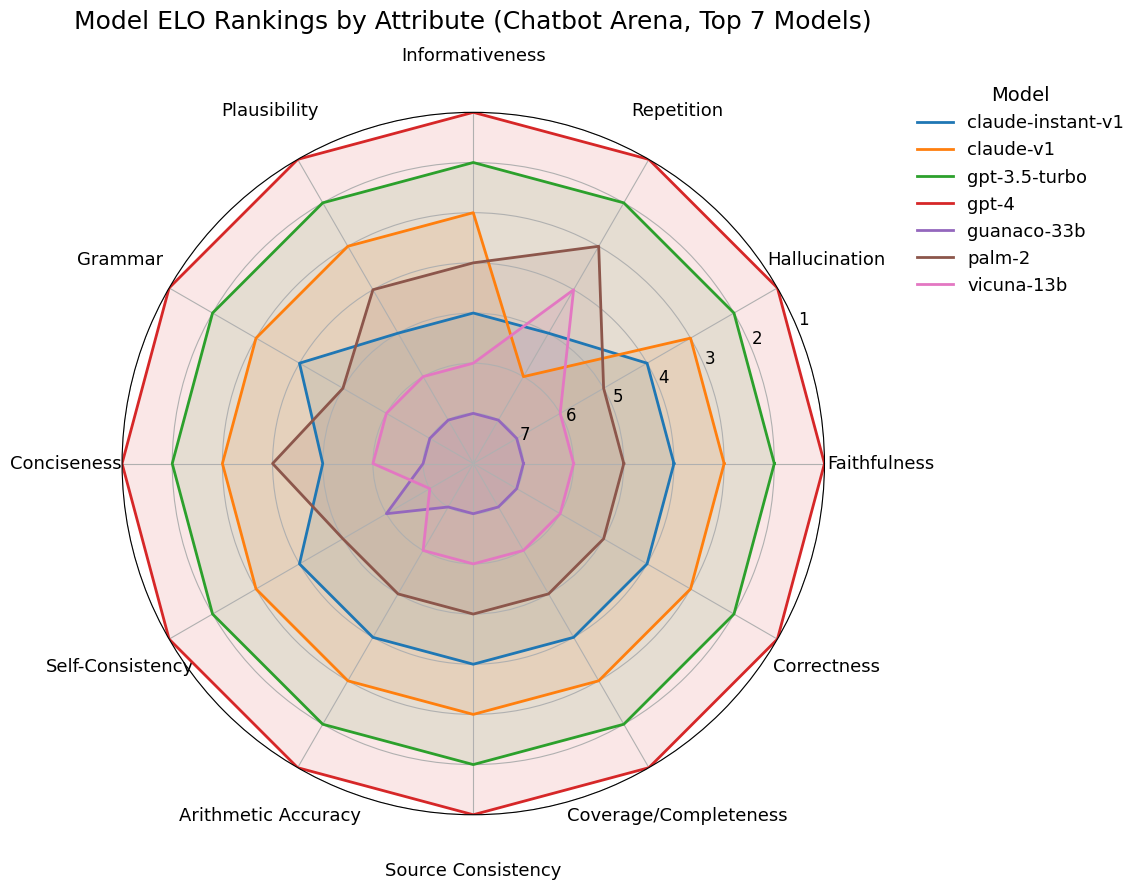}
        \caption{Chatbot Arena}
    \end{subfigure}
    \hfill
    \begin{subfigure}[t]{0.45\textwidth}
        \centering
        \includegraphics[width=\linewidth]{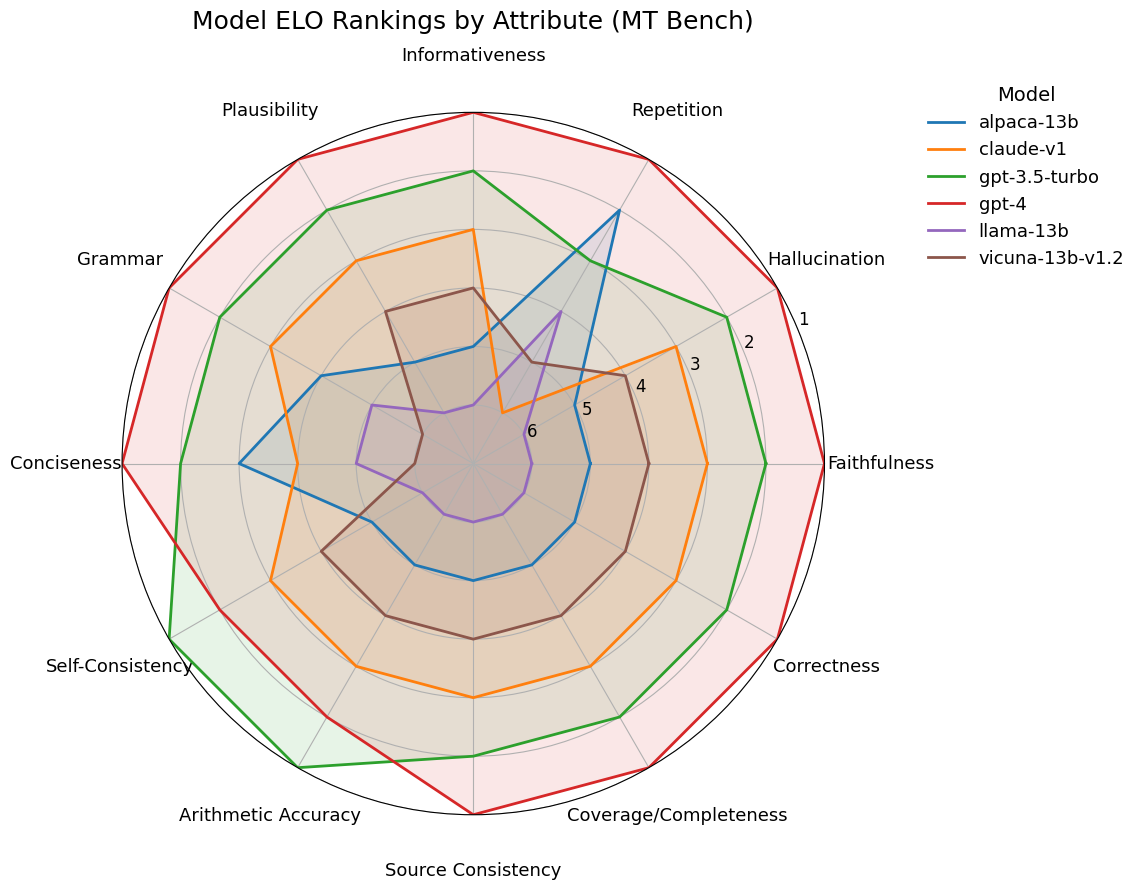}
        \caption{MT Bench}
    \end{subfigure}
    \caption{
    Radar charts of model ELO rankings by attribute on Chatbot Arena (a) and MT Bench (b). Lower rank (outer ring) indicates better performance. 
    }
    \label{fig:elo_rankings_radar_side_by_side}
\end{figure}


\vspace{-3mm}
\subsection{Results}
\vspace{-2mm}
\paragraph{Q2: Which attributes are most predictive of human preferences over rationales?}
From our feature importance analysis using SHAP values, we find that attributes Correctness, Plausibility, and Completeness are among the top predictors of human preference across models and datasets (Figure~\ref{fig:top-attributes-human-pref}), suggesting that human annotators place particular emphasis on the factual accuracy, plausibility, and thoroughness of rationales when making their judgments. Detailed SHAP values for each setting are presented in Appendix~\ref{human-results-appendix}.
\vspace{-3mm}
\paragraph{Q3: Can fine-grained attributes offer more informative evaluations?} 
We re-evaluate LLMs on the two preference datasets using fine-grained rationale attributes. 
We find that while attribute-specific ELO rankings generally align with overall human preference rankings, they reveal unique insights about models' strengths and weaknesses. Across both Chatbot Arena and MT-Bench datasets, GPT-4, GPT-3.5-Turbo, and Claude-v1 consistently occupy the top three positions, although the order varies with dataset and attribute (Figure~\ref{fig:elo_rankings_radar_side_by_side}). Interestingly, Claude-v1 scores poorly on Repetition in both datasets and on MT-Bench, GPT-3.5-Turbo unexpectedly outperforms GPT-4 on certain attributes such as Self-Consistency and Arithmetic Accuracy.

These examples suggest that while rankings based on generic human preferences coarsely identify overall strongest models, fine-grained attribute-based evaluations can facilitate comparison of models' strengths and weaknesses across multiple rationale quality dimensions, revealing subtle but meaningful differences between top-performing models. A detailed analysis of the ELO rankings for each attribute is presented in Appendix~\ref{elo-results-appendix}.

\section{Limitations and Future work}

\vspace{-3mm}
\subsection{Limitations of the Current Study}
\vspace{-2mm}
\begin{itemize}
    \item \textbf{Reliability and Bias of LLM Judges:} A primary limitation is the reliance on LLMs for both dataset filtering and attribute scoring. While we took steps to mitigate this by using multiple judges, including both open- and closed-source models, the "LLM-as-a-judge" paradigm has known challenges. These include non-deterministic outputs even at zero temperature, potential factual errors (as shown in Appendix~\ref{gpt4o-mistakes-example}), and the inherent bias of the judge models, which may have been trained on similar data or possess their own stylistic preferences. A more systematic evaluation of inter-judge consistency would be needed to fully quantify this variability.

    \item \textbf{Scope of Evaluation Tasks:} Our analysis focused on mathematical and logical reasoning tasks. While these domains are critical for evaluating reasoning, it remains an open question how the importance of specific attributes might shift in other contexts, such as creative writing, commonsense inference, or multi-step planning tasks. Future work should apply this attribute-based framework to a wider variety of tasks to test the generalizability of our findings.

    \item \textbf{Scope of Human Annotation:} Our human annotation study provided a crucial foundation for our analysis. However, these annotations were performed by the three co-first authors. Although this ensures expert-level evaluation, a future study involving a larger and more diverse pool of human annotators would strengthen the conclusions about which attributes are most predictive of general human preference. Such a study should also include formal measurements of inter-annotator agreement (IAA) to further validate the attribute definitions.
\end{itemize}

\vspace{-3mm}
\subsection{Directions for Future Work}
\vspace{-2mm}
\begin{itemize}
    \item \textbf{Improving Rationale Quality via Attribute-Based Fine-Tuning:} A key implication of our work is the potential to improve models by optimizing for specific, desirable attributes. Our fine-grained attribute labels offer a much richer training signal than binary preference data. Future work could leverage these attribute-level annotations to fine-tune models using techniques like Direct Preference Optimization (DPO), but with a multi-objective reward function. This could allow for targeted improvements, such as increasing a model's \textit{Correctness} and \textit{Self-Consistency} while simultaneously penalizing \textit{Repetition}.


    \item \textbf{Connecting to Interpretability and Human-AI Collaboration:} Finally, our framework can be more deeply connected to existing work on the human utility of explanations. Future studies could investigate how different rationale profiles (e.g., high \textit{Completeness} but low \textit{Conciseness}) affect human performance on downstream tasks, trust in the model, and the ability to detect model errors. This would bridge the gap between the quality of abstract rationale and the practical utility of human nature.
\end{itemize}

\section{Conclusion}

The evaluation of LLM-generated rationales has largely relied on coarse-grained, binary preferences, which obscure the specific qualities that make a rationale effective. We propose a fine-grained, attribute-based evaluation framework and show that human choices are strongly driven by attributes such as \textbf{Correctness}, \textbf{Plausibility}, and \textbf{Completeness}.

Our attribute-specific ELO rankings reveal nuanced model trade-offs—for example, GPT-3.5-Turbo surpasses GPT-4 on \textbf{Arithmetic Accuracy} and \textbf{Self-Consistency}. Such diagnostic insights, absent in holistic preference scores, are crucial for targeted model improvement. We argue that future evaluation should move beyond monolithic metrics toward fine-grained, interpretable frameworks that foster more transparent and trustworthy LLMs.

\bibliography{colm2025_conference}
\bibliographystyle{colm2025_conference}

\newpage
\appendix
\section{Appendix}
\subsection{ROSCOE Metrics} \label{roscoe-appendix}

\begin{table}[ht]
\centering
\begin{tabular}{p{4cm} p{9cm}}
\toprule
\textbf{ROSCOE Metric Attributes} & \textbf{Description} \\
\midrule
faithfulness & Mean alignment from the hypothesis chain to the context sentences; higher scores indicate better grounding by the context. \\
faithfulness\_ww & Mean alignment of the sentence and token embeddings from the hypothesis chain to the context sentences and tokens. \\
repetition\_word & For each step, gets the maximum alignment score between tokens in the current step and tokens in previous steps (token-level repetition). \\
repetition\_step & Maximum cosine similarity of each step to all previous steps (sentence-level repetition). \\
informativeness\_step & Mean alignment from the sentences in the context to all steps in the chain and vice versa, averaged. \\
informativeness\_chain & Cosine similarity between the overall hypothesis embedding and the context embedding. \\
discourse\_representation & Maximum predicted probability of contradiction (from NLI model) between each step in the hypothesis and each sentence in the context. \\
coherence\_step\_vs\_step & Maximum probability of contradiction (from NLI model) between each step and all previous steps in the chain. \\
perplexity\_step & Perplexity of each step, averaged over the chain. \\
perplexity\_step\_max & Maximum perplexity among all steps, where each step is scored individually. \\
perplexity\_chain & Perplexity of the entire chain taken as a continuous string. \\
grammar\_step & Grammatical correctness of each step, as predicted by a grammaticality classifier and averaged over the chain. \\
grammar\_step\_max & Grammatical correctness of each step; minimum score given to a step (most incorrect step's score is used). \\
\bottomrule
\end{tabular}
\caption{Descriptions of automated ROSCOE metrics used for rationale attribute evaluation.}
\label{tab:roscoe-metrics}
\end{table}

\newpage
\subsection{ROSCOE Metrics Results} \label{roscoe-results-appendix}
\subsubsection{Chatbot Arena}

\begin{figure}[H]
    \centering
    \includegraphics[width=0.9\linewidth]{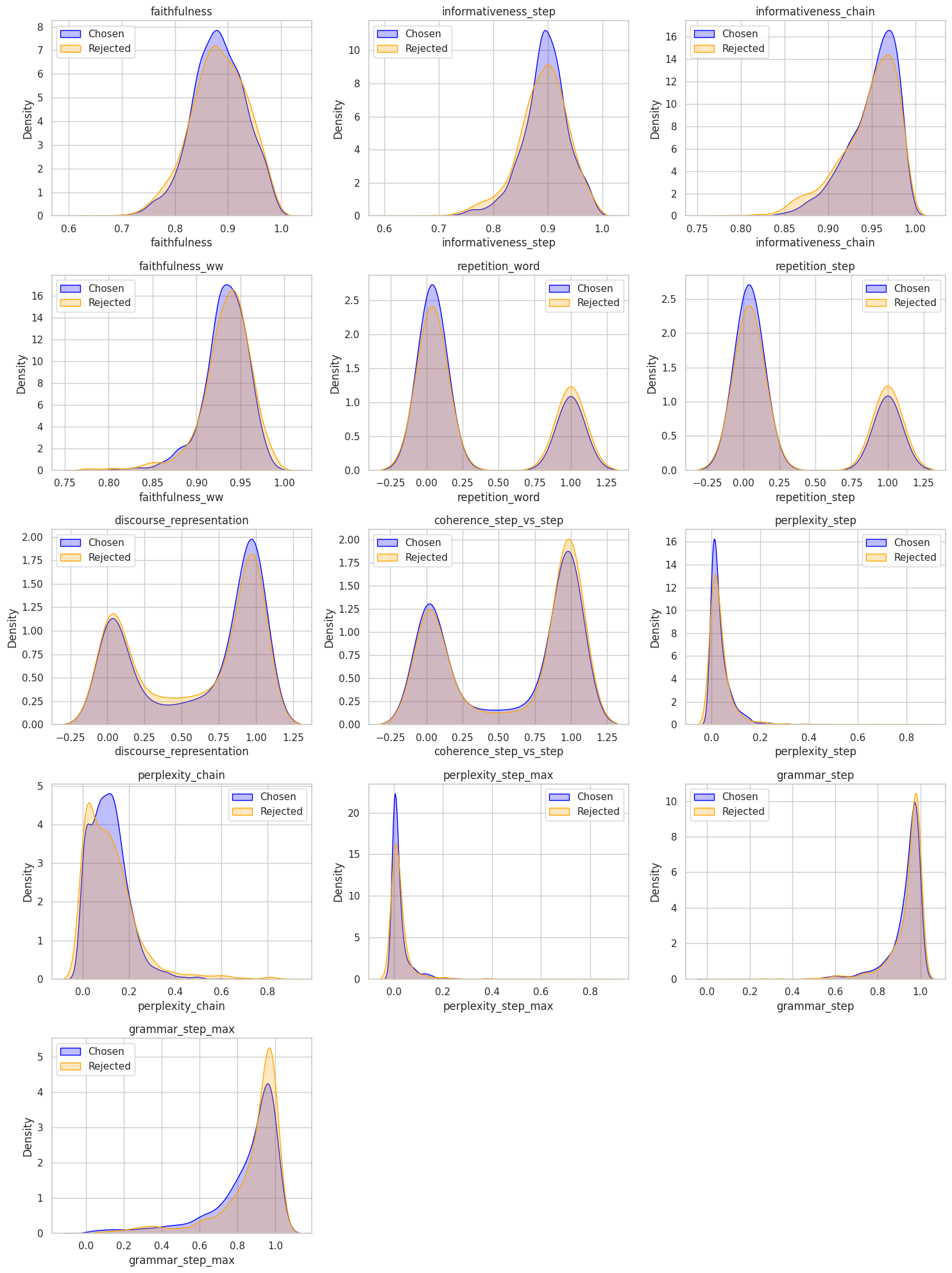}
    \caption{Distribution of the difference between chosen and rejected scores by attribute. Boxplots summarize the (chosen -- rejected) difference for each attribute.}
    \label{fig:roscoe-boxplot-chat}
\end{figure}

\begin{figure}[H]
    \centering
    \includegraphics[width=0.9\linewidth]{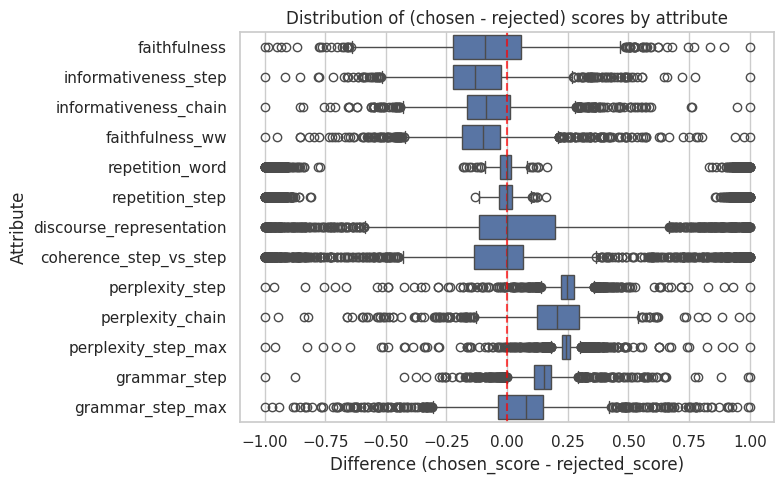}
    \caption{Distribution of attribute values for chosen vs.\ rejected rationales. Each subplot shows the density of scores for each attribute.}
    \label{fig:roscoe-density-chat}
\end{figure}

\begin{figure}[H]
    \centering
    \includegraphics[width=0.9\linewidth]{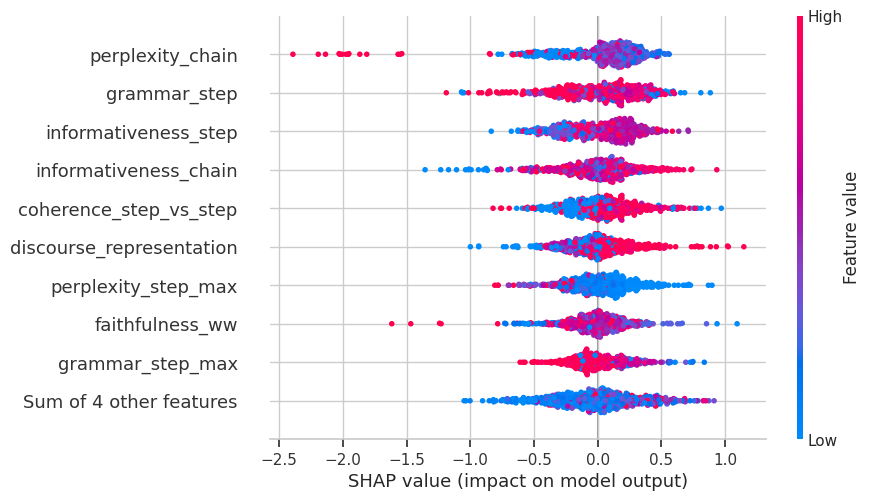}
    \caption{Mean absolute SHAP value plot for Chatbot Arena (ROSCOE). Shows the mean importance of each attribute in the model.}
    \label{fig:roscoe-shap-beam-chat}
\end{figure}

\begin{figure}[H]
    \centering
    \includegraphics[width=0.9\linewidth]{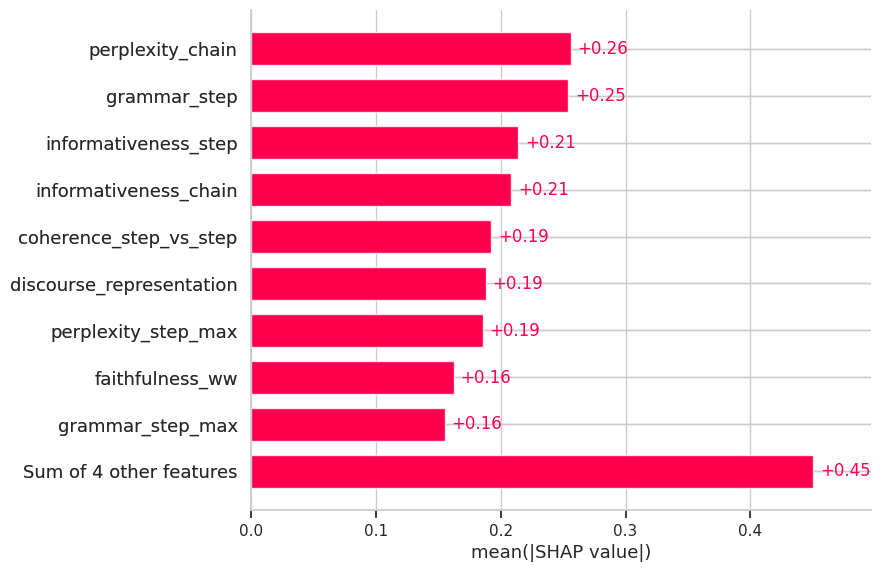}
    \caption{SHAP beeswarm plot for Chatbot Arena (ROSCOE). Visualizes the distribution and direction of SHAP values for each attribute.}
    \label{fig:roscoe-shap-beeswarm-chat}
\end{figure}

\subsubsection{Mt Bench}
\begin{figure}[H]
    \centering
    \includegraphics[width=0.9\linewidth]{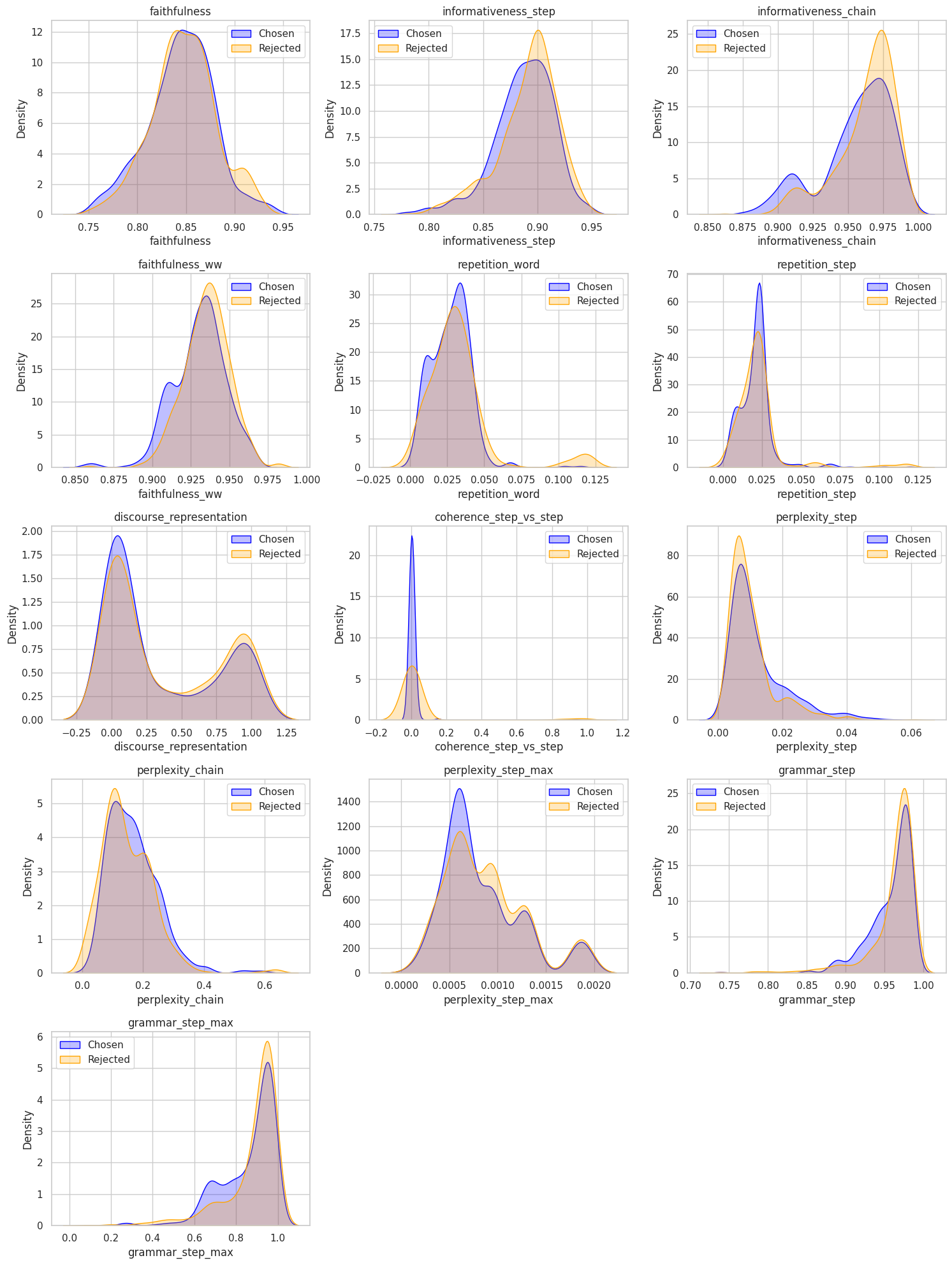}
    \caption{Distribution of the difference between chosen and rejected scores by attribute in MT Bench. Boxplots summarize the (chosen -- rejected) difference for each attribute.}
    \label{fig:roscoe-boxplot-mt}
\end{figure}

\begin{figure}[H]
    \centering
    \includegraphics[width=0.9\linewidth]{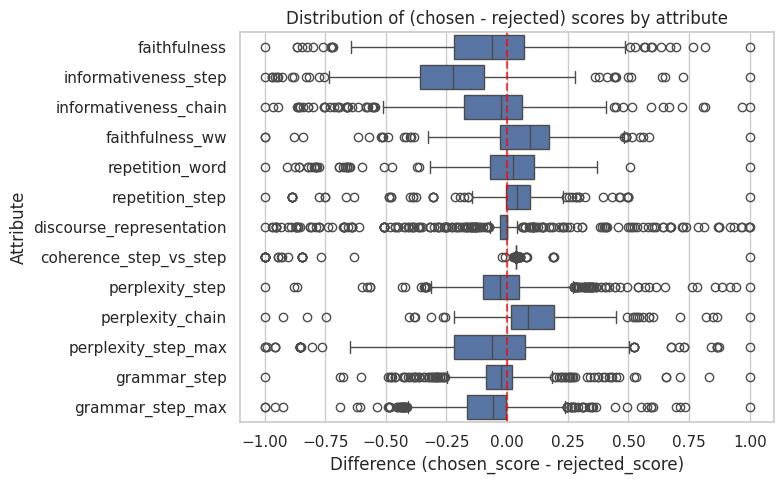}
    \caption{Distribution of attribute values for chosen vs.\ rejected rationales in MT Bench. Each subplot shows the density of scores for each attribute.}
    \label{fig:roscoe-density-mt}
\end{figure}

\begin{figure}[H]
    \centering
    \includegraphics[width=0.9\linewidth]{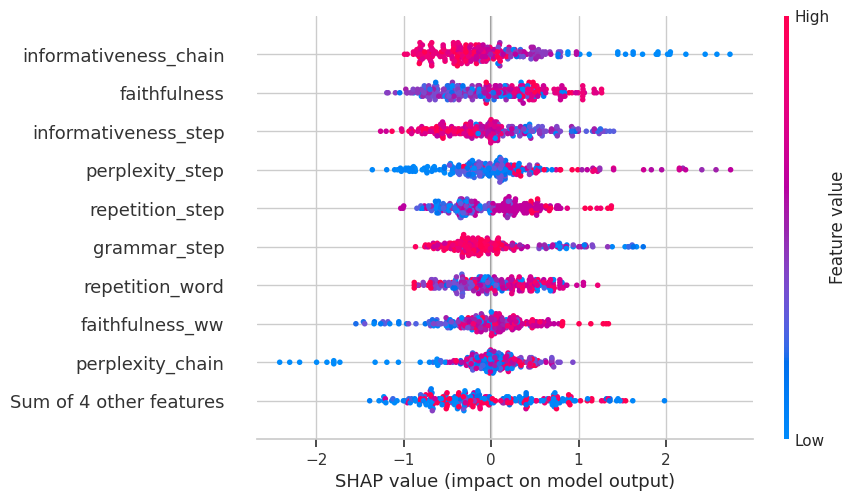}
    \caption{SHAP beeswarm plot for MT Bench (ROSCOE). Visualizes the distribution and direction of SHAP values for each attribute.}
    \label{fig:roscoe-shap-beeswarm-mt}
\end{figure}

\begin{figure}[H]
    \centering
    \includegraphics[width=0.9\linewidth]{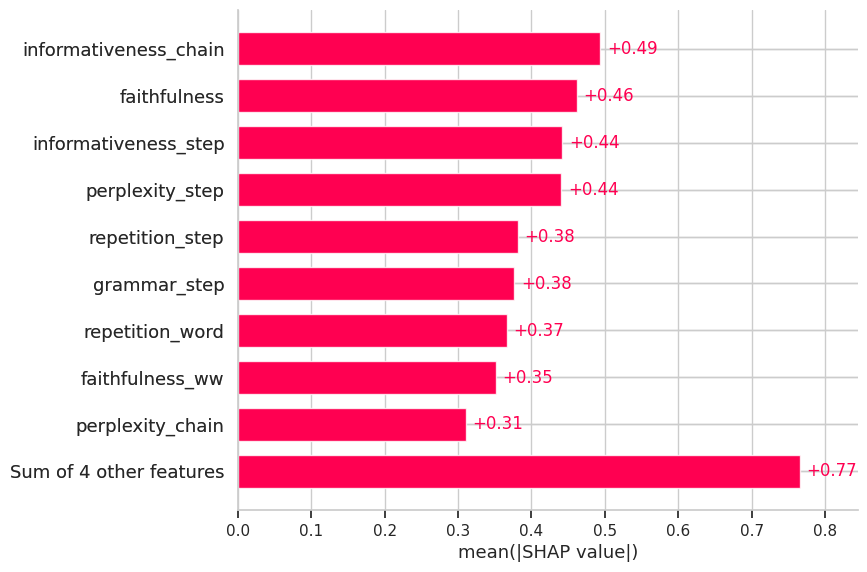}
    \caption{Mean absolute SHAP value plot for MT Bench (ROSCOE). Shows the mean importance of each attribute in the model.}
    \label{fig:roscoe-shap-beam-mt}
\end{figure}

\newpage
\subsection{Prompt Templates for Automated LLM Scoring} \label{prompts-appendix}
\subsubsection{Main Prompt Template (0--1 Scale; for GPT-4o and Gemini 2.5-Flash)}

\begin{verbatim}
You are an expert evaluator for mathematical and logical explanations. 
Given the following question and rationale, please assign a score between 0 (worst) and 1
(best) for each attribute listed below. For each attribute, also provide a brief justification.

IMPORTANT: A score of **1.0 always means BEST** and **0.0 always means WORST**, even for
attributes like Hallucination and Repetition. Interpret all attributes as “more of a good thing.”

Please return your response in two Python dictionaries:

- One called `scores` where each key is the name of the attribute and the value is the score
(a float between 0 and 1).

- One called `explanations` where each key is the name of the attribute and the value is your
explanation (1-2 sentences) for that score.

Please output the Python dictionaries as plain text only—do not include code blocks,
markdown, or any extra formatting.

Here are the attributes and their definitions:

Faithfulness:
Is the rationale supported by the model’s actual computation or the provided evidence?

Hallucination:
Does the rationale introduce information not present in the source/context?

Repetition:
Does the rationale unnecessarily repeat points or phrases?

Informativeness:
Does the rationale add meaningful, relevant details?

Plausibility:
Does the rationale “sound right” or seem believable, regardless of truth?

Self-Consistency:
The rationale does not contain steps that contradict each other; all reasoning is logically
aligned internally.

Source Consistency:
The rationale does not contradict the given context or information in the problem statement.

Grammar:
Is the rationale well-written, clear, and free of grammatical mistakes?

Arithmetic Accuracy:
Are any calculations in the rationale correct?

Conciseness:
Is it as short as possible, without losing information? Especially if length is a concern.

Coverage/Completeness:
Does it explain all necessary steps/evidence?

Correctness:
Are all steps and answers objectively correct?

Example output format:

scores = {
    "Faithfulness": 0.95,
    "Hallucination": 0.67,
    "Repetition": 0.89,
    ...
}

explanations = {
    "Faithfulness": "The rationale closely follows logical steps derived from the question.",
    "Hallucination": "Some external information or assumptions were introduced. For example, ...",
    "Repetition": "The rationale is does not repeat it self with similar points at different steps.",
    ...
}

Math/Logic Question:
{question}

Rationale:
{rationale}
\end{verbatim}

\newpage 
\subsubsection{OLMO Prompt Template (0--10 Scale)}

\begin{verbatim}
You are an expert evaluator for mathematical and logical explanations. 

Given the following question and rationale, please assign a score between 0 (worst) 
and 10 (best) for each attribute listed below. For each attribute, also provide a 
brief justification.

IMPORTANT: A score of **10 always means BEST** and **0 always means WORST**, 
even for attributes like Hallucination and Repetition. 
Interpret all attributes as “more of a good thing.”

Please return your response in two Python dictionaries:

- One called `scores` where each key is the name of the attribute and the value is the
score (a float between 0 and 10).

- One called `explanations` where each key is the name of the attribute and the value
is your explanation (1-2 sentences) for that score.

Please output the Python dictionaries as plain text only, do not include code blocks, 
markdown, or any extra formatting.

Here are the attributes and their definitions:

Faithfulness:
Is the rationale supported by the model’s actual computation or the provided evidence?

Hallucination:
Does the rationale introduce information not present in the source/context?

Repetition:
Does the rationale unnecessarily repeat points or phrases?

Informativeness:
Does the rationale add meaningful, relevant details?

Plausibility:
Does the rationale “sound right” or seem believable, regardless of truth?

Self-Consistency:
The rationale does not contain steps that contradict each other;
all reasoning is logically aligned internally.

Source Consistency:
The rationale does not contradict the given context or information in the problem statement.

Grammar:
Is the rationale well-written, clear, and free of grammatical mistakes?

Arithmetic Accuracy:
Are any calculations in the rationale correct?

Conciseness:
Is it as short as possible, without losing information? Especially if length is a concern.

Coverage/Completeness:
Does it explain all necessary steps/evidence?

Correctness:
Are all steps and answers objectively correct?

Example output format:

scores = {
    "Faithfulness": 9.5,
    "Hallucination": 6.8,
    "Repetition": 8.9,
    ...
}

explanations = {
    "Faithfulness": "The rationale closely follows logical steps derived from the question.",
    "Hallucination": "Some external information or assumptions were introduced. For example, ...",
    "Repetition": "The rationale is does not repeat it self with similar points at different steps.",
    ...
}

Math/Logic Question:
{question}

Rationale:
{rationale}
\end{verbatim}

\newpage
\subsection{LLM Judges Results} \label{judge-results-appendix}
\subsubsection{OpenAI GPT-4o}
\begin{enumerate}
    \item \textbf{Chatbot Arena}
    \begin{figure}[H]
        \centering
        \includegraphics[width=0.8\linewidth]{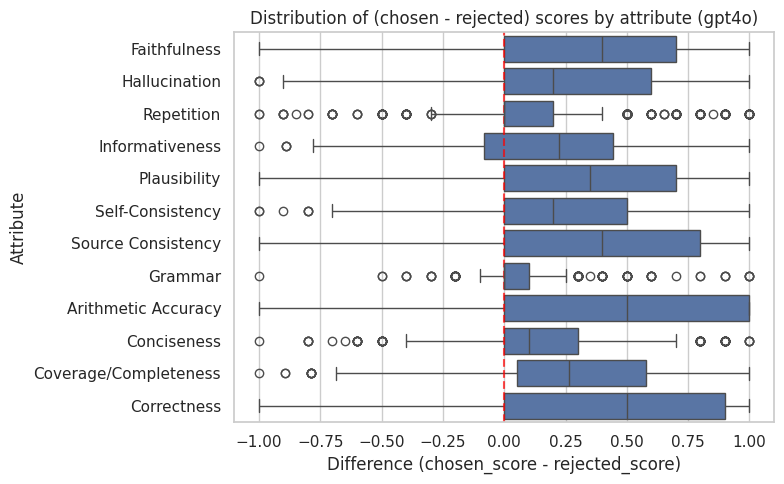}
        \caption{Distribution of the difference between chosen and rejected scores by attribute in Chatbot Arena (GPT-4o). Boxplots summarize the (chosen -- rejected) difference for each attribute.}
        \label{fig:gpt4o-diff-chat}
    \end{figure}
    
    \begin{figure}[H]
        \centering
        \includegraphics[width=0.8\linewidth]{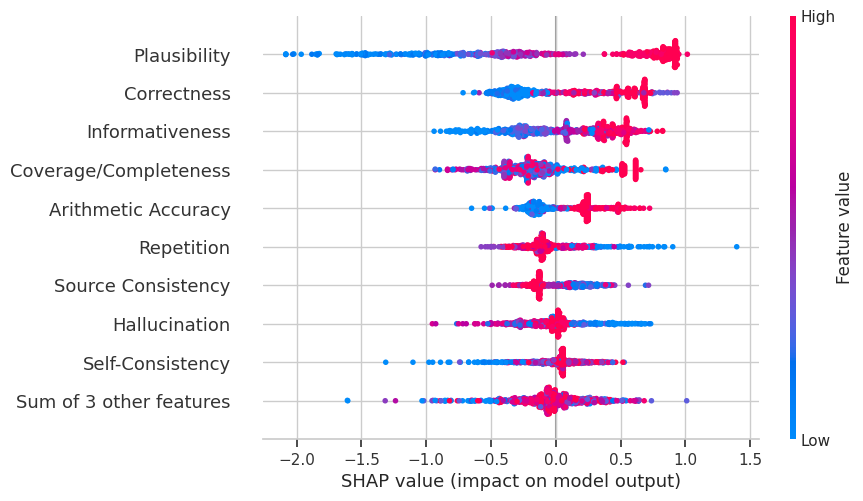}
        \caption{SHAP beeswarm plot for Chatbot Arena (GPT-4o). Visualizes the distribution and direction of SHAP values for each attribute.}
        \label{fig:gpt4o-beeswarm-chat}
    \end{figure}
    
    \begin{figure}[H]
        \centering
        \includegraphics[width=0.8\linewidth]{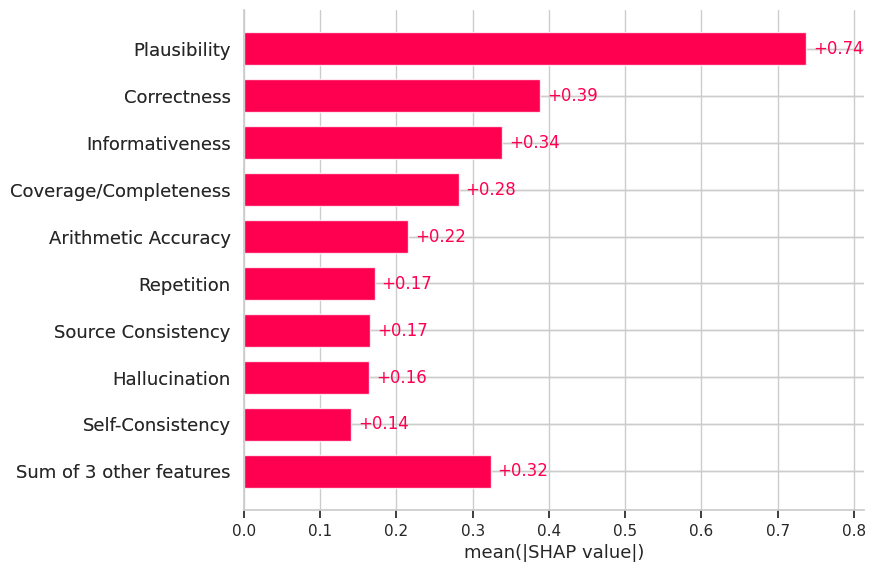}
        \caption{Mean absolute SHAP value plot for Chatbot Arena (GPT-4o). Shows the mean importance of each attribute in the model.}
        \label{fig:gpt4o-shap-chat}
    \end{figure}
    
    \begin{figure}[H]
        \centering
        \includegraphics[width=0.9\linewidth]{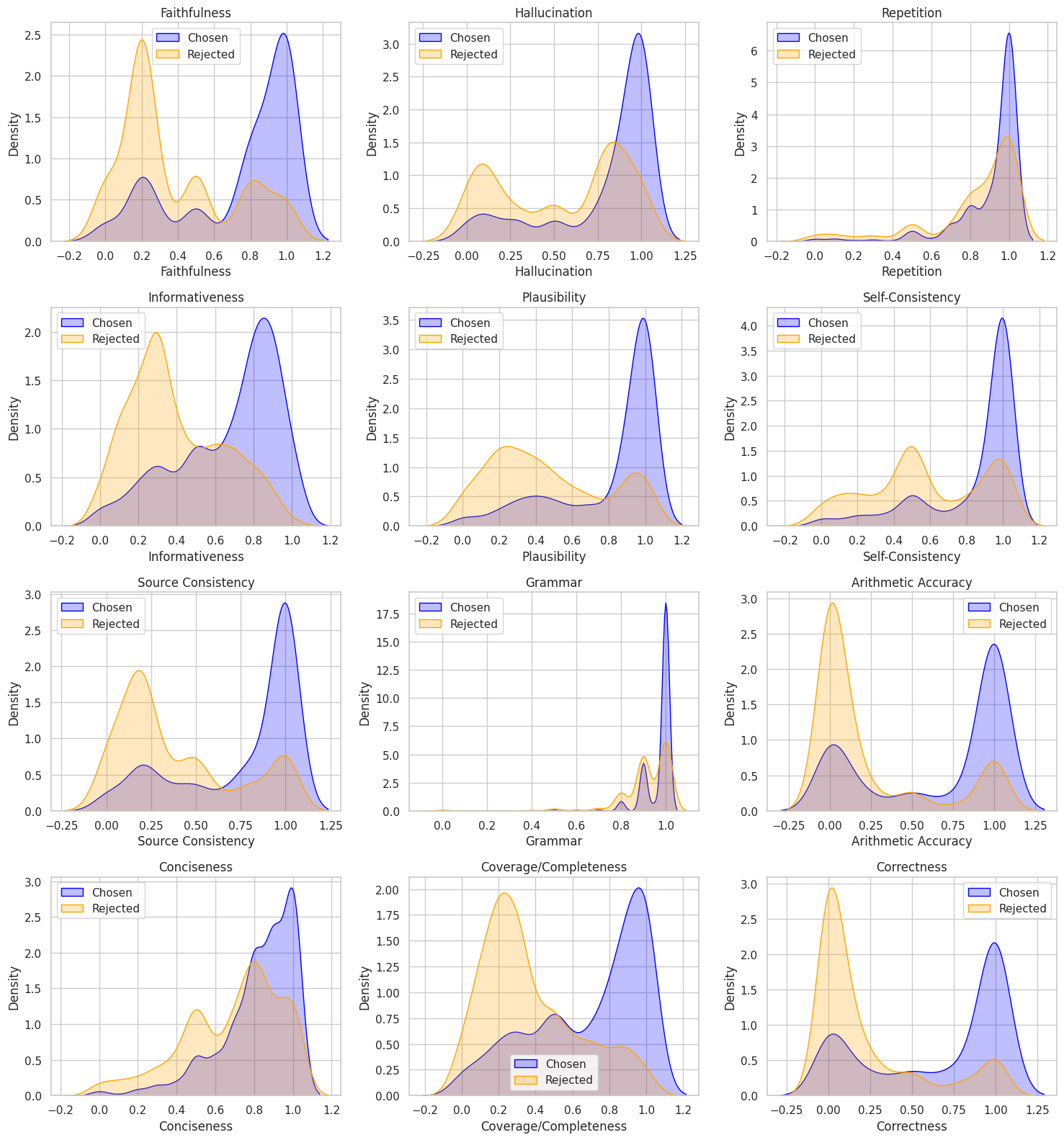}
        \caption{Distribution of attribute values for chosen vs.\ rejected rationales in Chatbot Arena (GPT-4o). Each subplot shows the density of scores for each attribute.}
        \label{fig:gpt4o-density-chat}
    \end{figure}

    \newpage
    \item \textbf{Mt Bench}
    \begin{figure}[H]
        \centering
        \includegraphics[width=0.8\linewidth]{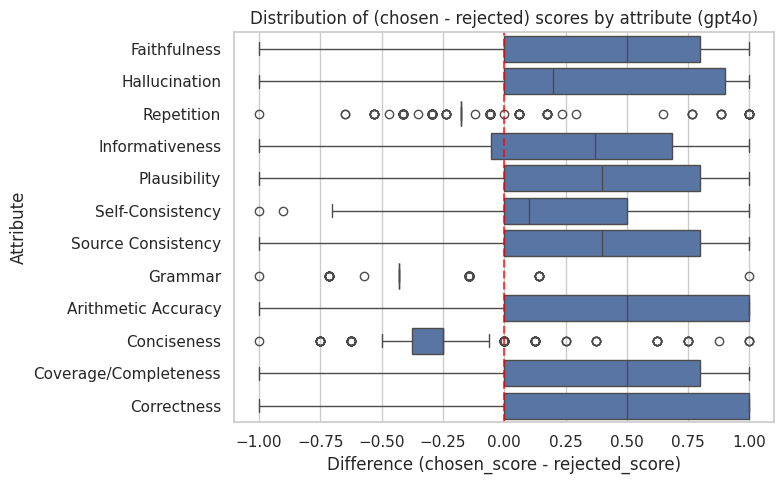}
        \caption{Distribution of the difference between chosen and rejected scores by attribute in MT Bench (GPT-4o). Boxplots summarize the (chosen -- rejected) difference for each attribute.}
        \label{fig:gpt4o-diff-mt}
    \end{figure}
    
    \begin{figure}[H]
        \centering
        \includegraphics[width=0.8\linewidth]{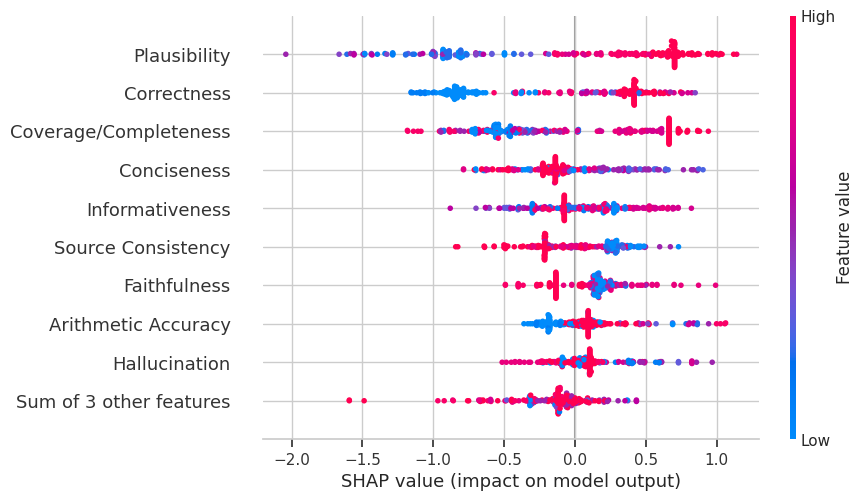}
        \caption{SHAP beeswarm plot for MT Bench (GPT-4o). Visualizes the distribution and direction of SHAP values for each attribute.}
        \label{fig:gpt4o-beeswarm-mt}
    \end{figure}
    
    \begin{figure}[H]
        \centering
        \includegraphics[width=0.8\linewidth]{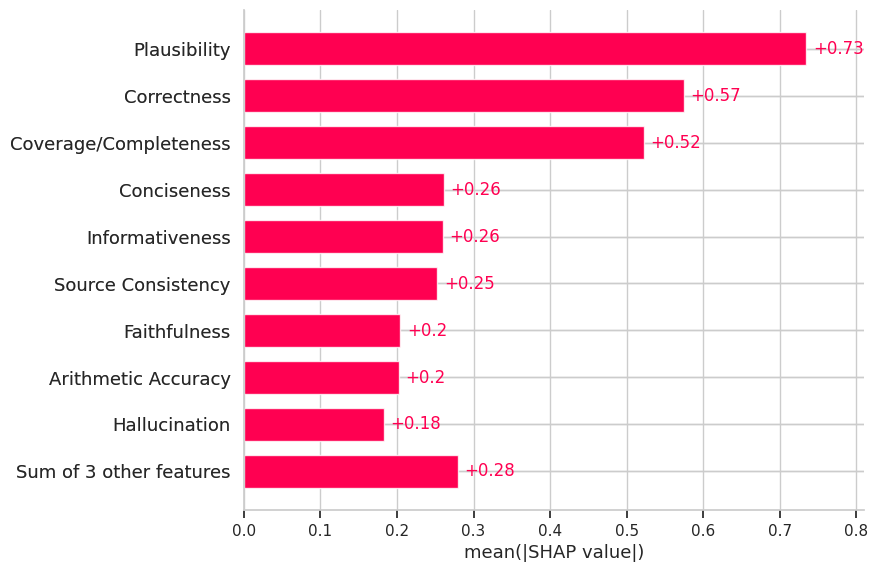}
        \caption{Mean absolute SHAP value plot for MT Bench (GPT-4o). Shows the mean importance of each attribute in the model.}
        \label{fig:gpt4o-shap-mt}
    \end{figure}
    
    \begin{figure}[H]
        \centering
        \includegraphics[width=0.9\linewidth]{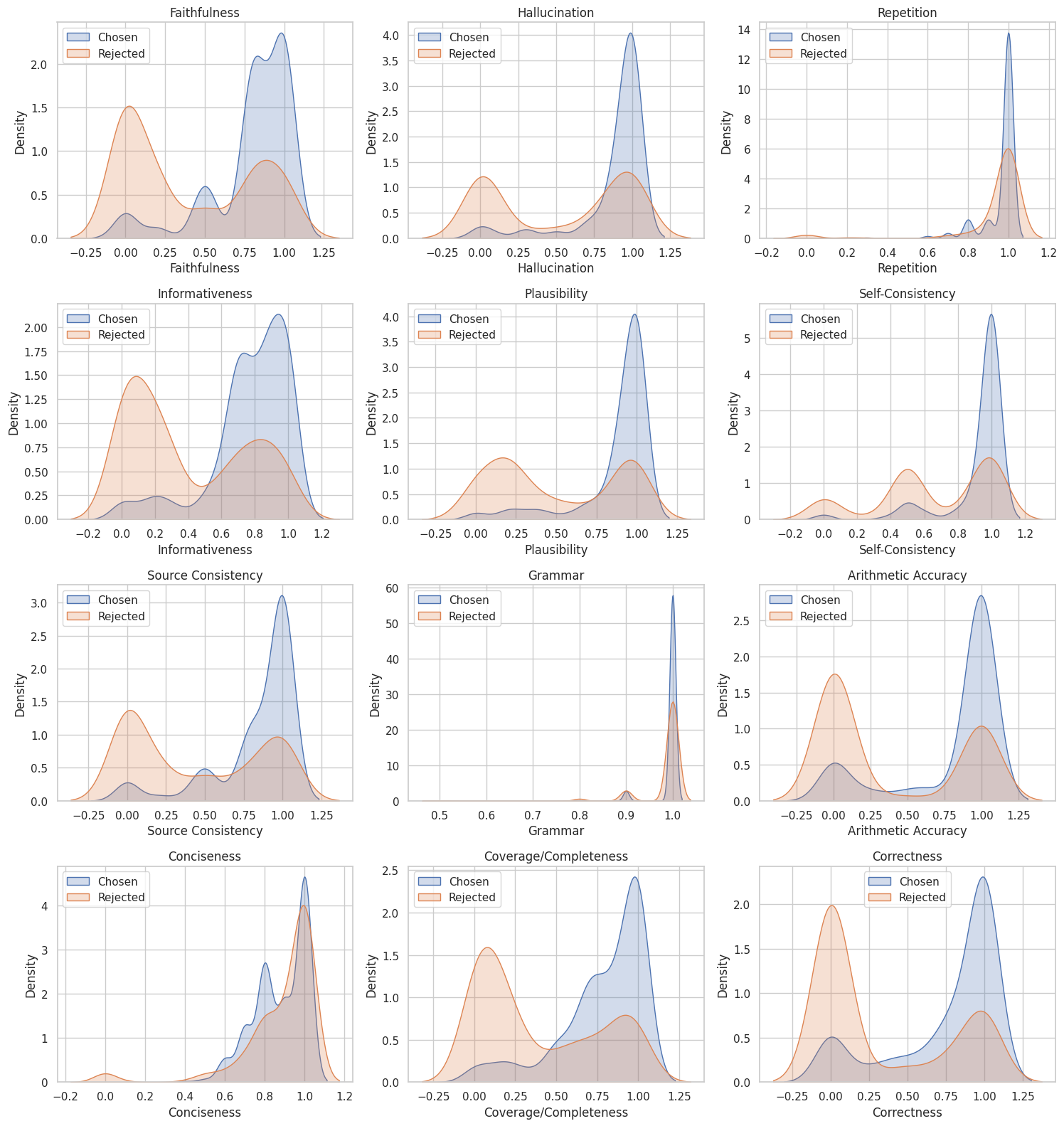}
        \caption{Distribution of attribute values for chosen vs.\ rejected rationales in MT Bench (GPT-4o). Each subplot shows the density of scores for each attribute.}
        \label{fig:gpt4o-density-mt}
    \end{figure}
\end{enumerate}

\newpage
\subsubsection{Google Gemini-2.5-Flash}
\begin{enumerate}
    \item \textbf{Chatbot Arena}
    \begin{figure}[H]
        \centering
        \includegraphics[width=0.8\linewidth]{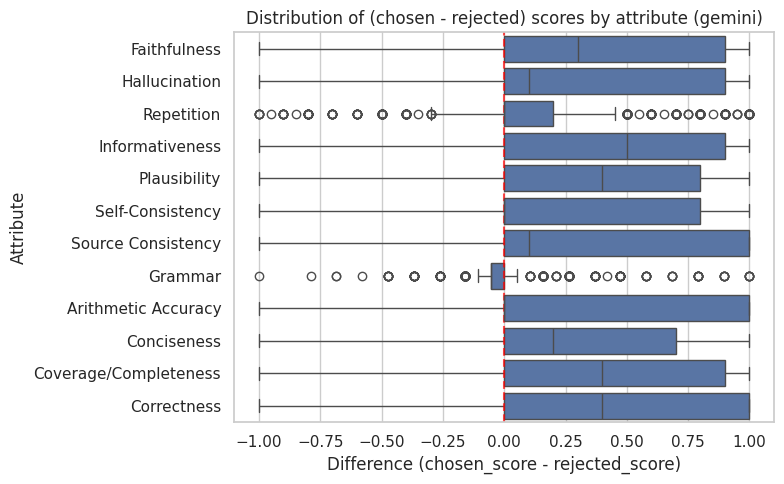}
        \caption{Distribution of the difference between chosen and rejected scores by attribute in Chatbot Arena (Gemini 2.5-Flash). Boxplots summarize the (chosen -- rejected) difference for each attribute.}
        \label{fig:gemini-diff-chat}
    \end{figure}
    
    \begin{figure}[H]
        \centering
        \includegraphics[width=0.8\linewidth]{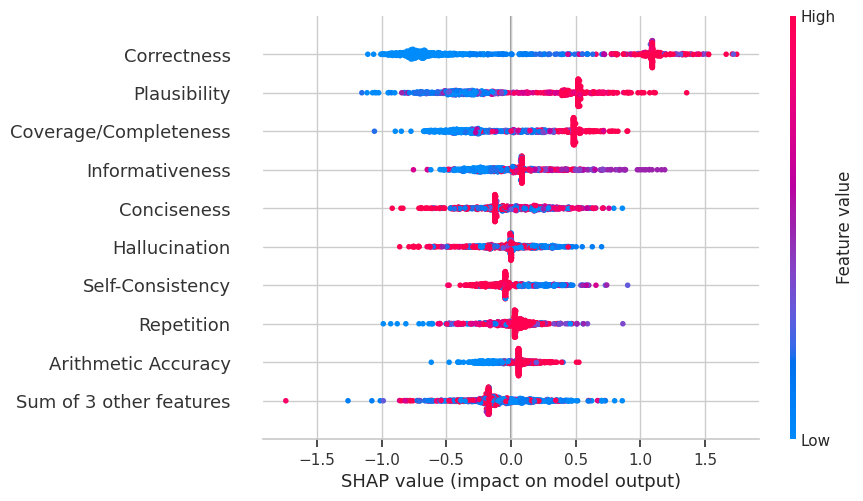}
        \caption{SHAP beeswarm plot for Chatbot Arena (Gemini 2.5-Flash). Visualizes the distribution and direction of SHAP values for each attribute.}
        \label{fig:gemini-beeswarm-chat}
    \end{figure}
    
    \begin{figure}[H]
        \centering
        \includegraphics[width=0.8\linewidth]{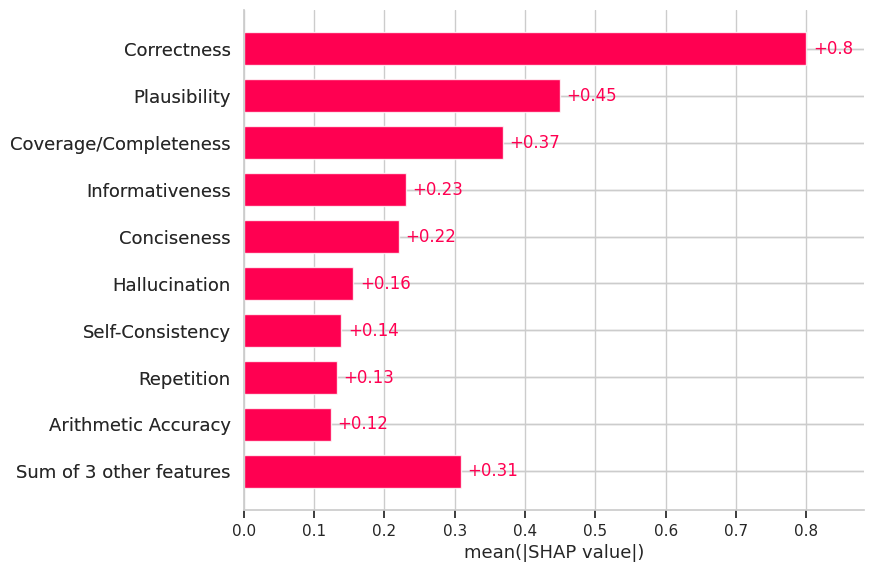}
        \caption{Mean absolute SHAP value plot for Chatbot Arena (Gemini 2.5-Flash). Shows the mean importance of each attribute in the model.}
        \label{fig:gemini-shap-chat}
    \end{figure}
    
    \begin{figure}[H]
        \centering
        \includegraphics[width=0.9\linewidth]{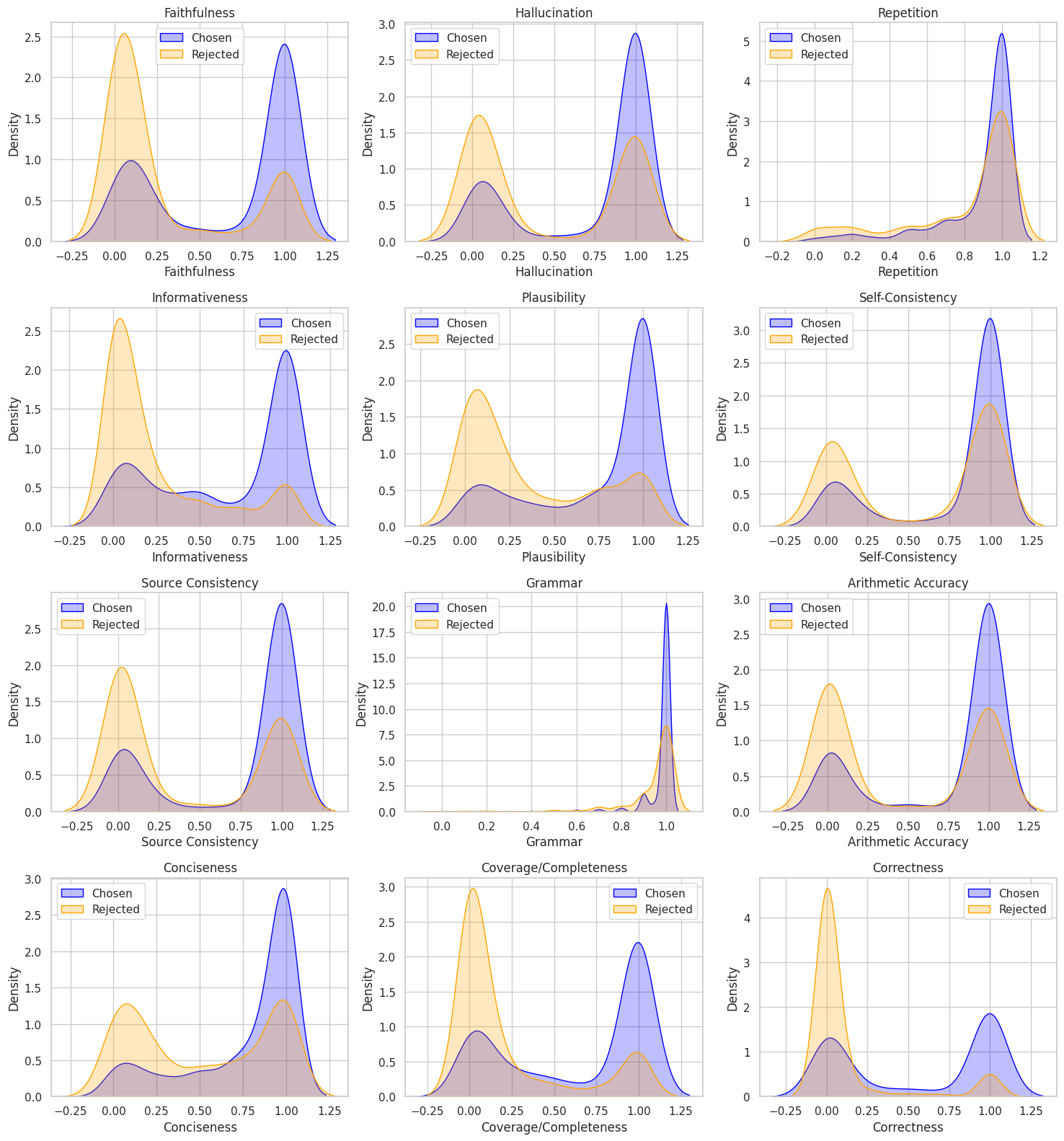}
        \caption{Distribution of attribute values for chosen vs.\ rejected rationales in Chatbot Arena (Gemini 2.5-Flash). Each subplot shows the density of scores for each attribute.}
        \label{fig:gemini-density-chat}
    \end{figure}

\newpage
    \item \textbf{Mt Bench}
\begin{figure}[H]
    \centering
    \includegraphics[width=0.8\linewidth]{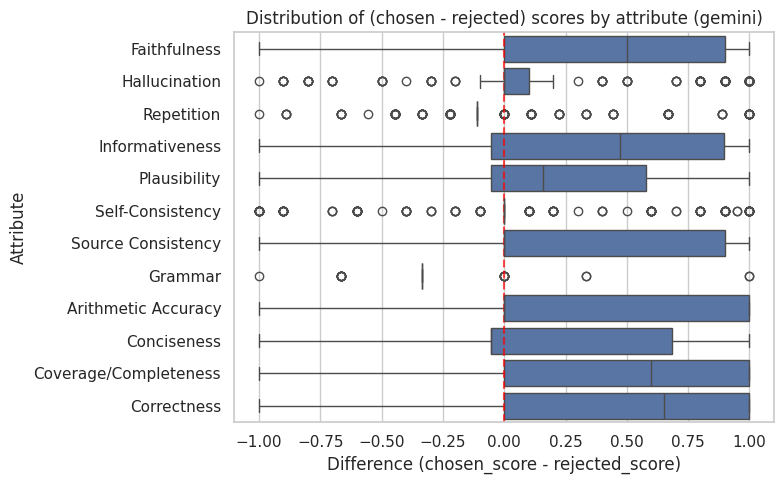}
    \caption{Distribution of the difference between chosen and rejected scores by attribute in MT Bench (Gemini 2.5-Flash). Boxplots summarize the (chosen -- rejected) difference for each attribute.}
    \label{fig:gemini-diff-mt}
\end{figure}

\begin{figure}[H]
    \centering
    \includegraphics[width=0.8\linewidth]{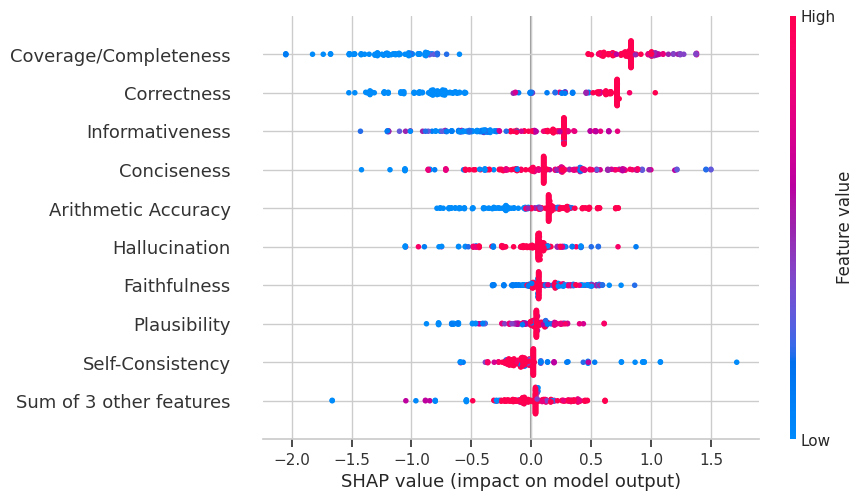}
    \caption{SHAP beeswarm plot for MT Bench (Gemini 2.5-Flash). Visualizes the distribution and direction of SHAP values for each attribute.}
    \label{fig:gemini-beeswarm-mt}
\end{figure}

\begin{figure}[H]
    \centering
    \includegraphics[width=0.8\linewidth]{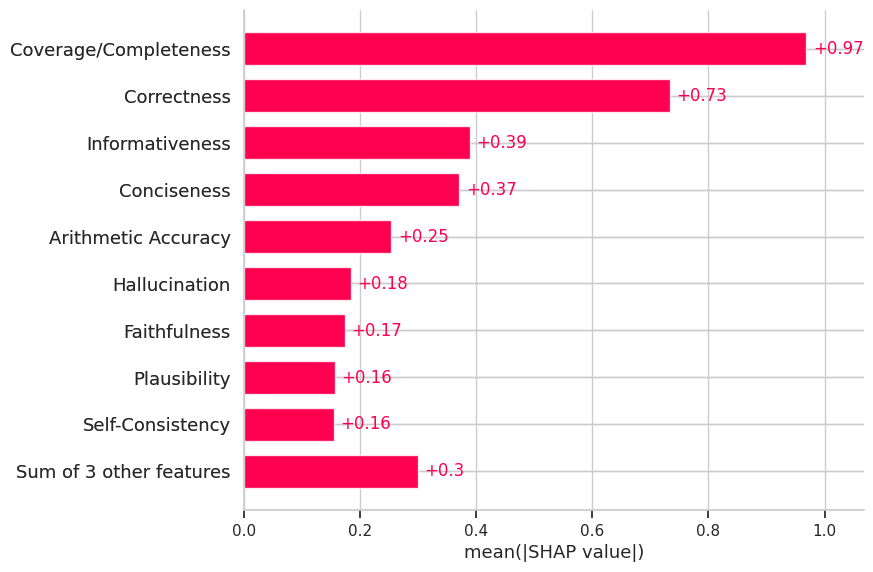}
    \caption{Mean absolute SHAP value plot for MT Bench (Gemini 2.5-Flash). Shows the mean importance of each attribute in the model.}
    \label{fig:gemini-shap-mt}
\end{figure}

\begin{figure}[H]
    \centering
    \includegraphics[width=0.9\linewidth]{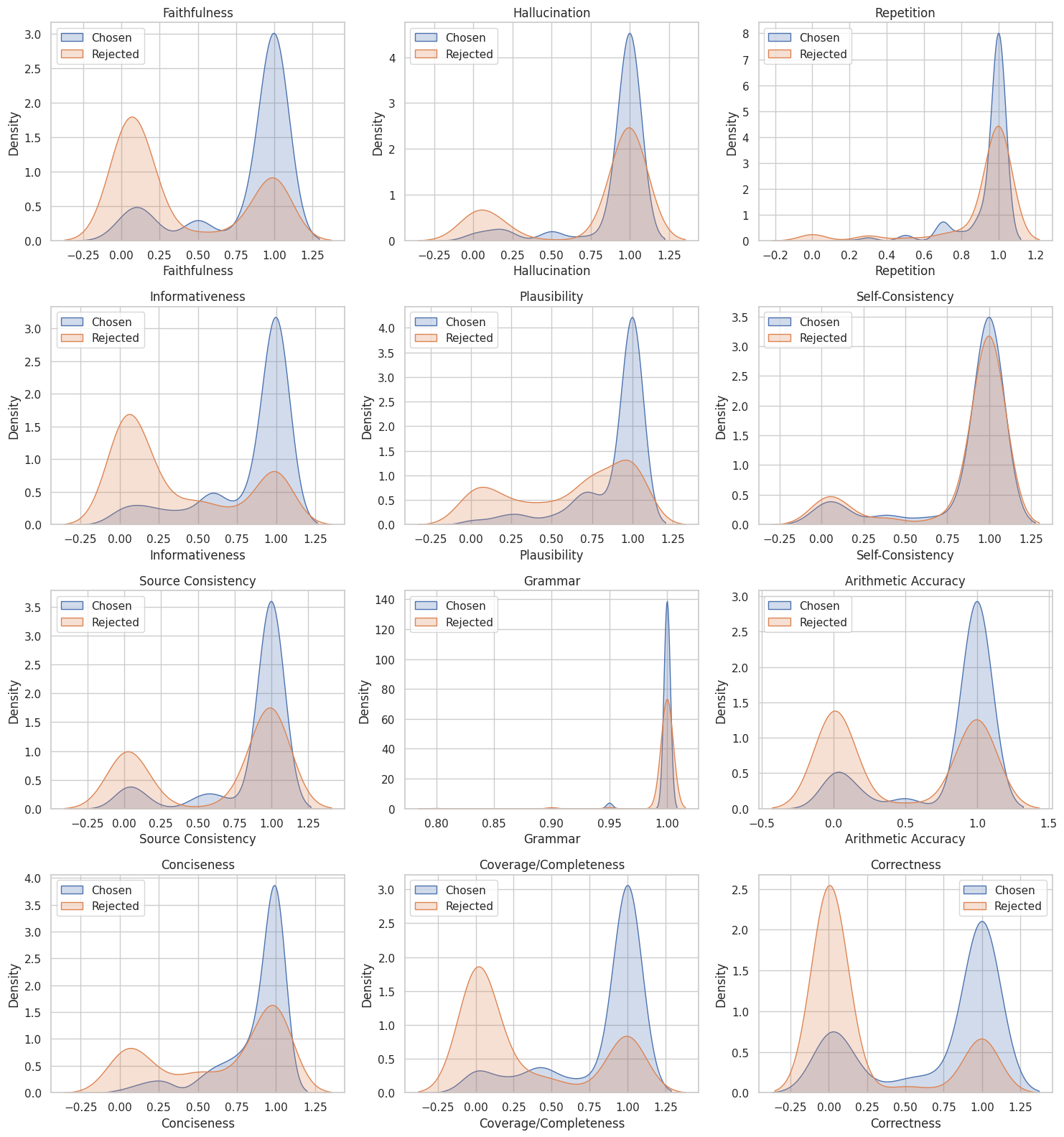}
    \caption{Distribution of attribute values for chosen vs.\ rejected rationales in MT Bench (Gemini 2.5-Flash). Each subplot shows the density of scores for each attribute.}
    \label{fig:gemini-density-mt}
\end{figure}

\end{enumerate}

\newpage

\subsubsection{OLMo 2-32b}
\begin{enumerate}
    \item \textbf{Chatbot Arena}
    \begin{figure}[H]
    \centering
    \includegraphics[width=0.8\linewidth]{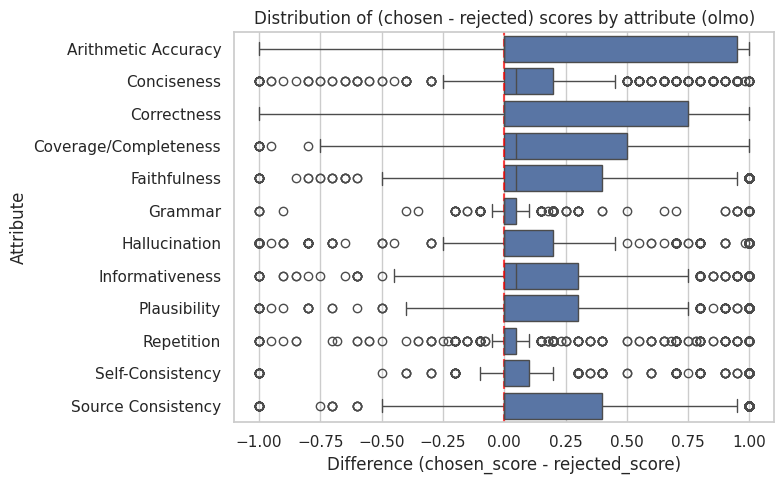}
    \caption{Distribution of the difference between chosen and rejected scores by attribute in Chatbot Arena (OLMo 2-32b). Boxplots summarize the (chosen -- rejected) difference for each attribute.}
    \label{fig:olmo-diff-chat}
\end{figure}

\begin{figure}[H]
    \centering
    \includegraphics[width=0.8\linewidth]{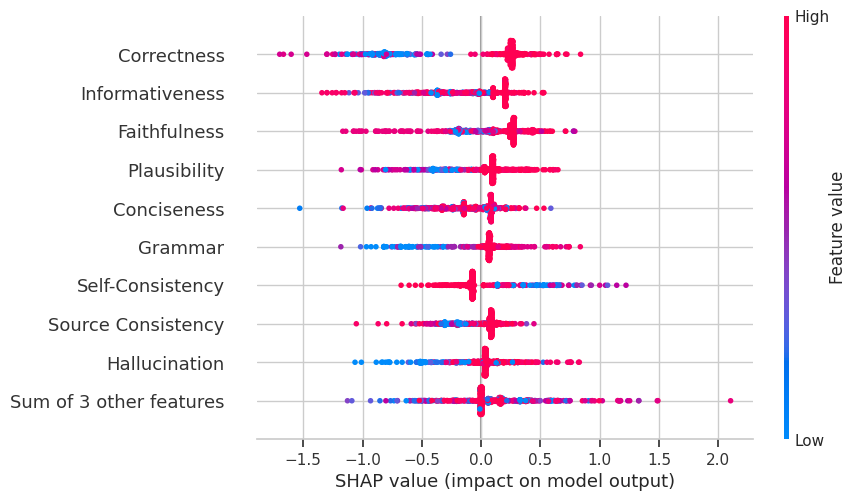}
    \caption{SHAP beeswarm plot for Chatbot Arena (OLMo 2-32b). Visualizes the distribution and direction of SHAP values for each attribute.}
    \label{fig:olmo-beeswarm-chat}
\end{figure}

\begin{figure}[H]
    \centering
    \includegraphics[width=0.8\linewidth]{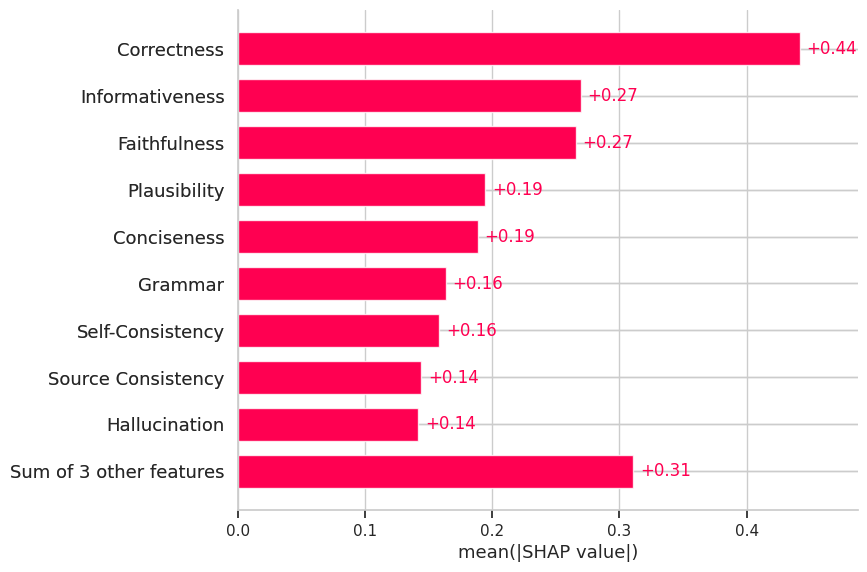}
    \caption{Mean absolute SHAP value plot for Chatbot Arena (OLMo 2-32b). Shows the mean importance of each attribute in the model.}
    \label{fig:olmo-shap-chat}
\end{figure}

\begin{figure}[H]
    \centering
    \includegraphics[width=0.9\linewidth]{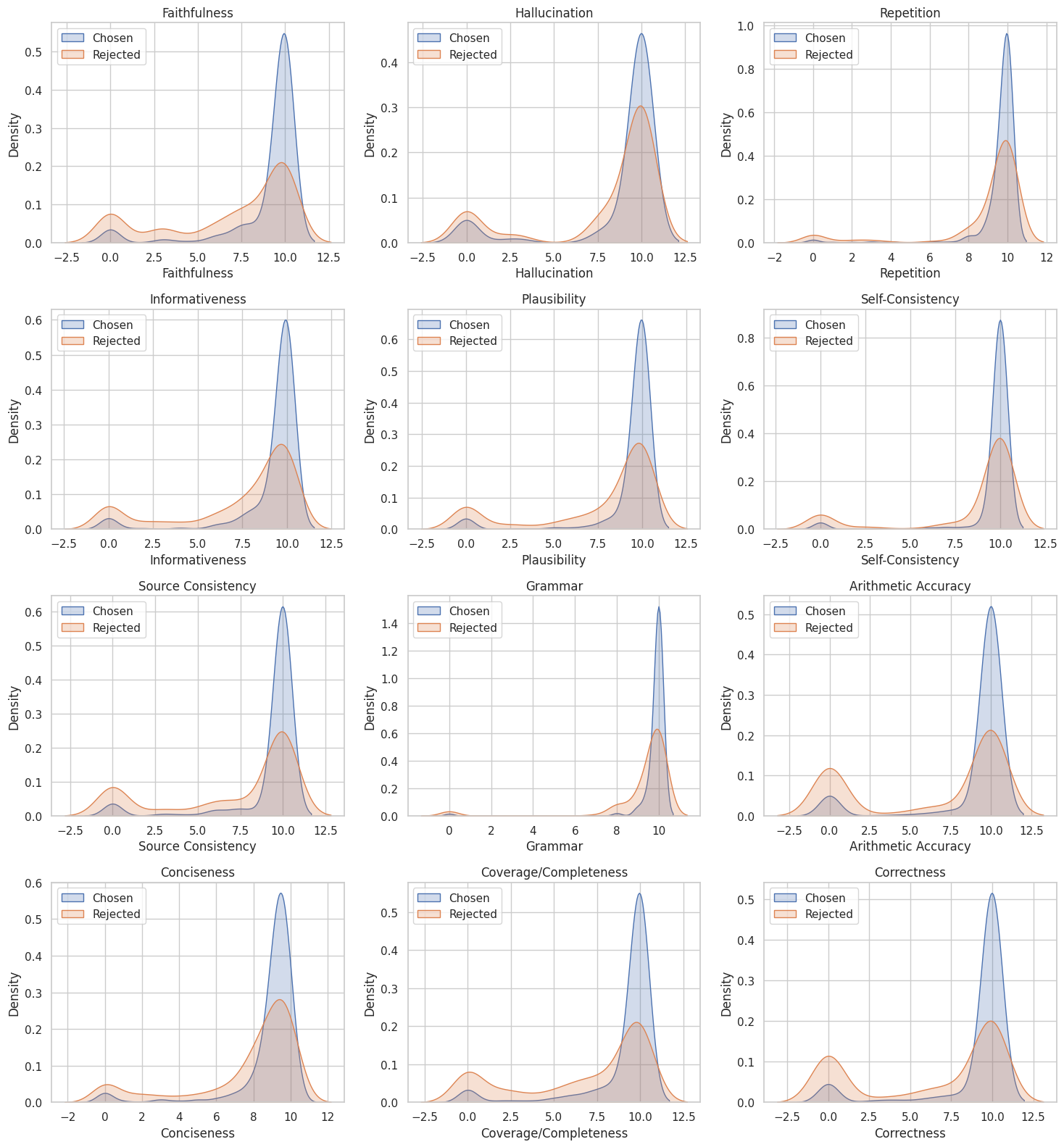}
    \caption{Distribution of attribute values for chosen vs.\ rejected rationales in Chatbot Arena (OLMo 2-32b). Each subplot shows the density of scores for each attribute.}
    \label{fig:olmo-density-chat}
\end{figure}

\newpage
    \item \textbf{Mt Bench}

    \begin{figure}[H]
    \centering
    \includegraphics[width=0.8\linewidth]{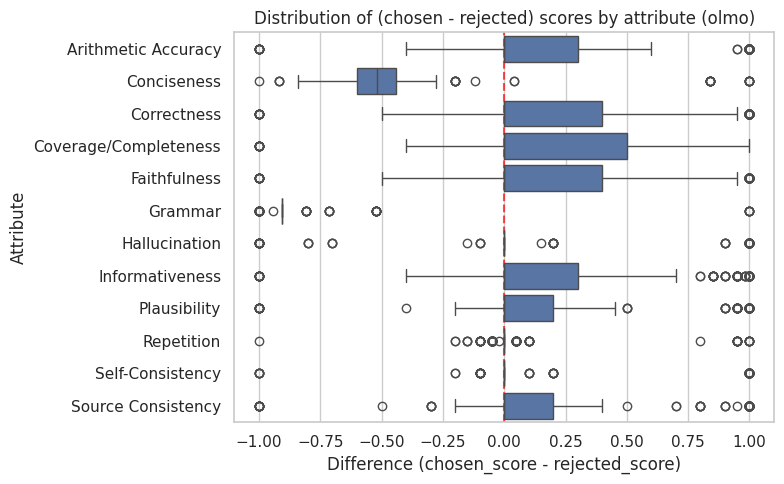}
    \caption{Distribution of the difference between chosen and rejected scores by attribute in MT Bench (OLMo 2-32b). Boxplots summarize the (chosen -- rejected) difference for each attribute.}
    \label{fig:olmo-diff-mt}
    \end{figure}
    
    \begin{figure}[H]
        \centering
        \includegraphics[width=0.8\linewidth]{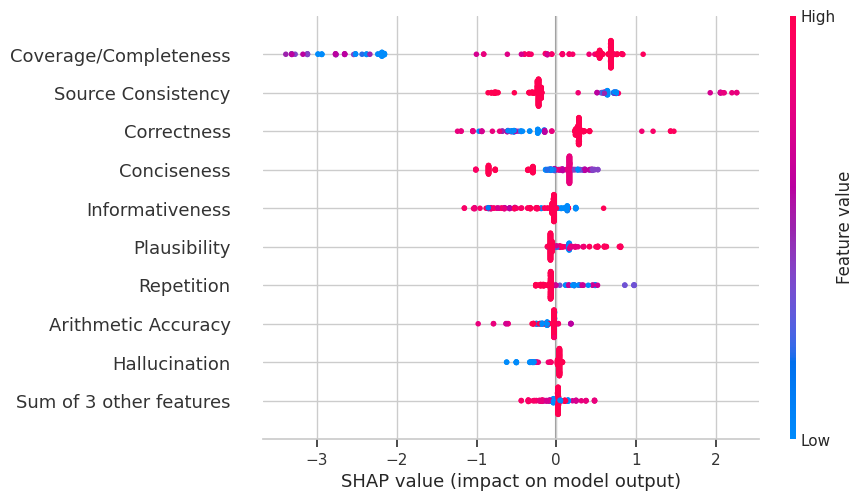}
        \caption{SHAP beeswarm plot for MT Bench (OLMo 2-32b). Visualizes the distribution and direction of SHAP values for each attribute.}
        \label{fig:olmo-beeswarm-mt}
    \end{figure}
    
    \begin{figure}[H]
        \centering
        \includegraphics[width=0.8\linewidth]{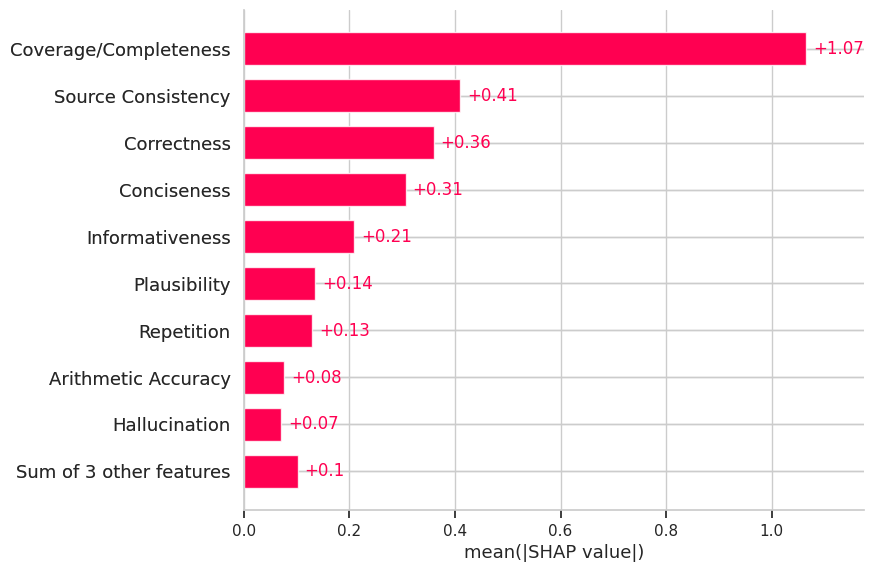}
        \caption{Mean absolute SHAP value plot for MT Bench (OLMo 2-32b). Shows the mean importance of each attribute in the model.}
        \label{fig:olmo-shap-mt}
    \end{figure}
    
    \begin{figure}[H]
        \centering
        \includegraphics[width=0.9\linewidth]{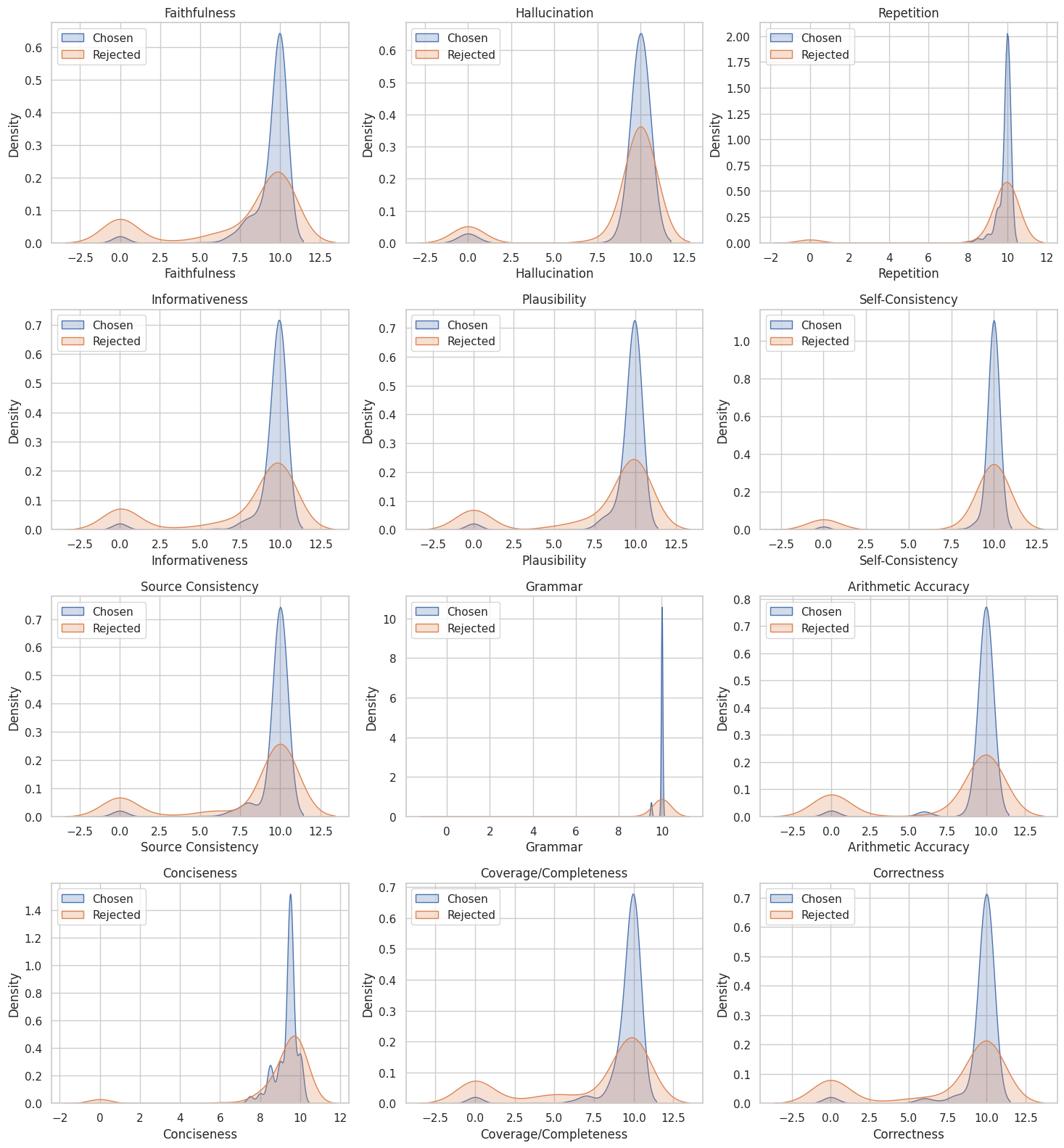}
        \caption{Distribution of attribute values for chosen vs.\ rejected rationales in MT Bench (OLMo 2-32b). Each subplot shows the density of scores for each attribute.}
        \label{fig:olmo-density-mt}
    \end{figure}
\end{enumerate}

\newpage
\subsection{Human Annotators Results} \label{human-results-appendix}
The human evaluation results reported in this section are based on the average scores assigned by three independent annotators for each question. For every rationale, we take the mean of the three annotators' scores to obtain a single human score per attribute. This approach helps mitigate individual annotator bias and provides a more robust measure of human preference.

\subsubsection{Chatbot Arena}
\begin{figure}[H]
    \centering
    \includegraphics[width=0.8\linewidth]{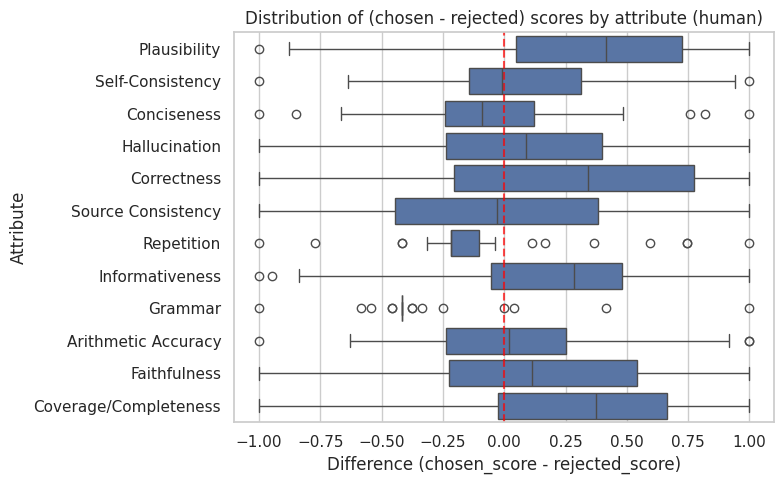}
    \caption{Distribution of the difference between chosen and rejected scores by attribute in Chatbot Arena (Human Annotators). Boxplots summarize the (chosen -- rejected) difference for each attribute.}
    \label{fig:human-diff-chat}
\end{figure}

\begin{figure}[H]
    \centering
    \includegraphics[width=0.8\linewidth]{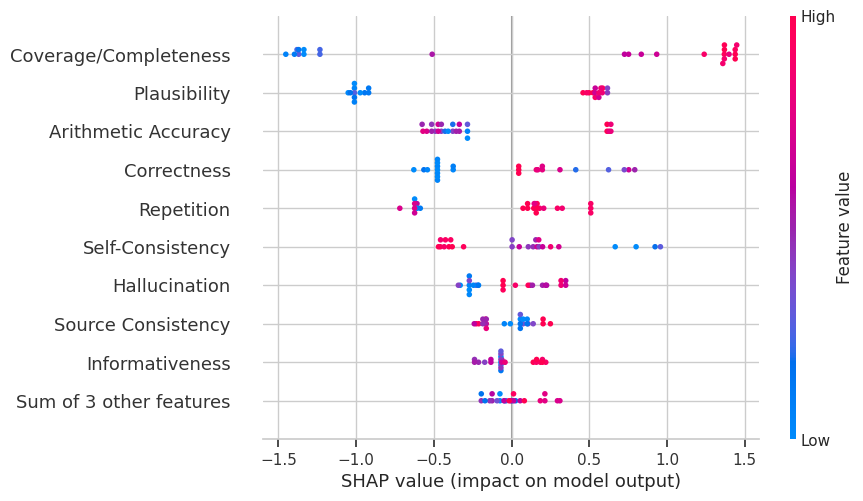}
    \caption{SHAP beeswarm plot for Chatbot Arena (Human Annotators). Visualizes the distribution and direction of SHAP values for each attribute.}
    \label{fig:human-beeswarm-chat}
\end{figure}

\begin{figure}[H]
    \centering
    \includegraphics[width=0.8\linewidth]{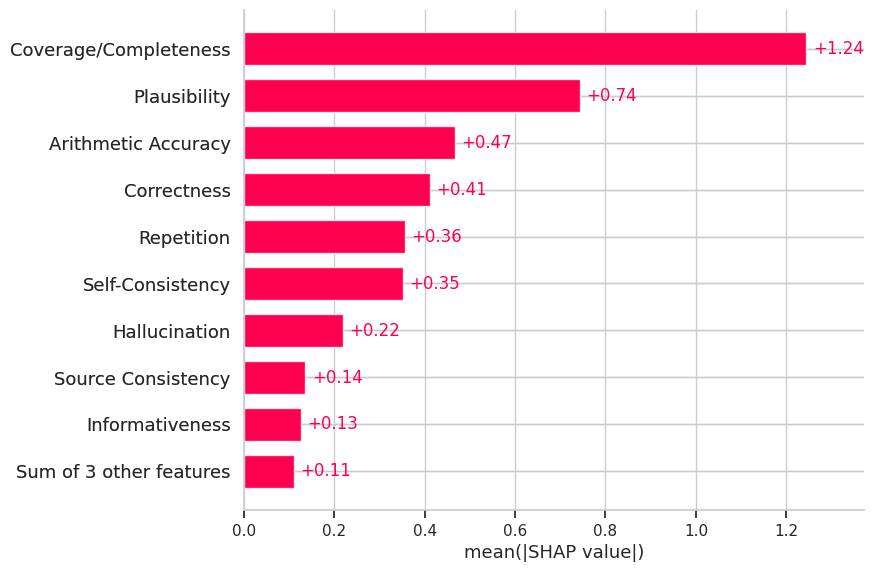}
    \caption{Mean absolute SHAP value plot for Chatbot Arena (Human Annotators). Shows the mean importance of each attribute in the model.}
    \label{fig:human-shap-chat}
\end{figure}

\begin{figure}[H]
    \centering
    \includegraphics[width=0.9\linewidth]{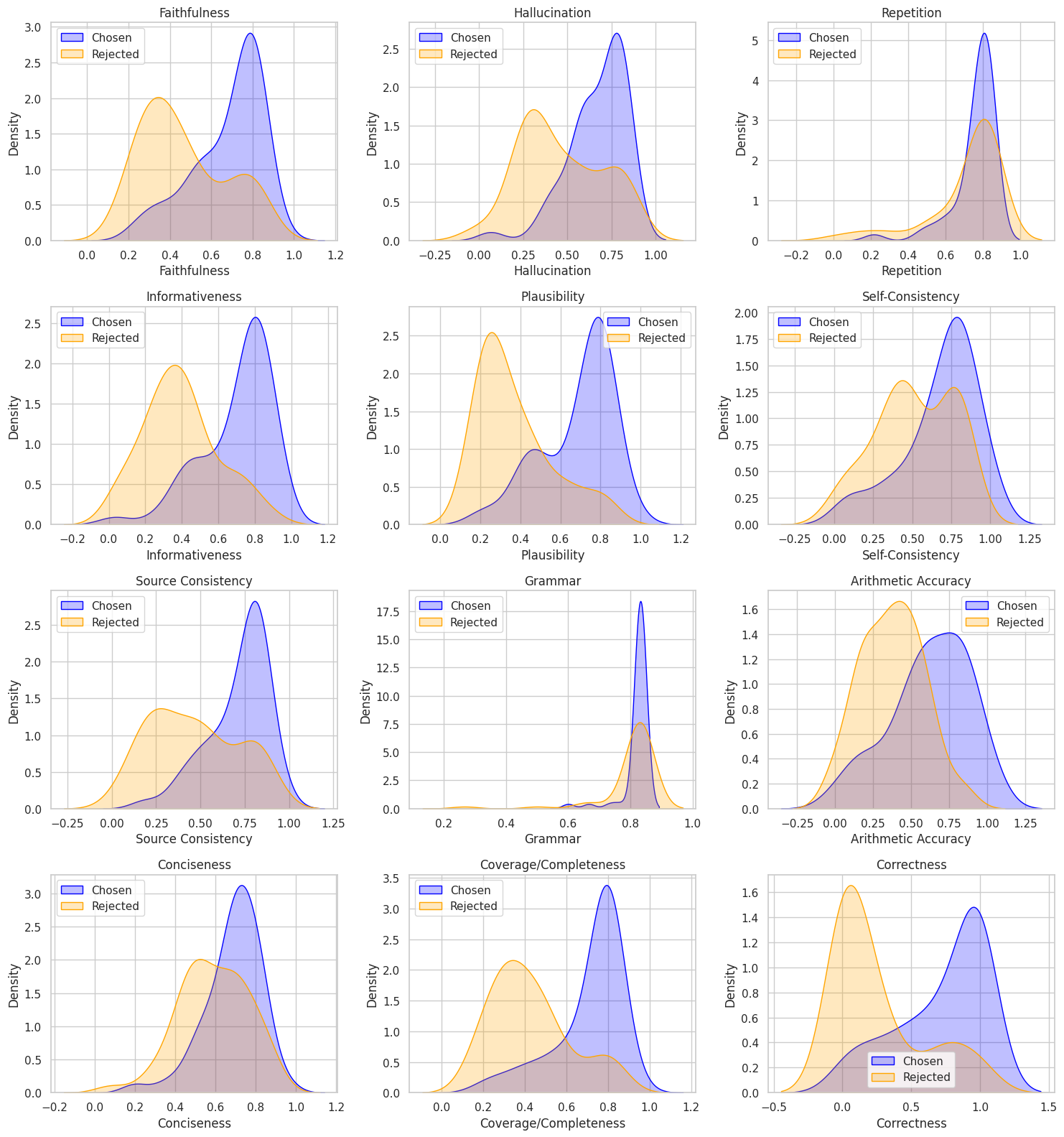}
    \caption{Distribution of attribute values for chosen vs.\ rejected rationales in Chatbot Arena (Human Annotators). Each subplot shows the density of scores for each attribute.}
    \label{fig:human-density-chat}
\end{figure}

\newpage
\subsubsection{Mt Bench}
\begin{figure}[H]
    \centering
    \includegraphics[width=0.8\linewidth]{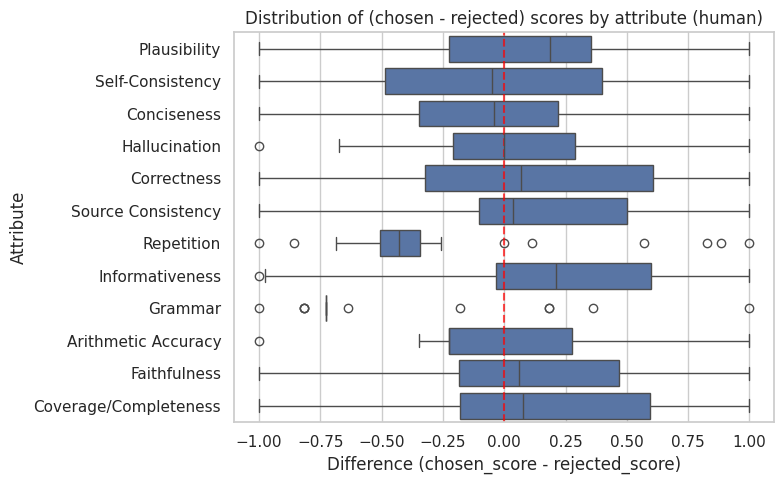}
    \caption{Distribution of the difference between chosen and rejected scores by attribute in MT Bench (Human Annotators). Boxplots summarize the (chosen -- rejected) difference for each attribute.}
    \label{fig:human-diff-mt}
\end{figure}

\begin{figure}[H]
    \centering
    \includegraphics[width=0.8\linewidth]{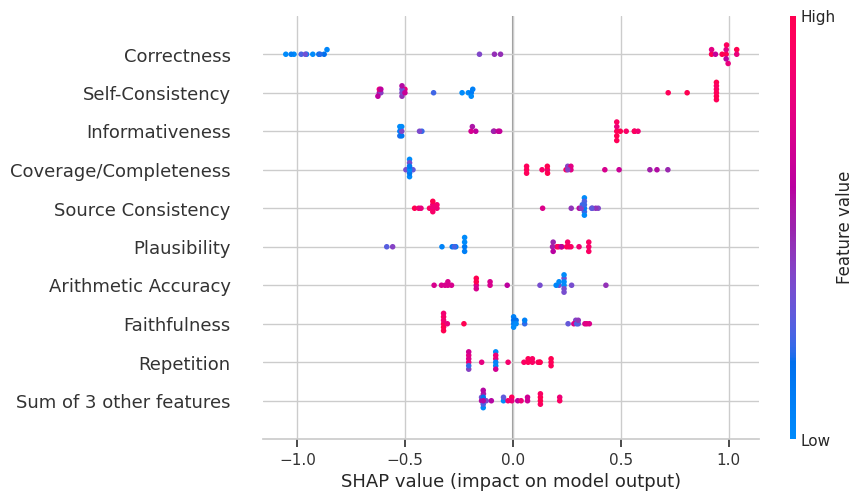}
    \caption{SHAP beeswarm plot for MT Bench (Human Annotators). Visualizes the distribution and direction of SHAP values for each attribute.}
    \label{fig:human-beeswarm-mt}
\end{figure}

\begin{figure}[H]
    \centering
    \includegraphics[width=0.8\linewidth]{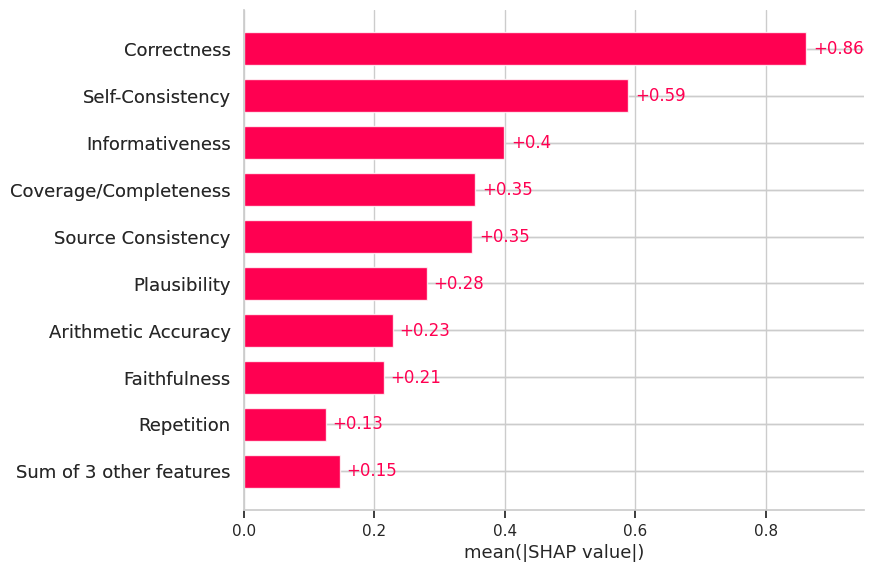}
    \caption{Mean absolute SHAP value plot for MT Bench (Human Annotators). Shows the mean importance of each attribute in the model.}
    \label{fig:human-shap-mt}
\end{figure}

\begin{figure}[H]
    \centering
    \includegraphics[width=0.9\linewidth]{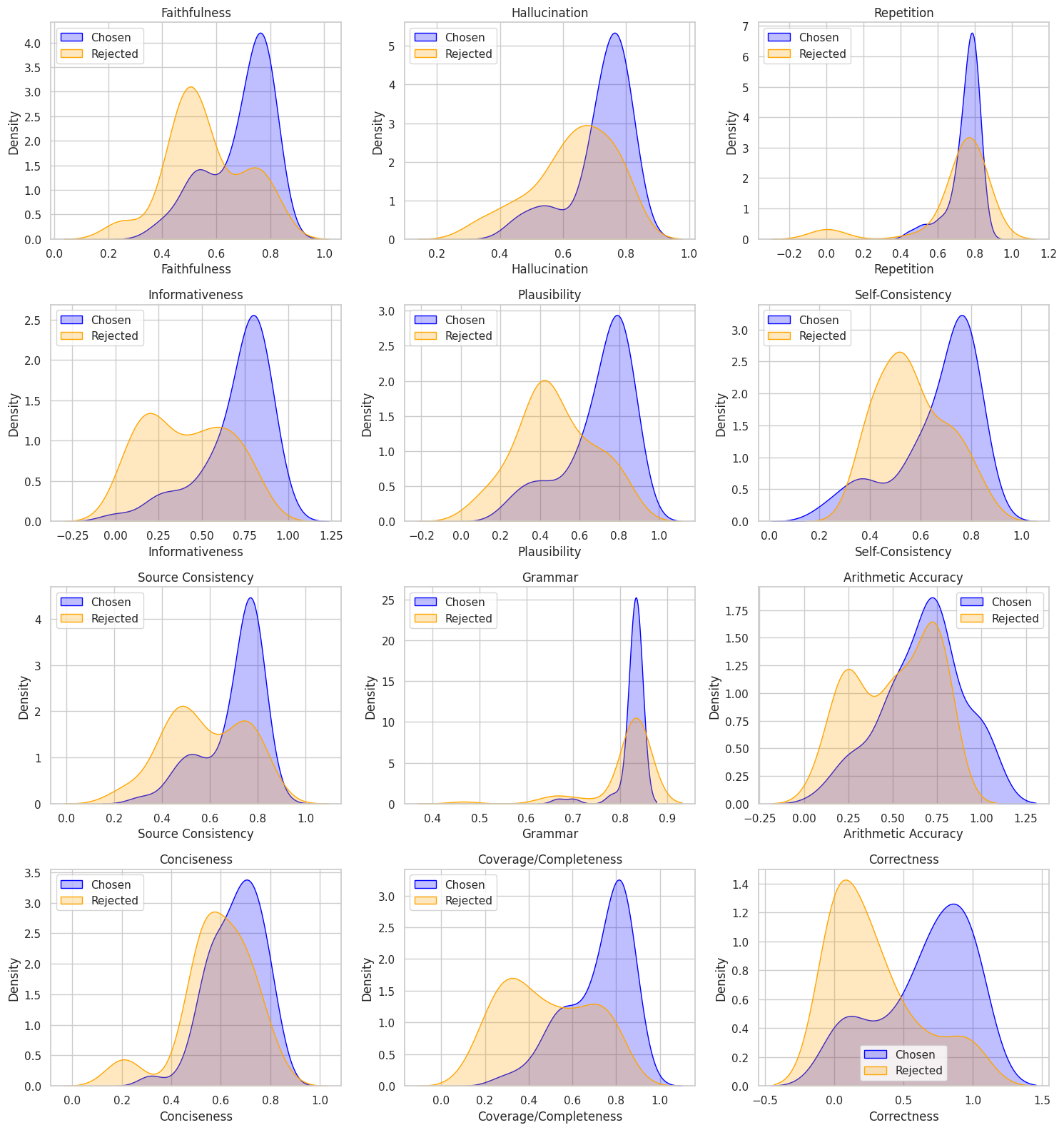}
    \caption{Distribution of attribute values for chosen vs.\ rejected rationales in MT Bench (Human Annotators). Each subplot shows the density of scores for each attribute.}
    \label{fig:human-density-mt}
\end{figure}

\newpage
\subsection{ELO ranking results} \label{elo-results-appendix}
\subsubsection{LLM Judges}
\begin{enumerate}
    \item \textbf{Chatbot Arena}
    \begin{figure}[H]
    \centering
    \includegraphics[width=0.99\textwidth]{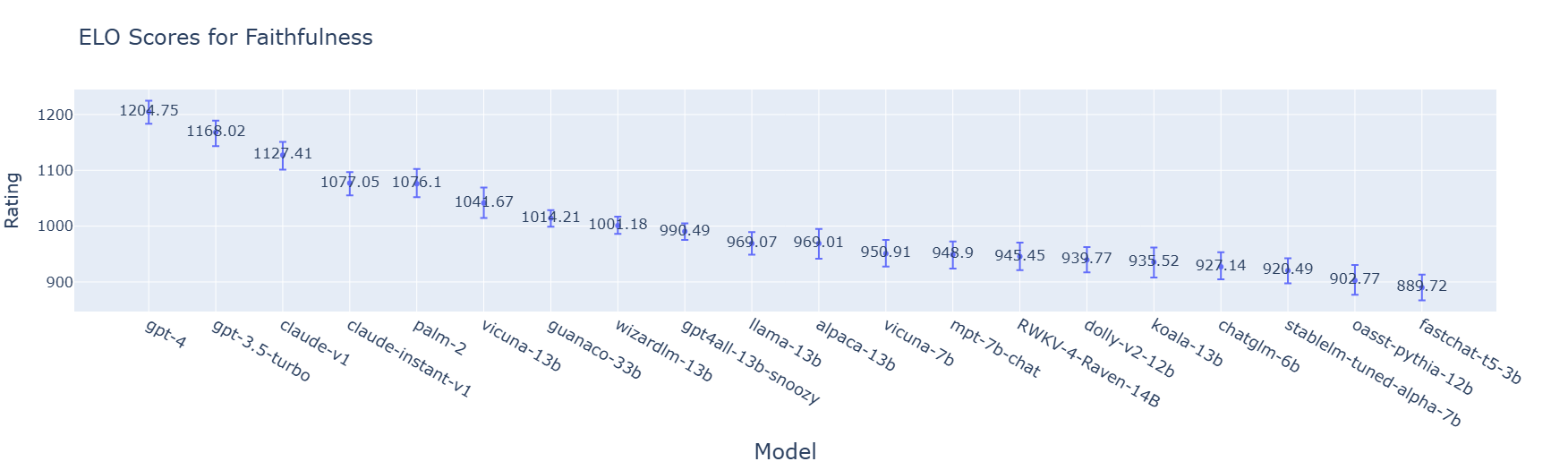}
    \caption{ELO Scores for Faithfulness across all models in Chatbot Arena, scored by the mean score of three LLMs.}
    \label{fig:elo-faithfulness-c}
\end{figure}

\begin{figure}[H]
    \centering
    \includegraphics[width=0.99\textwidth]{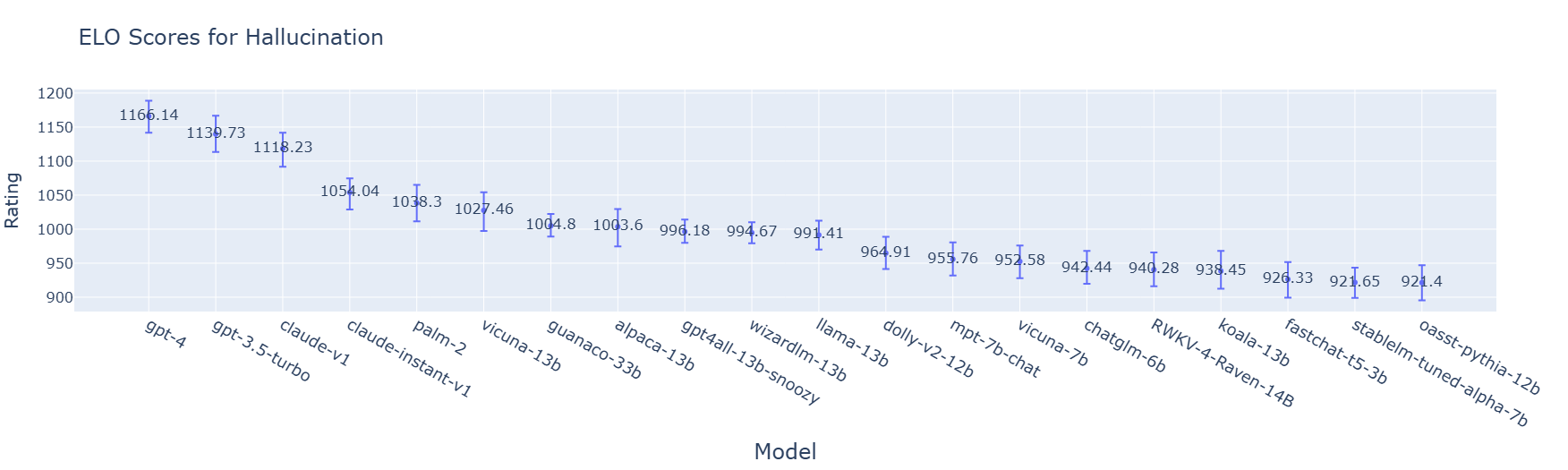}
    \caption{ELO Scores for Hallucination across all models in Chatbot Arena, scored by the mean score of three LLMs.}
    \label{fig:elo-hallucination-c}
\end{figure}

\begin{figure}[H]
    \centering
    \includegraphics[width=0.99\textwidth]{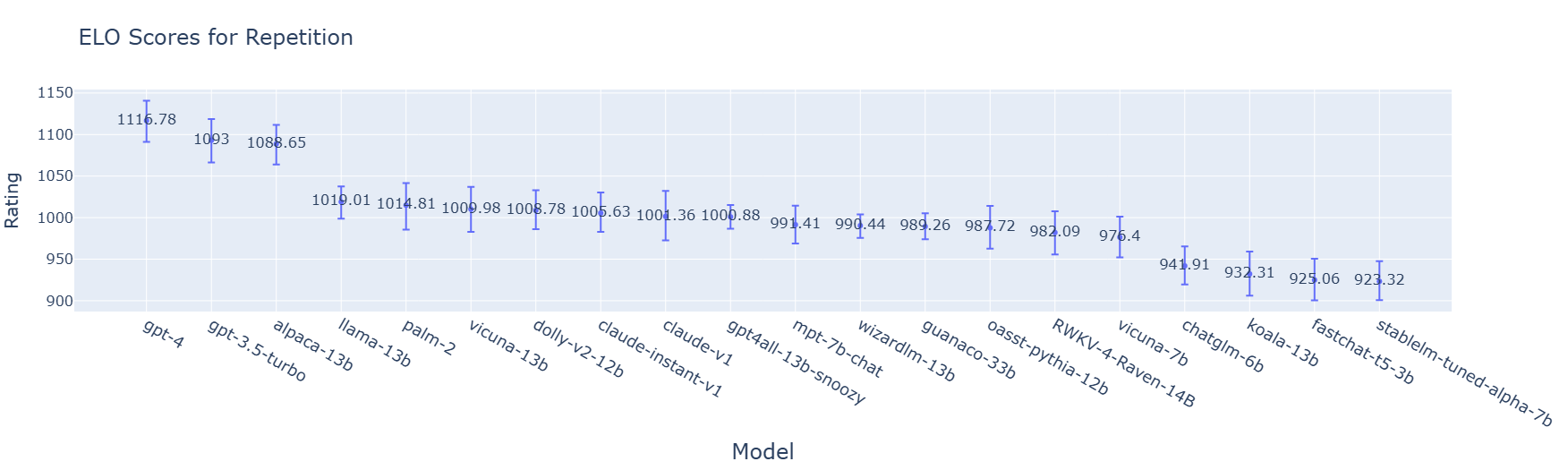}
    \caption{ELO Scores for Repetition across all models in Chatbot Arena, scored by the mean score of three LLMs.}
    \label{fig:elo-repetition-c}
\end{figure}

\begin{figure}[H]
    \centering
    \includegraphics[width=0.99\textwidth]{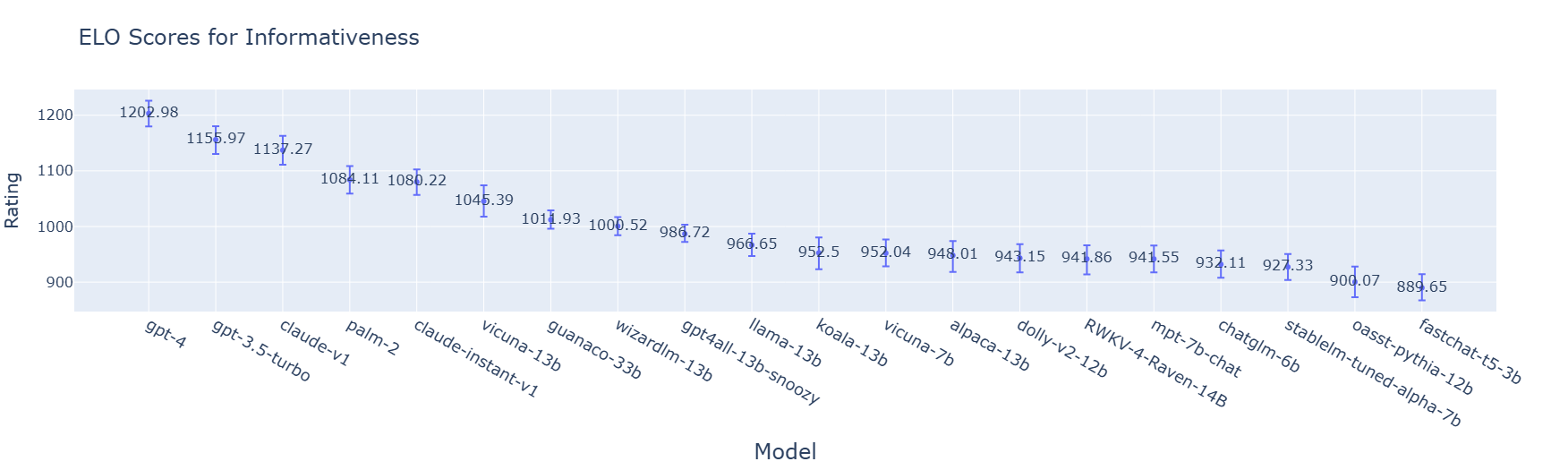}
    \caption{ELO Scores for Informativeness across all models in Chatbot Arena, scored by the mean score of three LLMs.}
    \label{fig:elo-informativeness-c}
\end{figure}

\begin{figure}[H]
    \centering
    \includegraphics[width=0.99\textwidth]{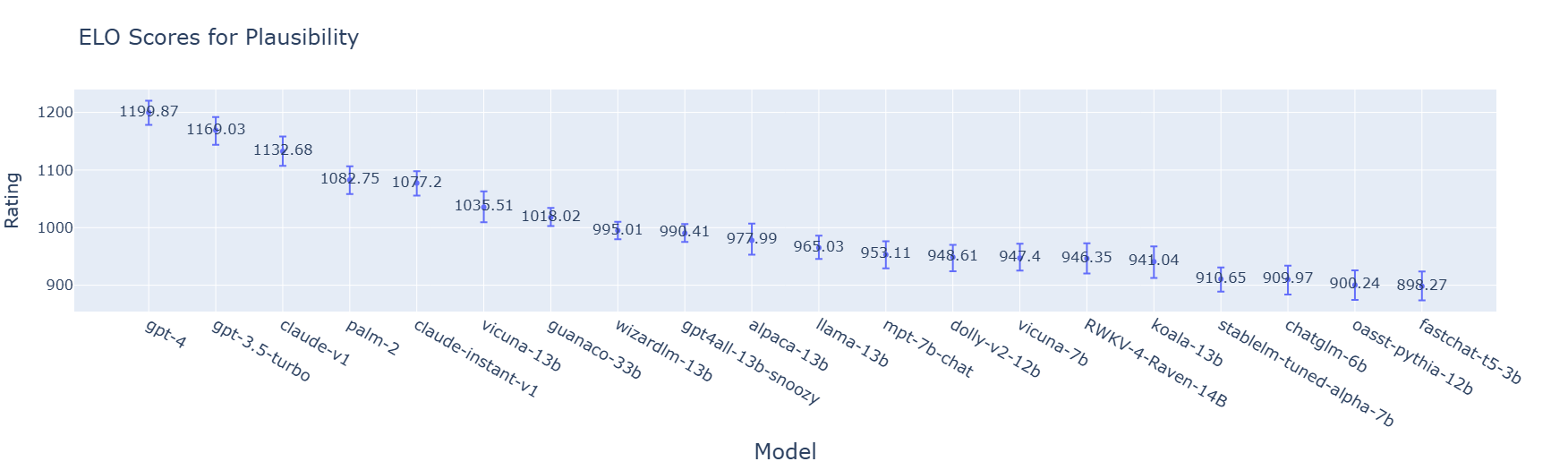}
    \caption{ELO Scores for Plausibility across all models in Chatbot Arena, scored by the mean score of three LLMs.}
    \label{fig:elo-plausibility-c}
\end{figure}

\begin{figure}[H]
    \centering
    \includegraphics[width=0.99\textwidth]{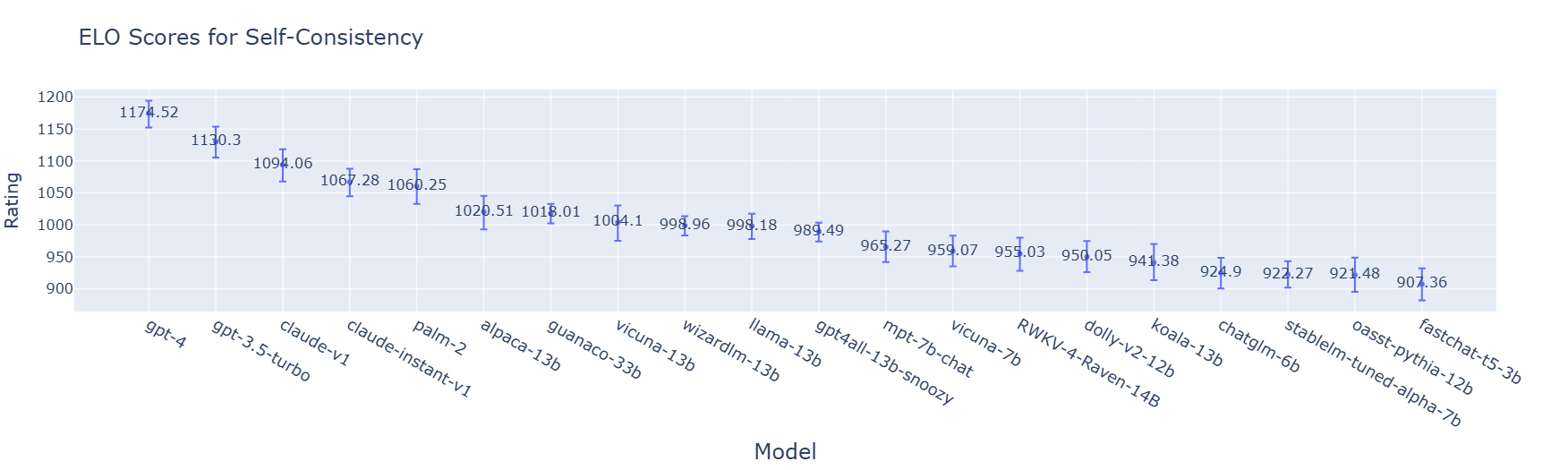}
    \caption{ELO Scores for Self-Consistency across all models in Chatbot Arena, scored by the mean score of three LLMs.}
    \label{fig:elo-selfconsistency-c}
\end{figure}

\begin{figure}[H]
    \centering
    \includegraphics[width=0.99\textwidth]{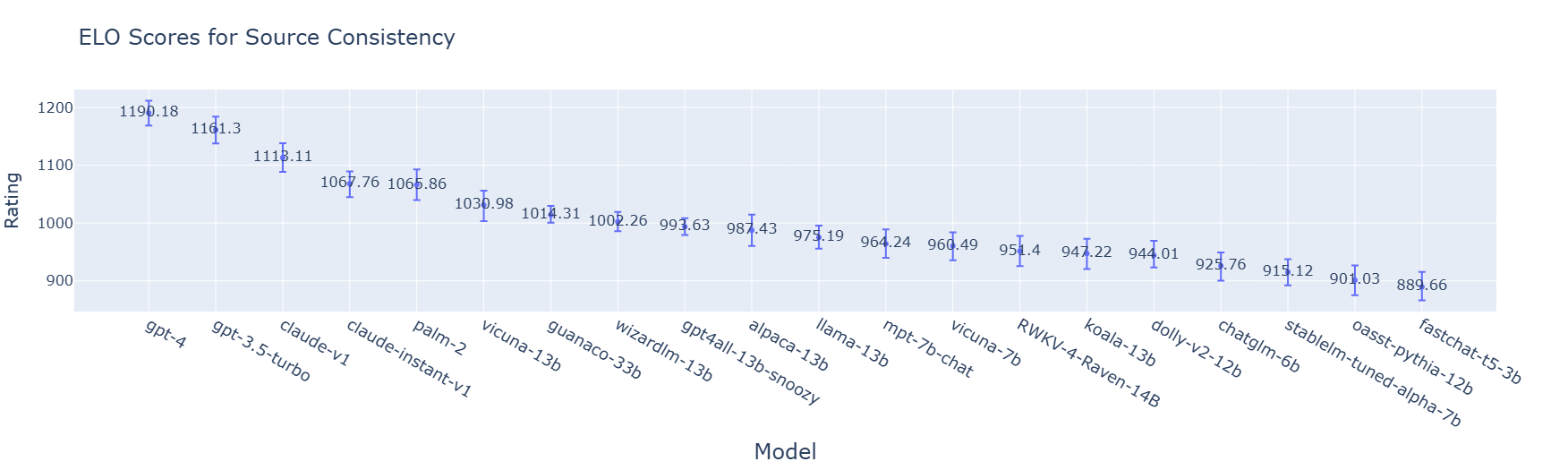}
    \caption{ELO Scores for Source Consistency across all models in Chatbot Arena, scored by the mean score of three LLMs.}
    \label{fig:elo-sourceconsistency-c}
\end{figure}

\begin{figure}[H]
    \centering
    \includegraphics[width=0.99\textwidth]{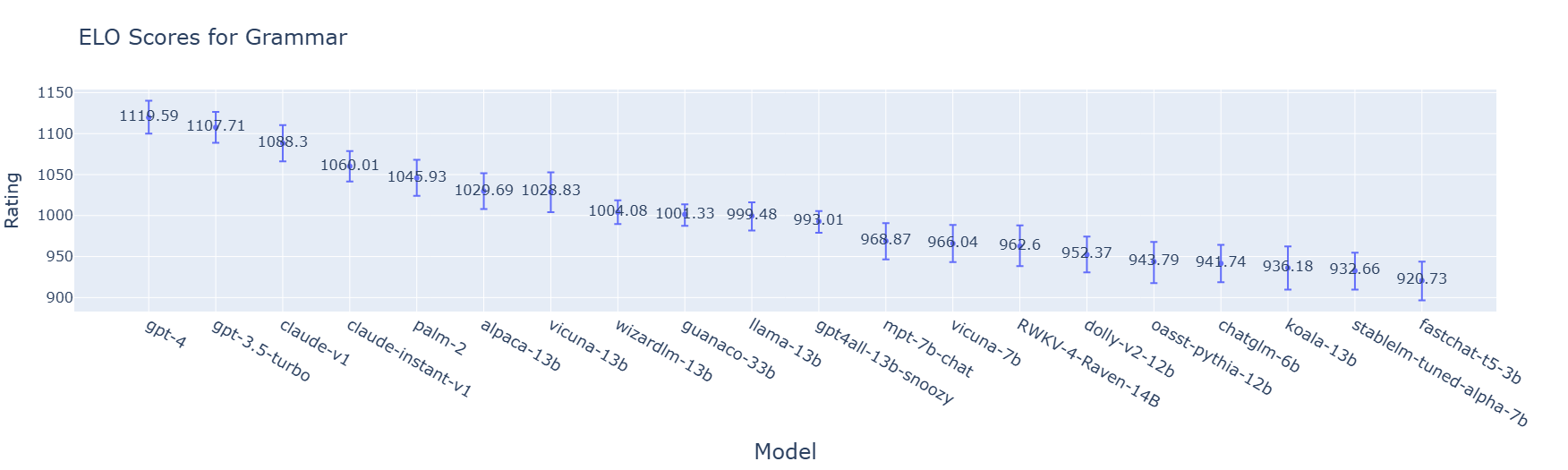}
    \caption{ELO Scores for Grammar across all models in Chatbot Arena, scored by the mean score of three LLMs.}
    \label{fig:elo-grammar-c}
\end{figure}

\begin{figure}[H]
    \centering
    \includegraphics[width=0.99\textwidth]{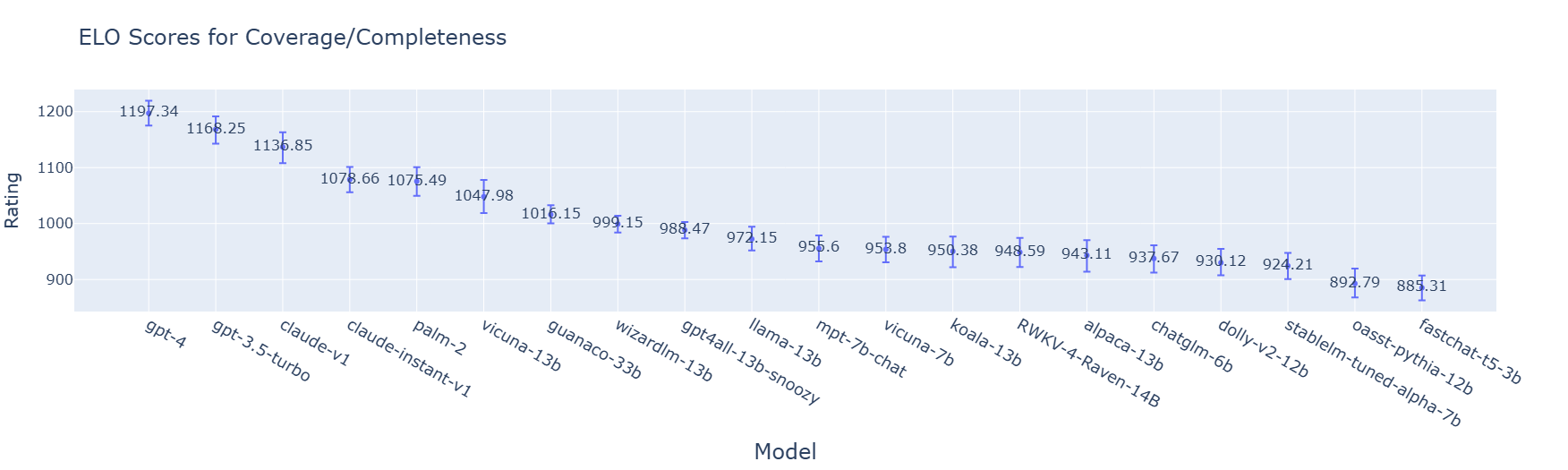}
    \caption{ELO Scores for Completeness across all models in Chatbot Arena, scored by the mean score of three LLMs.}
    \label{fig:elo-correctness-c}
\end{figure}

\begin{figure}[H]
    \centering
    \includegraphics[width=0.99\textwidth]{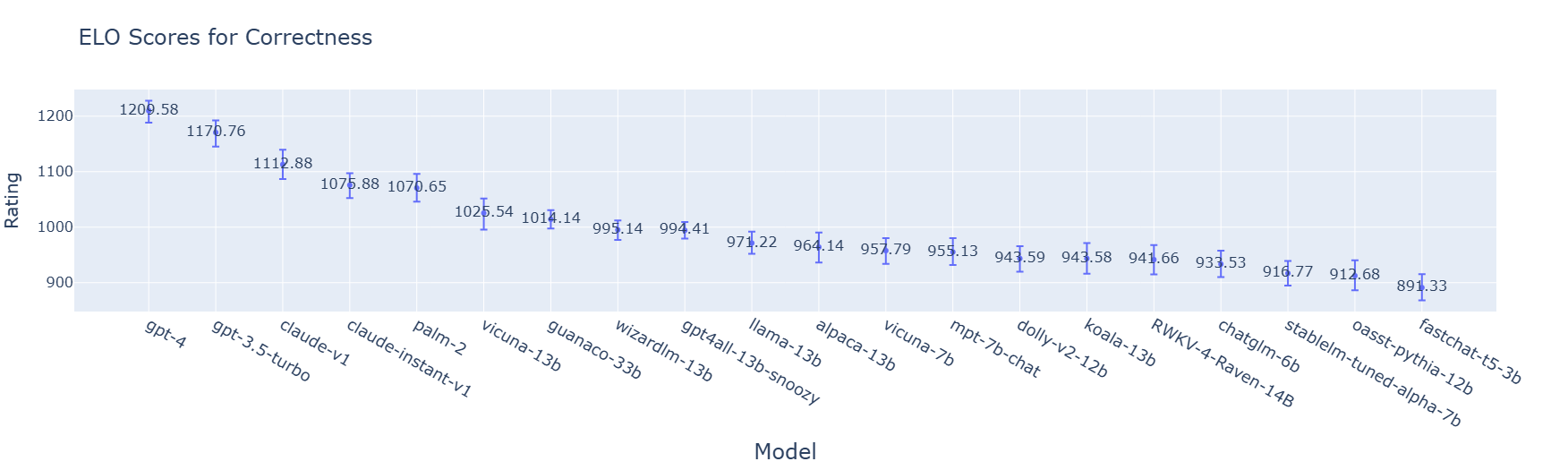}
    \caption{ELO Scores for Correctness across all models in Chatbot Arena, scored by the mean score of three LLMs.}
    \label{fig:elo-coverage-c}
\end{figure}

\begin{figure}[H]
    \centering
    \includegraphics[width=0.99\textwidth]{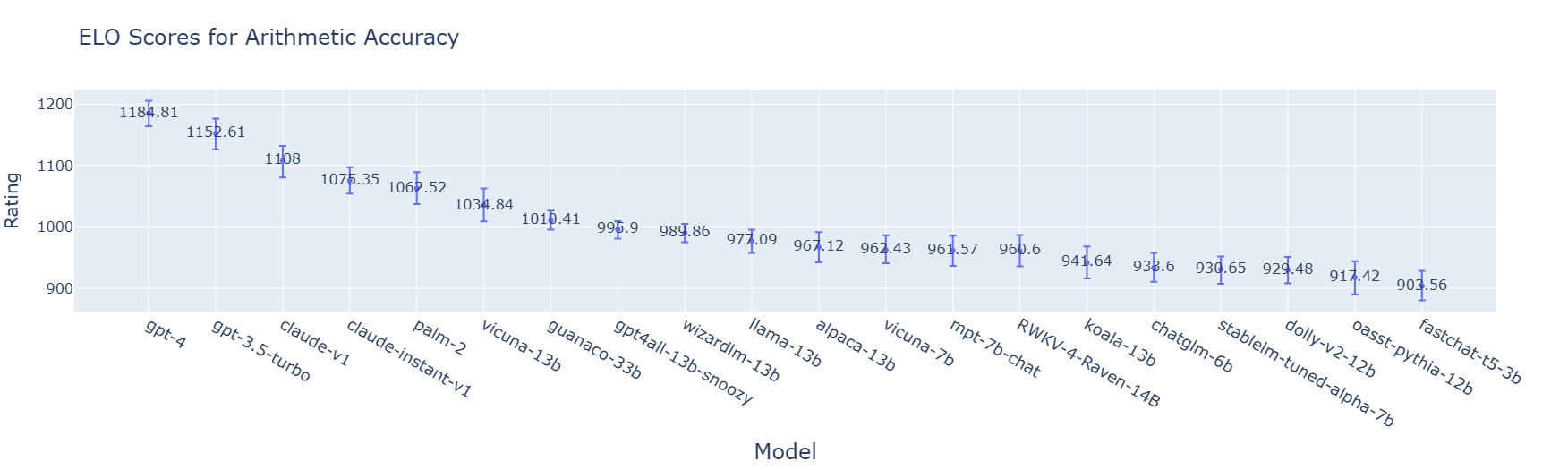}
    \caption{ELO Scores for Arithmetic Accuracy across all models in Chatbot Arena, scored by the mean score of three LLMs.}
    \label{fig:elo-arithmetic-c}
\end{figure}

\begin{figure}[H]
    \centering
    \includegraphics[width=0.99\textwidth]{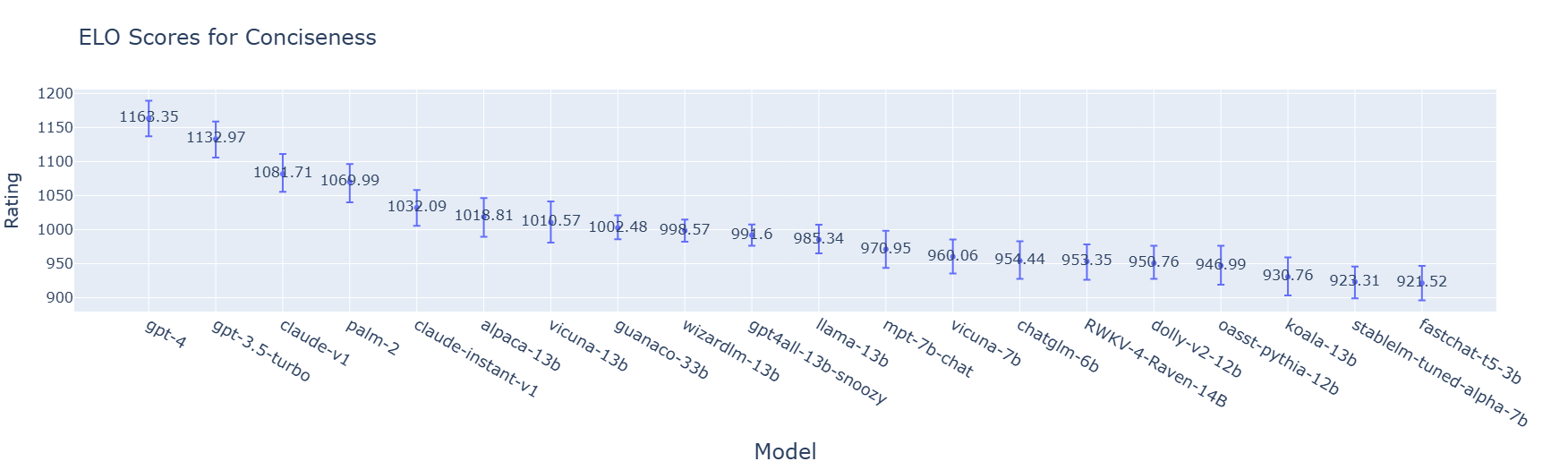}
    \caption{ELO Scores for Conciseness across all models in Chatbot Arena, scored by the mean score of three LLMs.}
    \label{fig:elo-conciseness-c}
\end{figure}
    \begin{figure}[H]
        \centering
        \includegraphics[width=0.8\linewidth]{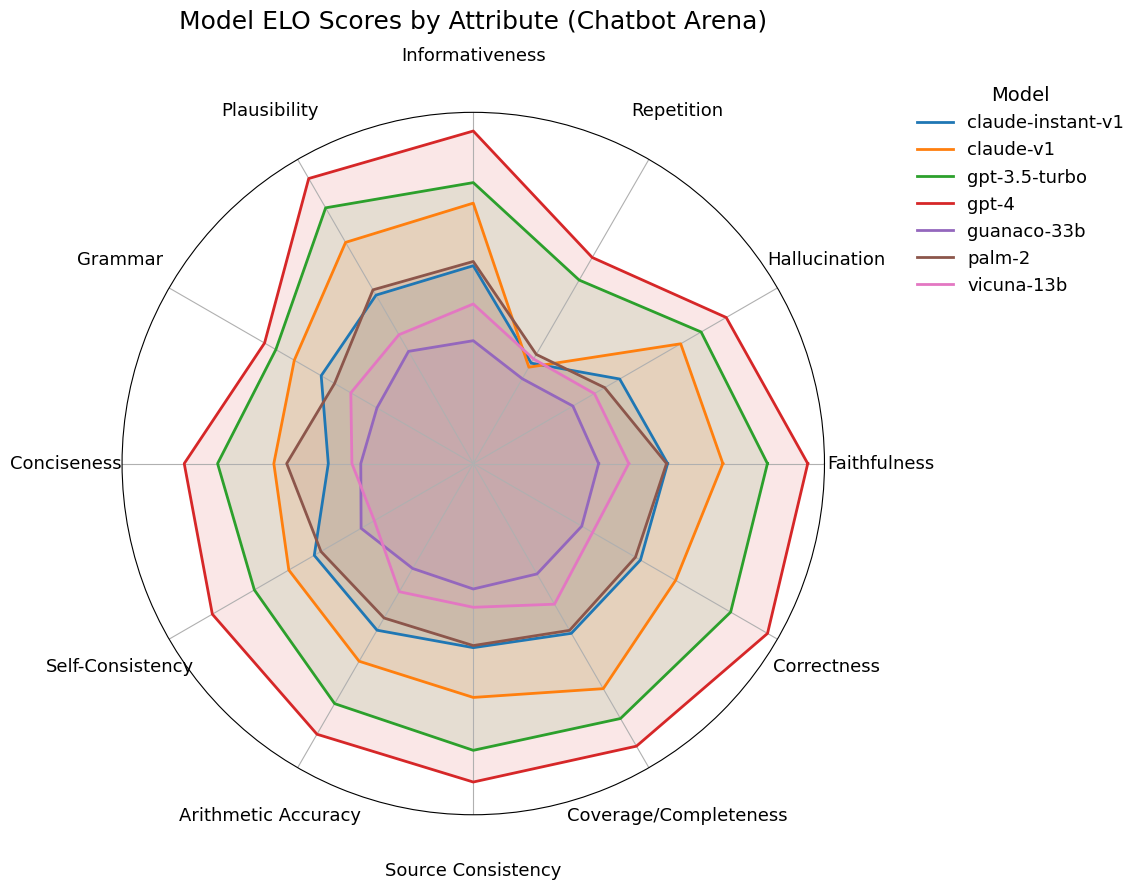}
        \caption{Radar chart of model ELO scores by attribute in Chatbot Arena, computed as the mean of the three LLM judges.}
        \label{fig:elo-radar-llm-chat-c}
    \end{figure}

    \begin{figure}[H]
        \centering
        \includegraphics[width=0.99\linewidth]{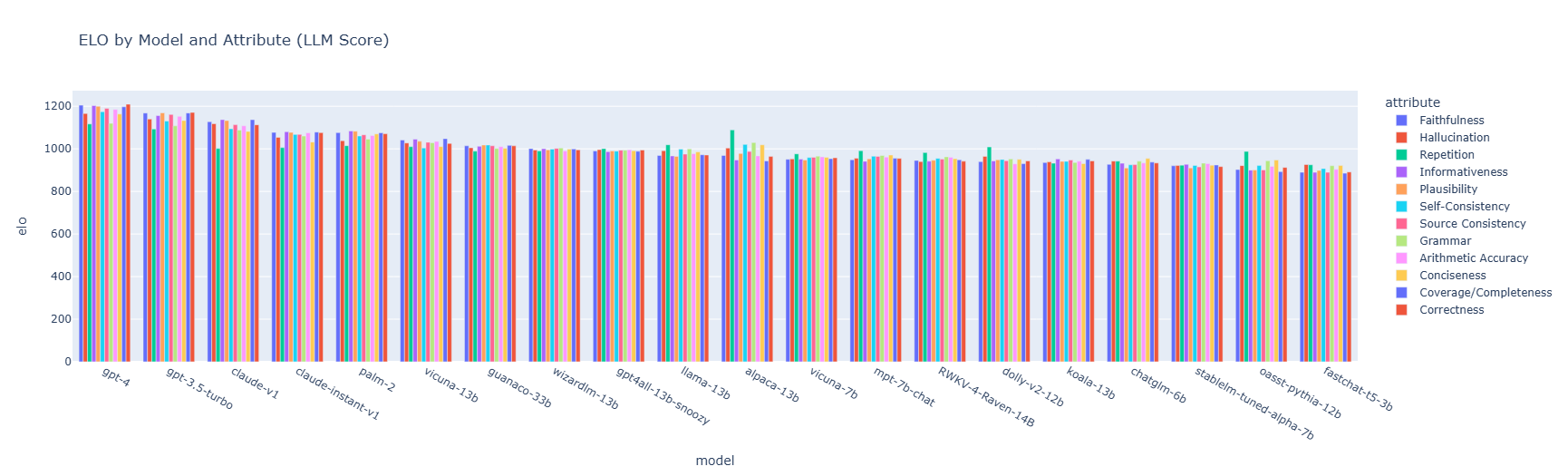}
        \caption{Bar plot of model ELO scores by model and attribute in Chatbot Arena, averaged across all LLM judges.}
        \label{fig:elo-bar-llm-chat-c}
    \end{figure}
    
    \begin{figure}[H]
        \centering
        \includegraphics[width=0.99\linewidth]{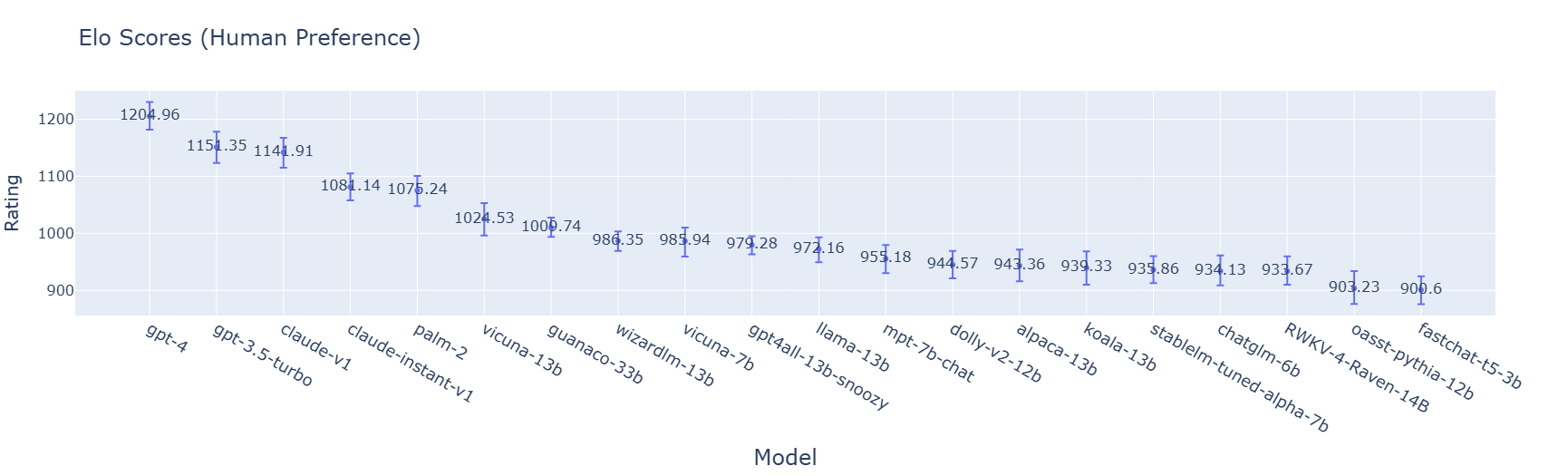}
        \caption{ELO scores for all models based on human preference labels in Chatbot Arena. Here, the score is computed directly from the human-chosen vs.\ rejected outcomes. Error bars indicate the confidence interval.}
        \label{fig:elo-humanpref-llm-chat-c}
    \end{figure}

    \newpage
    \item \textbf{Mt Bench}
    \begin{figure}[H]
    \centering
    \includegraphics[width=0.99\textwidth]{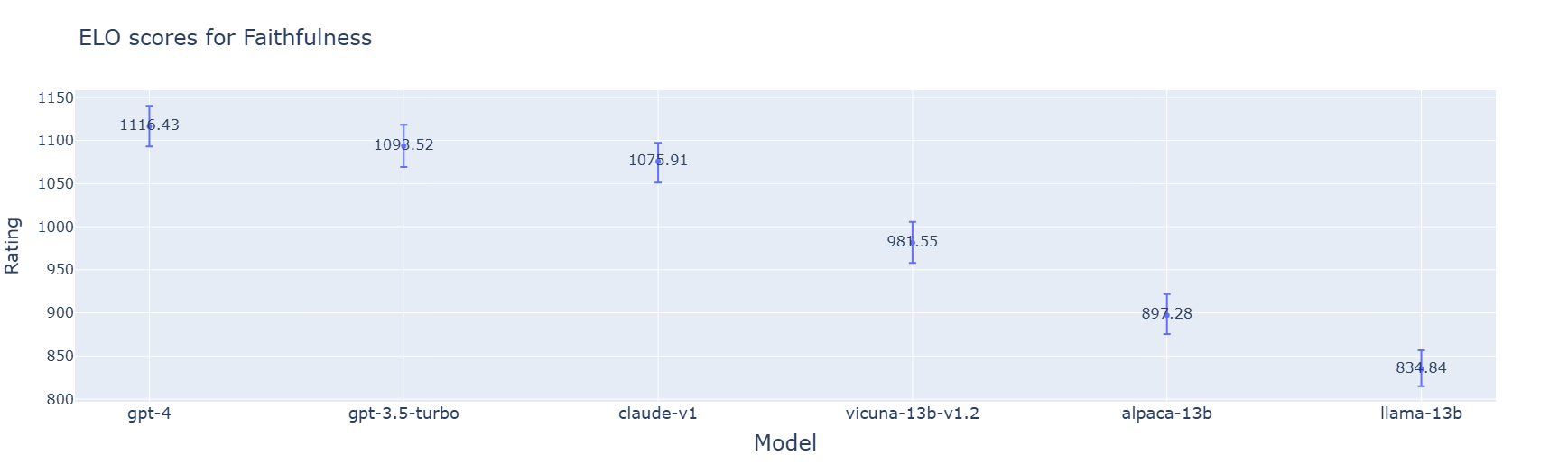}
    \caption{ELO Scores for Faithfulness across all models in Mt Bench, scored by the mean score of three LLMs.}
    \label{fig:elo-faithfulness}
\end{figure}

\begin{figure}[H]
    \centering
    \includegraphics[width=0.99\textwidth]{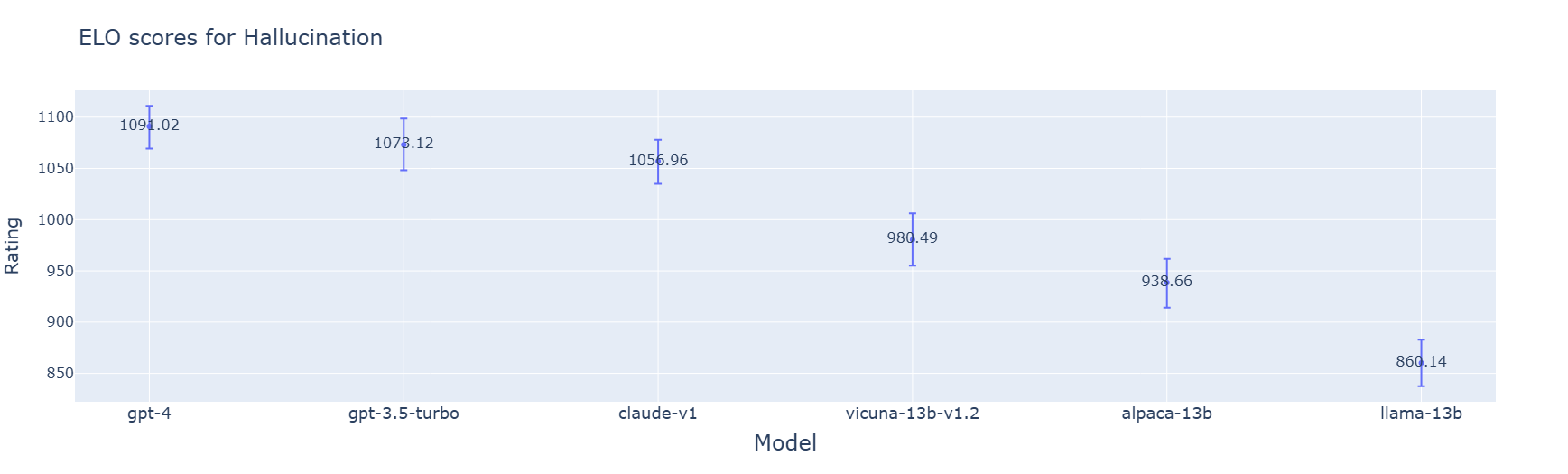}
    \caption{ELO Scores for Hallucination across all models in Mt Bench, scored by the mean score of three LLMs.}
    \label{fig:elo-hallucination}
\end{figure}

\begin{figure}[H]
    \centering
    \includegraphics[width=0.99\textwidth]{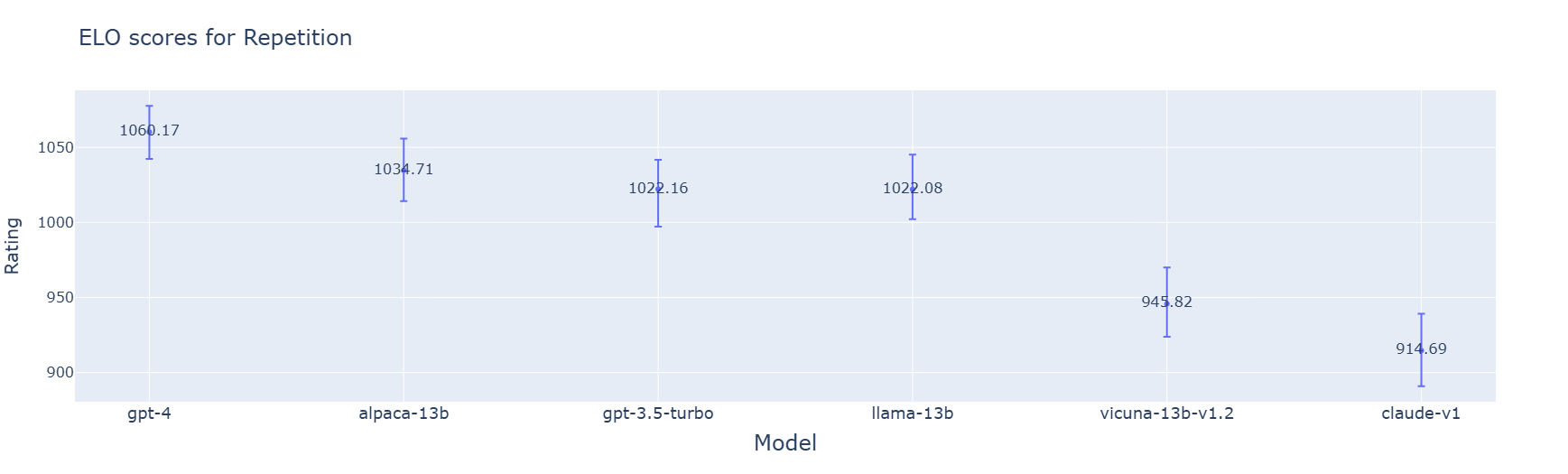}
    \caption{ELO Scores for Repetition across all models in Mt Bench, scored by the mean score of three LLMs.}
    \label{fig:elo-repetition}
\end{figure}

\begin{figure}[H]
    \centering
    \includegraphics[width=0.99\textwidth]{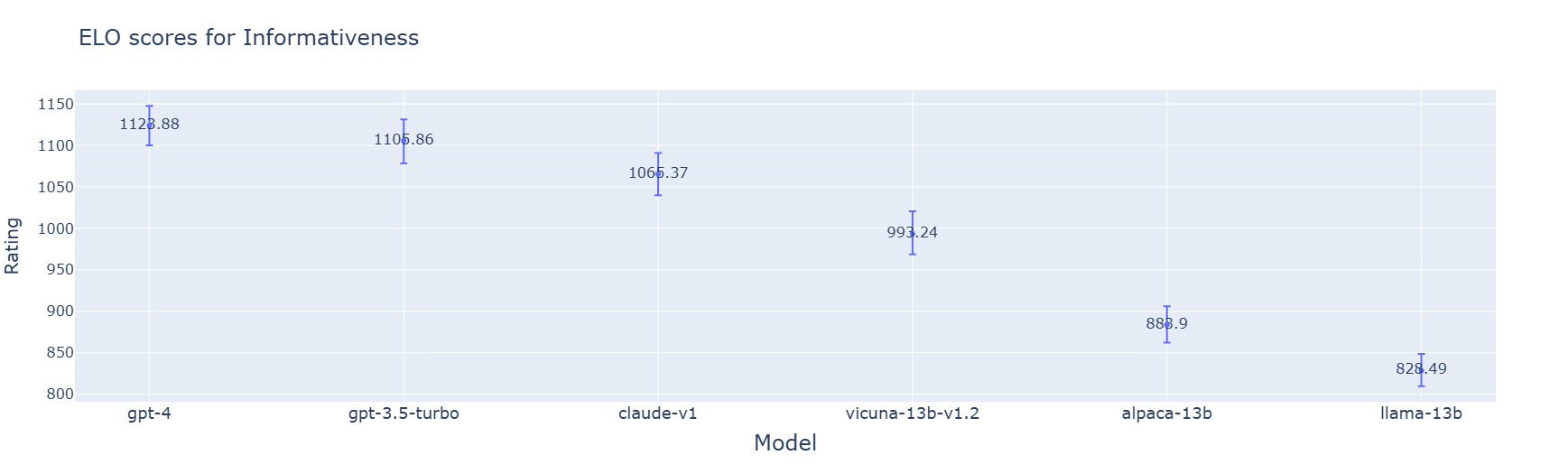}
    \caption{ELO Scores for Informativeness across all models in Mt Bench, scored by the mean score of three LLMs.}
    \label{fig:elo-informativeness}
\end{figure}

\begin{figure}[H]
    \centering
    \includegraphics[width=0.99\textwidth]{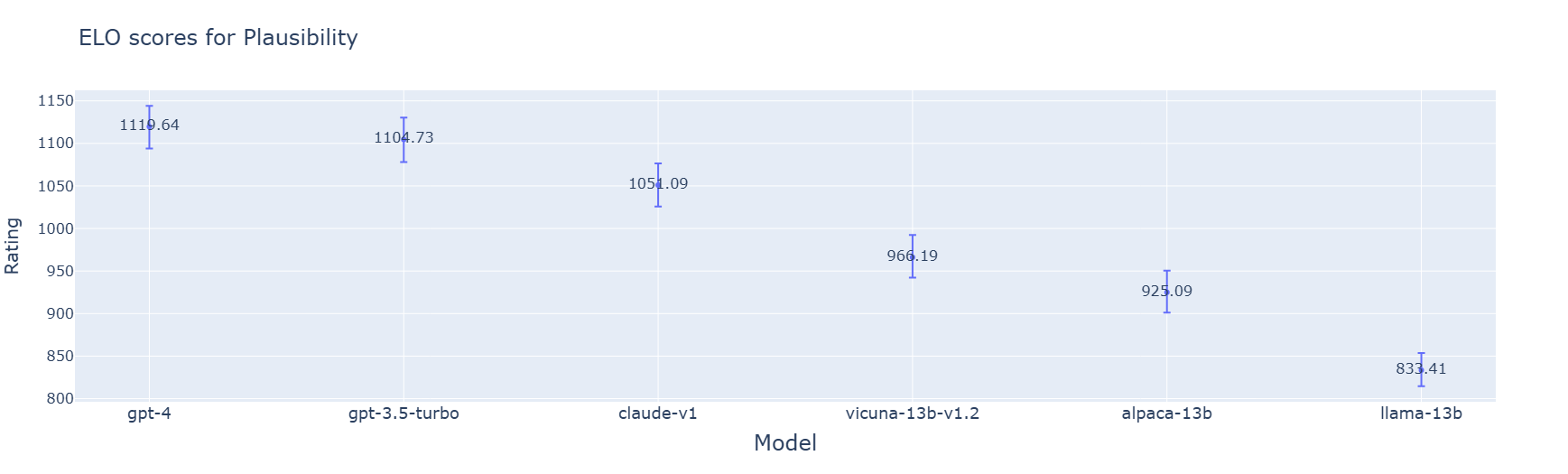}
    \caption{ELO Scores for Plausibility across all models in Mt Bench, scored by the mean score of three LLMs.}
    \label{fig:elo-plausibility}
\end{figure}

\begin{figure}[H]
    \centering
    \includegraphics[width=0.99\textwidth]{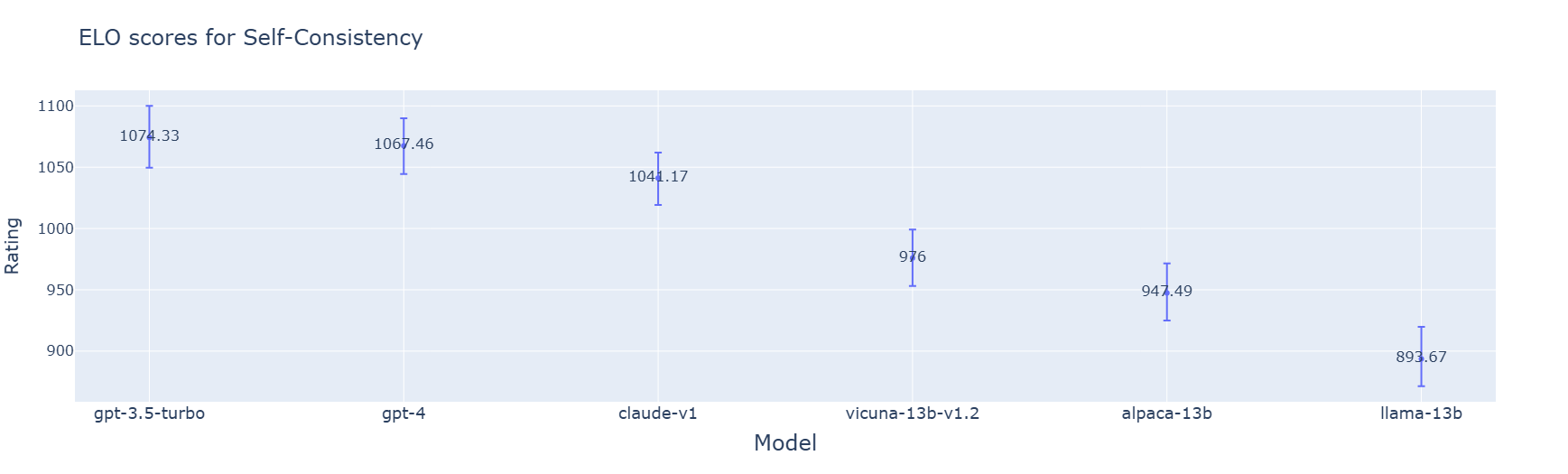}
    \caption{ELO Scores for Self-Consistency across all models in Mt Bench, scored by the mean score of three LLMs.}
    \label{fig:elo-selfconsistency}
\end{figure}

\begin{figure}[H]
    \centering
    \includegraphics[width=0.99\textwidth]{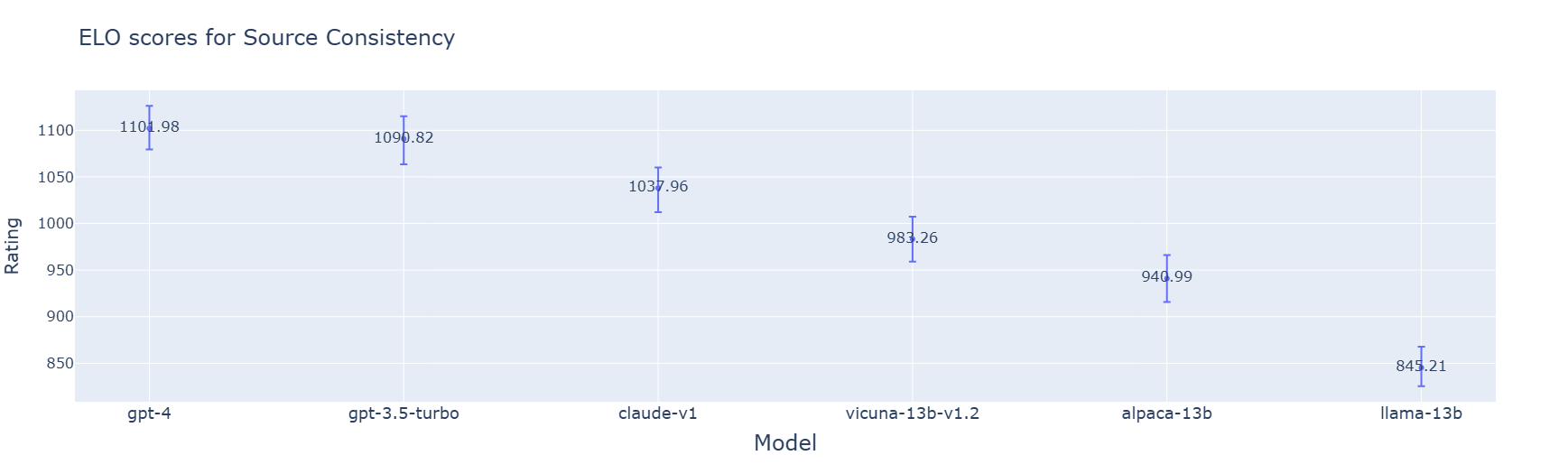}
    \caption{ELO Scores for Source Consistency across all models in Mt Bench, scored by the mean score of three LLMs.}
    \label{fig:elo-sourceconsistency}
\end{figure}

\begin{figure}[H]
    \centering
    \includegraphics[width=0.99\textwidth]{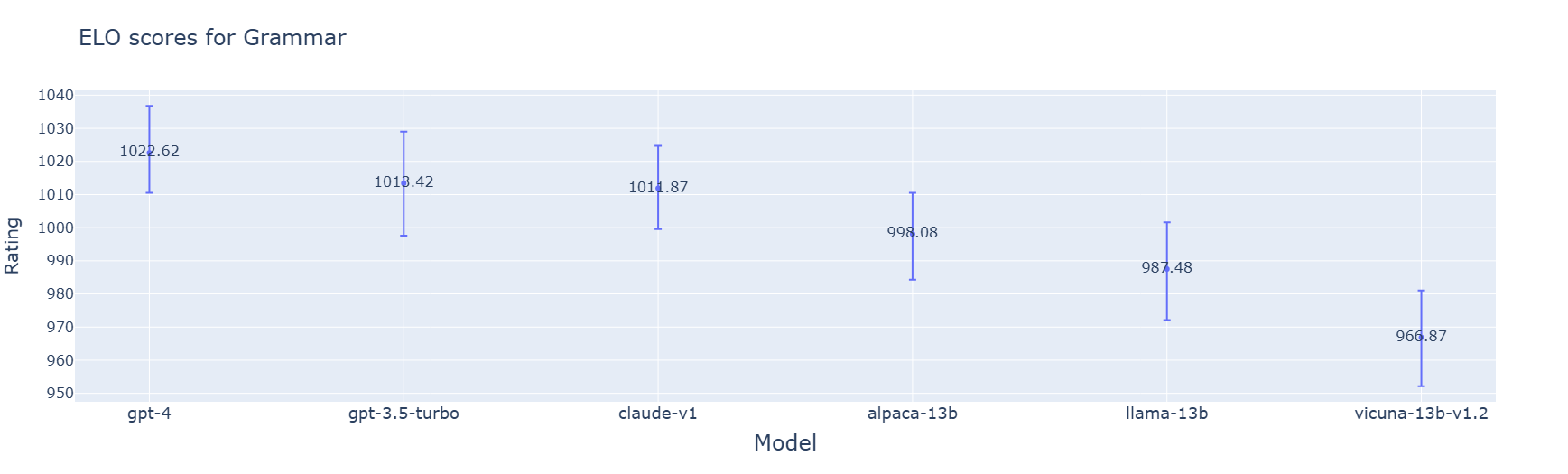}
    \caption{ELO Scores for Grammar across all models in Mt Bench, scored by the mean score of three LLMs.}
    \label{fig:elo-grammar}
\end{figure}

\begin{figure}[H]
    \centering
    \includegraphics[width=0.99\textwidth]{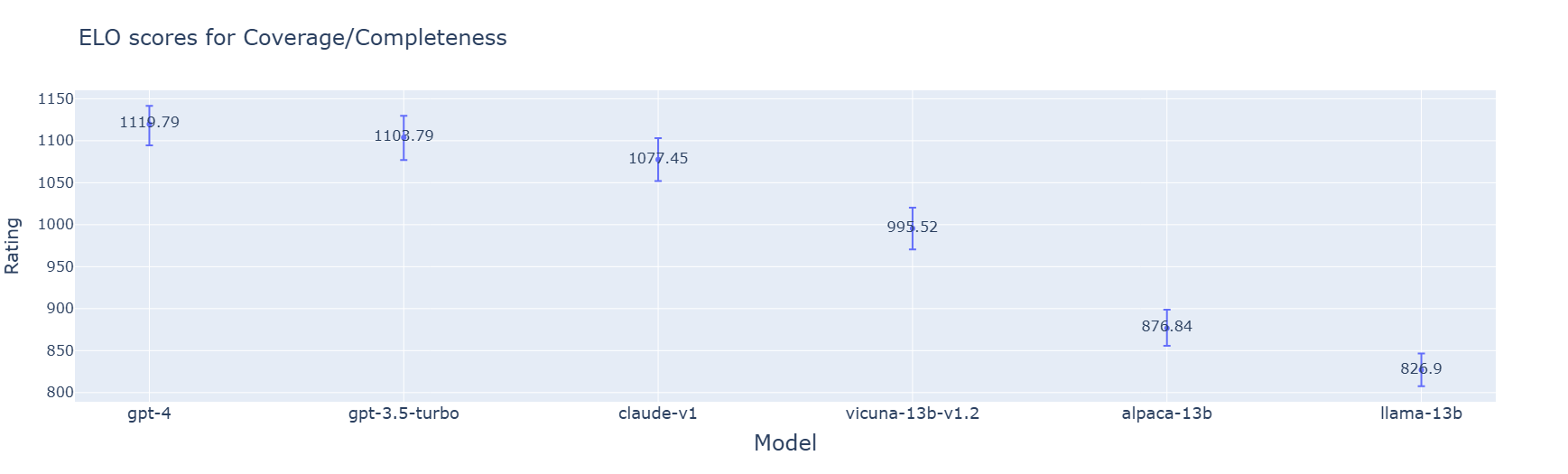}
    \caption{ELO Scores for Completeness across all models in Mt Bench, scored by the mean score of three LLMs.}
    \label{fig:elo-correctness}
\end{figure}

\begin{figure}[H]
    \centering
    \includegraphics[width=0.99\textwidth]{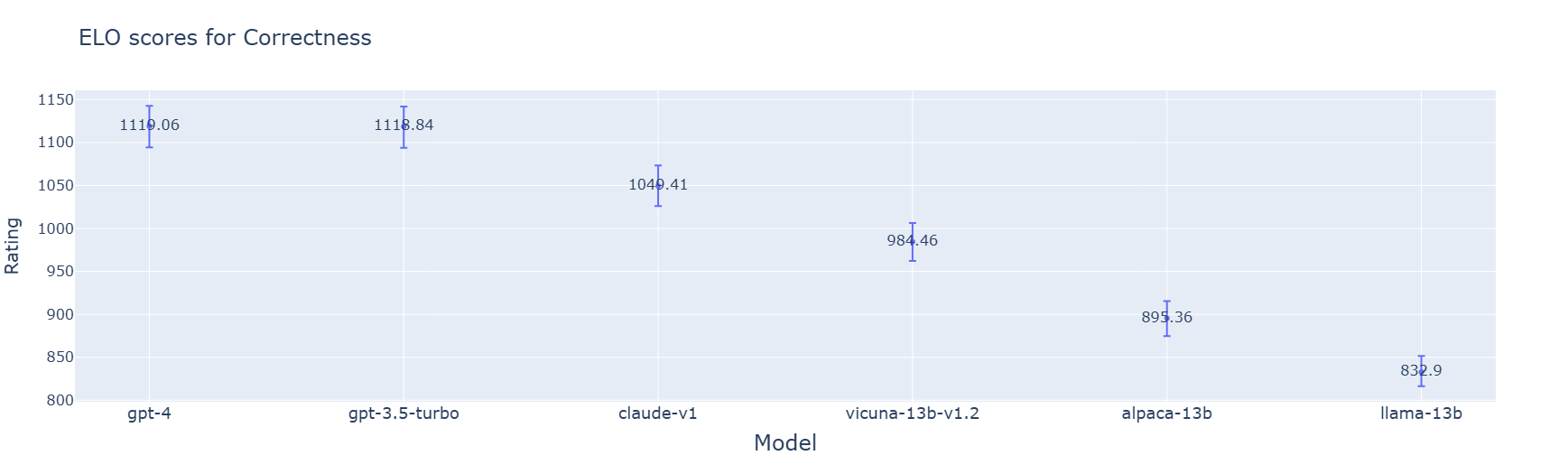}
    \caption{ELO Scores for Correctness across all models in Mt Bench, scored by the mean score of three LLMs.}
    \label{fig:elo-coverage}
\end{figure}

\begin{figure}[H]
    \centering
    \includegraphics[width=0.99\textwidth]{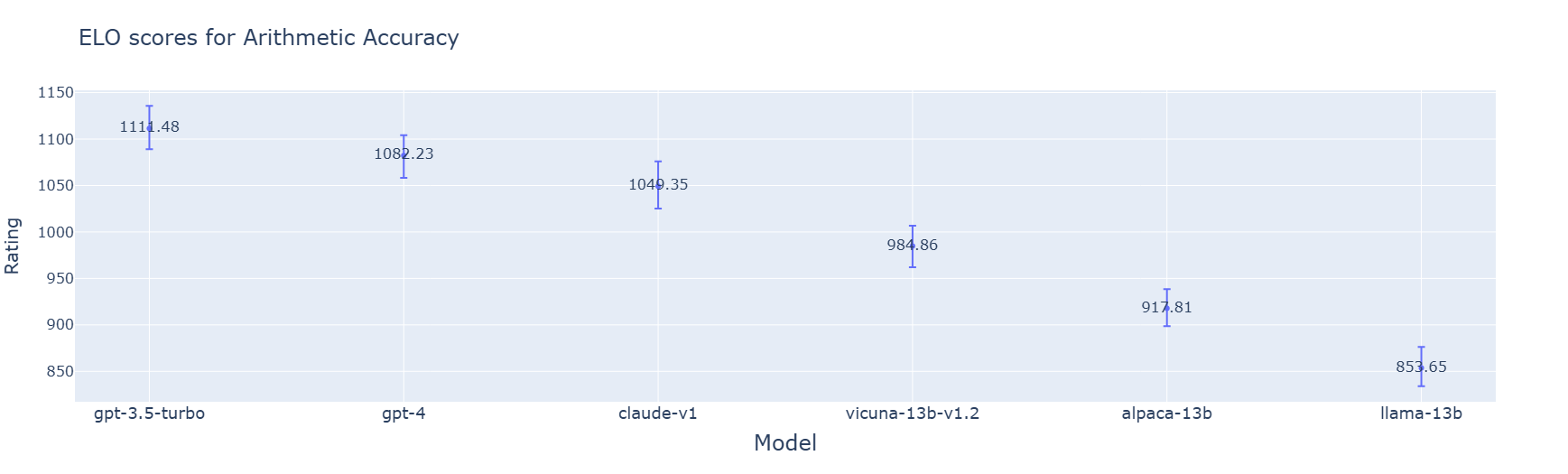}
    \caption{ELO Scores for Arithmetic Accuracy across all models in Mt Bench, scored by the mean score of three LLMs.}
    \label{fig:elo-arithmetic}
\end{figure}

\begin{figure}[H]
    \centering
    \includegraphics[width=0.99\textwidth]{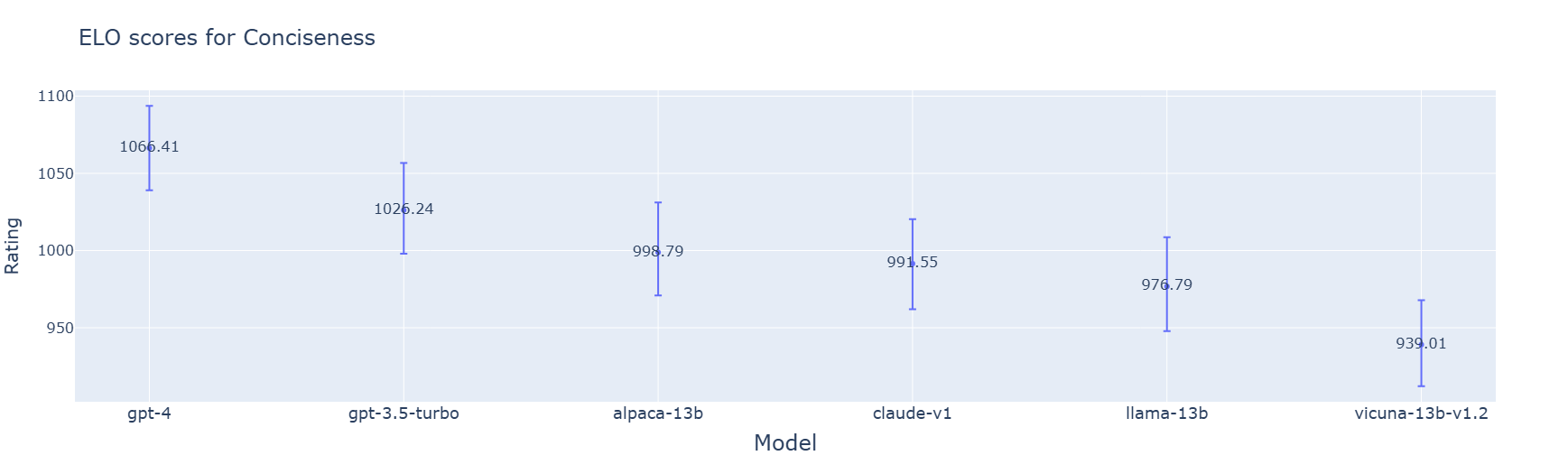}
    \caption{ELO Scores for Conciseness across all models in Mt Bench, scored by the mean score of three LLMs.}
    \label{fig:elo-conciseness}
\end{figure}

    \begin{figure}[H]
    \centering
    \includegraphics[width=0.8\textwidth]{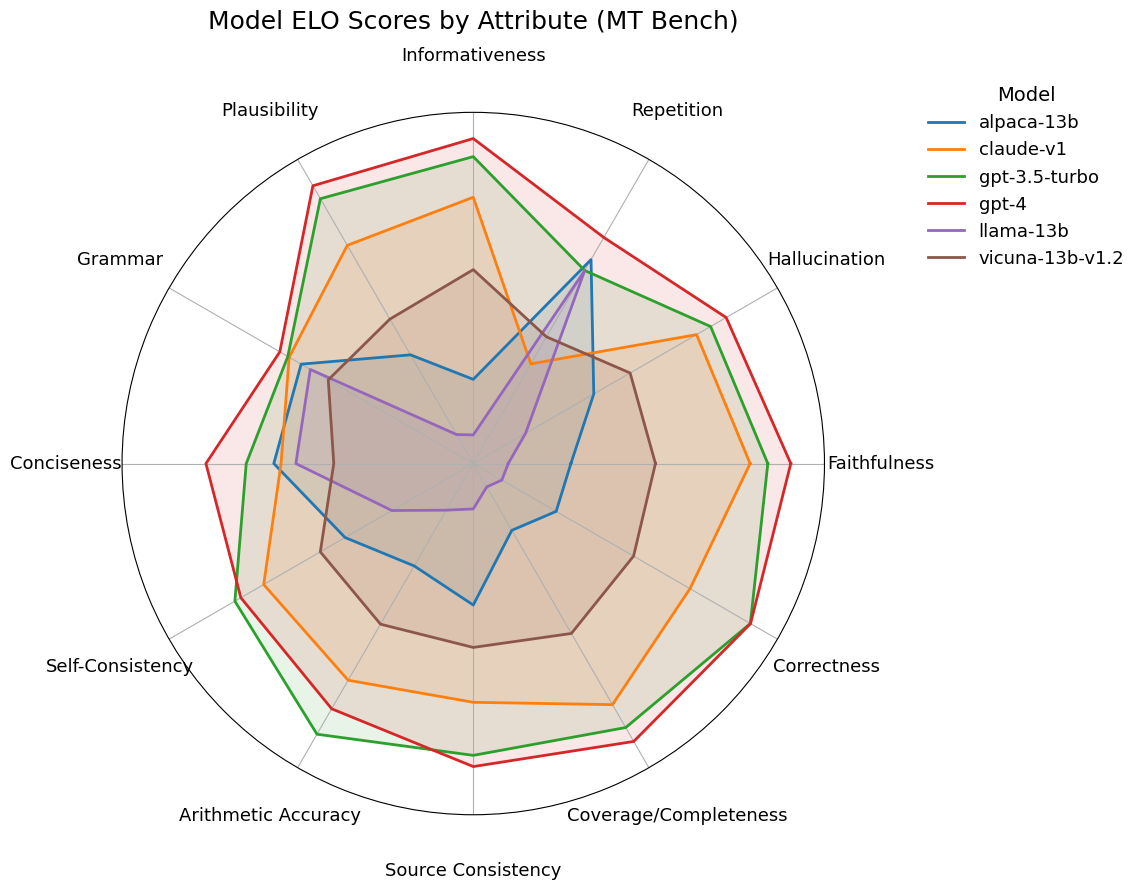}
    \caption{
        Radar chart comparing model ELO scores by attribute on MT Bench (LLM scores). Each axis represents one evaluation attribute, and each polygon represents a different model.
    }
    \label{fig:mtbench-elo-radar}
\end{figure}

\begin{figure}[H]
    \centering
    \includegraphics[width=0.99\textwidth]{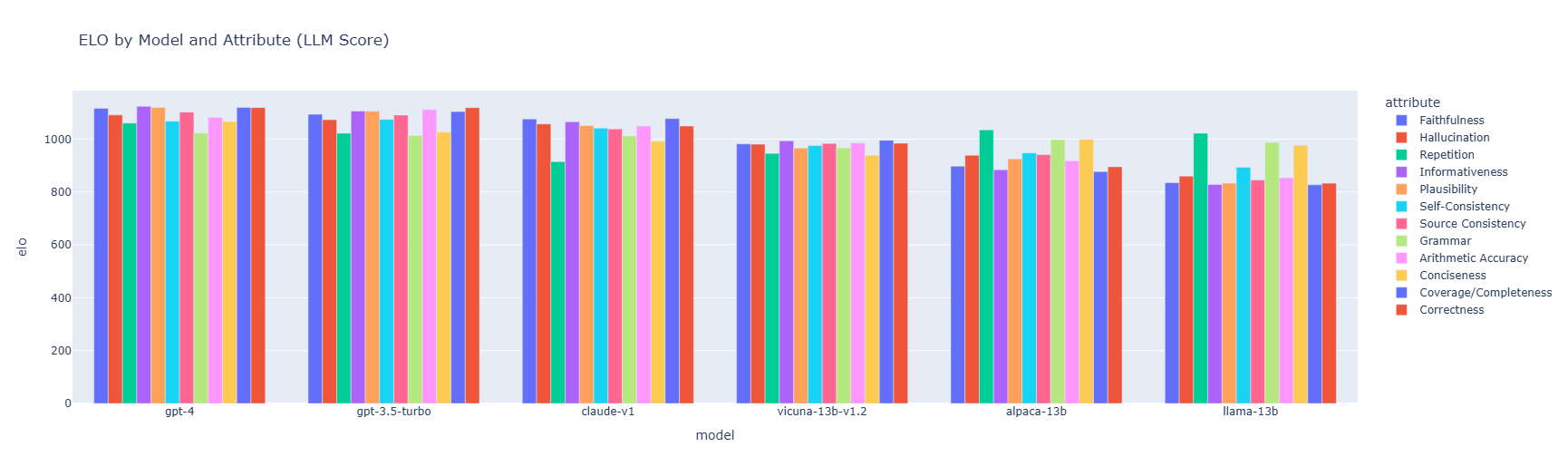}
    \caption{
        Bar chart showing ELO scores for each model and attribute on MT Bench (LLM scores). Each color indicates a different evaluation attribute.
    }
    \label{fig:mtbench-elo-bar}
\end{figure}

\begin{figure}[H]
    \centering
    \includegraphics[width=0.99\textwidth]{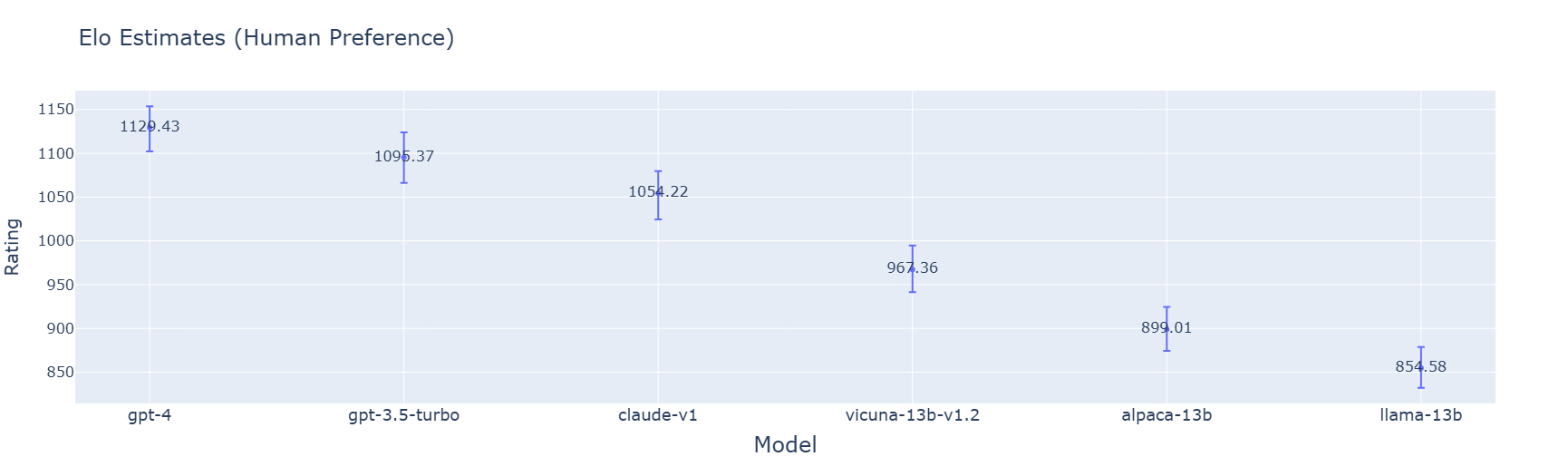}
    \caption{
        Bar plot of ELO estimates based on human preferences for each model on MT Bench. Error bars indicate uncertainty in ELO estimation.
    }
    \label{fig:mtbench-elo-human}
\end{figure}
\end{enumerate}

\subsubsection{Human Annotator}

\begin{enumerate}
    \item \textbf{Chatbot Arena}
    \begin{figure}[H]
    \centering
    \includegraphics[width=0.8\textwidth]{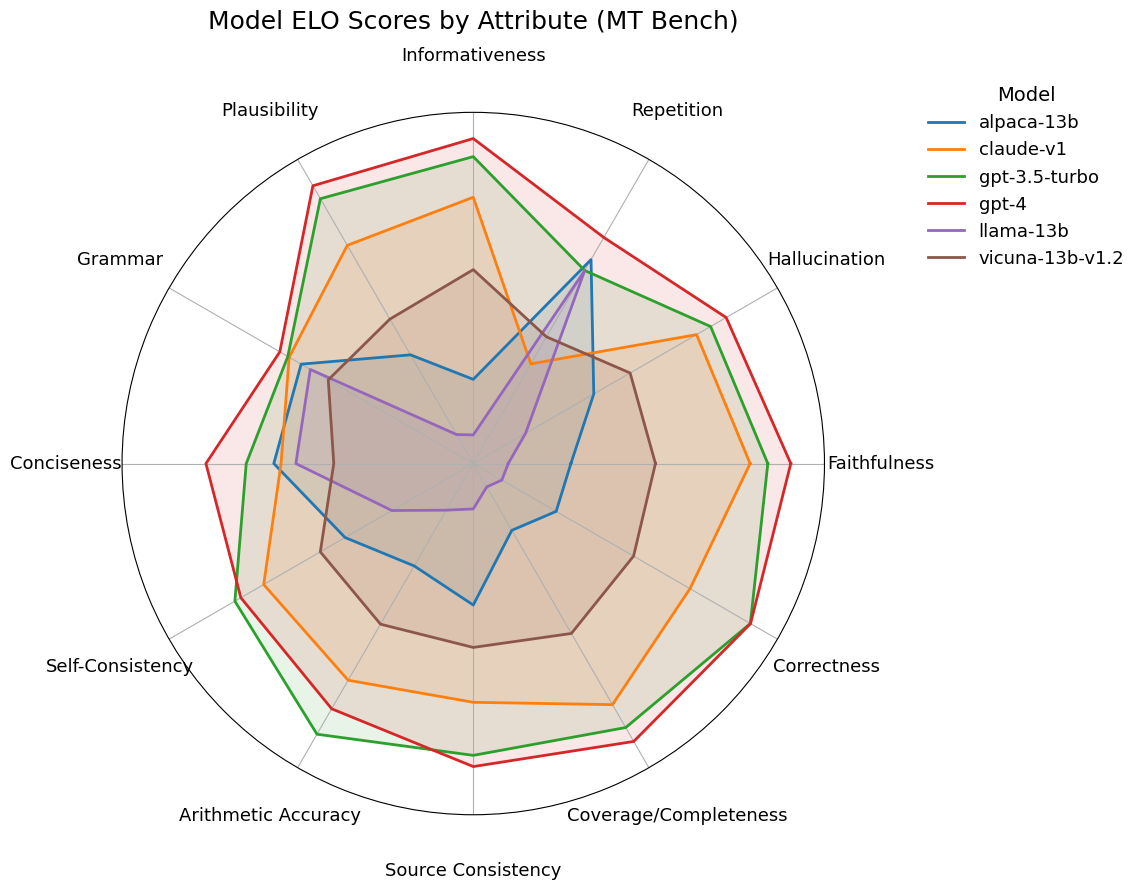}
    \caption{
        Radar chart comparing model ELO scores by attribute on MT Bench (LLM Judges). Each axis corresponds to a specific evaluation attribute, and each line represents a different model.
    }
    \label{fig:mtbench-elo-radar-2}
\end{figure}

\begin{figure}[H]
    \centering
    \includegraphics[width=0.8\textwidth]{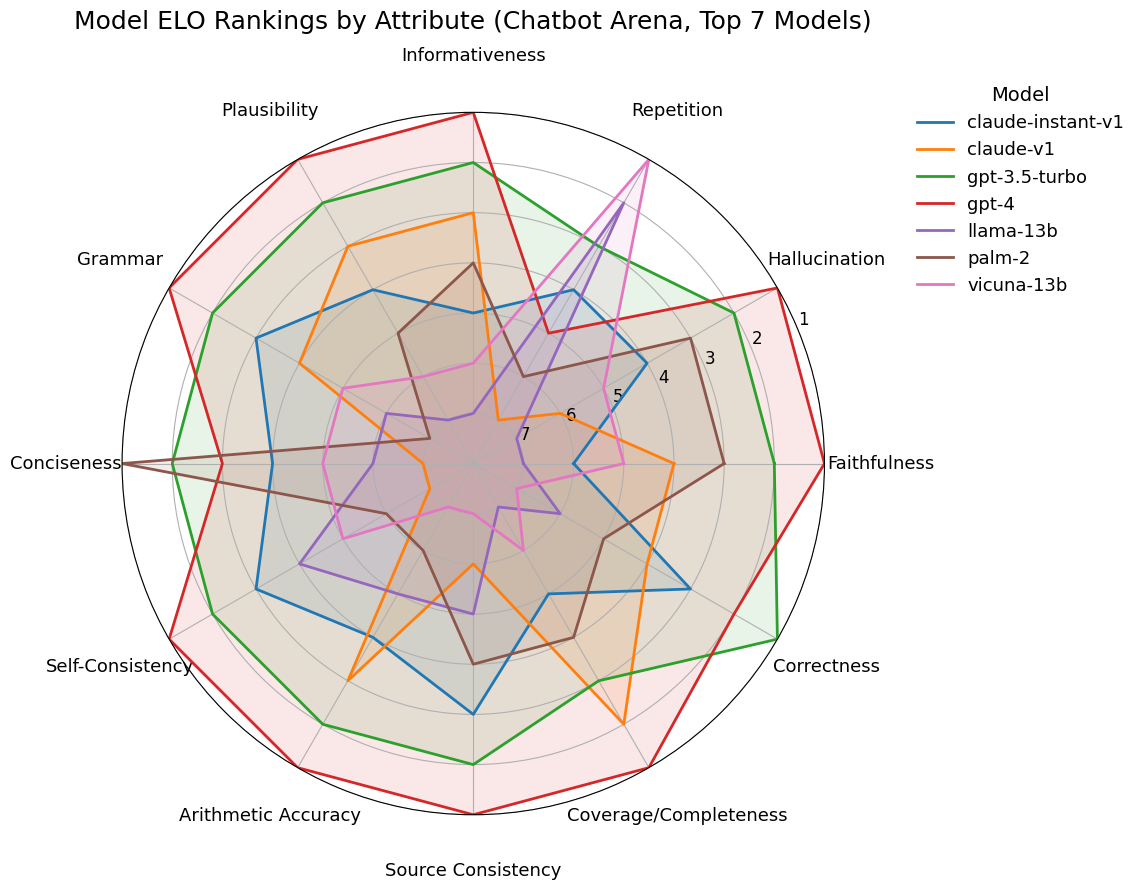}
    \caption{
        Radar chart comparing model ELO rankings by attribute for the top 7 models in Chatbot Arena (LLM Judges). Each axis is an evaluation attribute, and each polygon represents a model.
    }
    \label{fig:chatbotarena-elo-radar}
\end{figure}

\newpage
    \item \textbf{Mt Bench}
    \begin{figure}[H]
    \centering
    \includegraphics[width=0.8\textwidth]{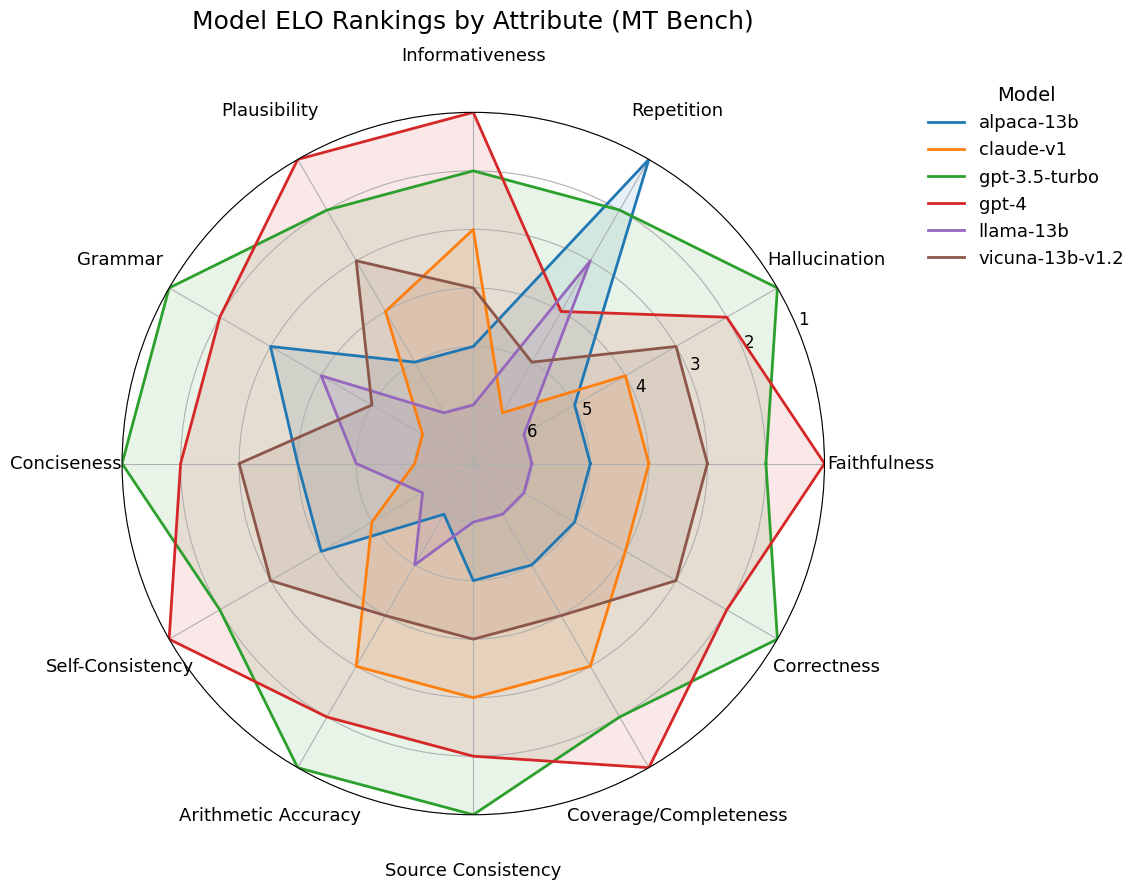}
    \caption{
        Radar chart showing model ELO rankings by attribute on MT Bench. Each axis represents an evaluation attribute, and each polygon represents a model. (Lower is better)
    }
    \label{fig:mtbench-elo-ranking-radar}
\end{figure}

\begin{figure}[H]
    \centering
    \includegraphics[width=0.8\textwidth]{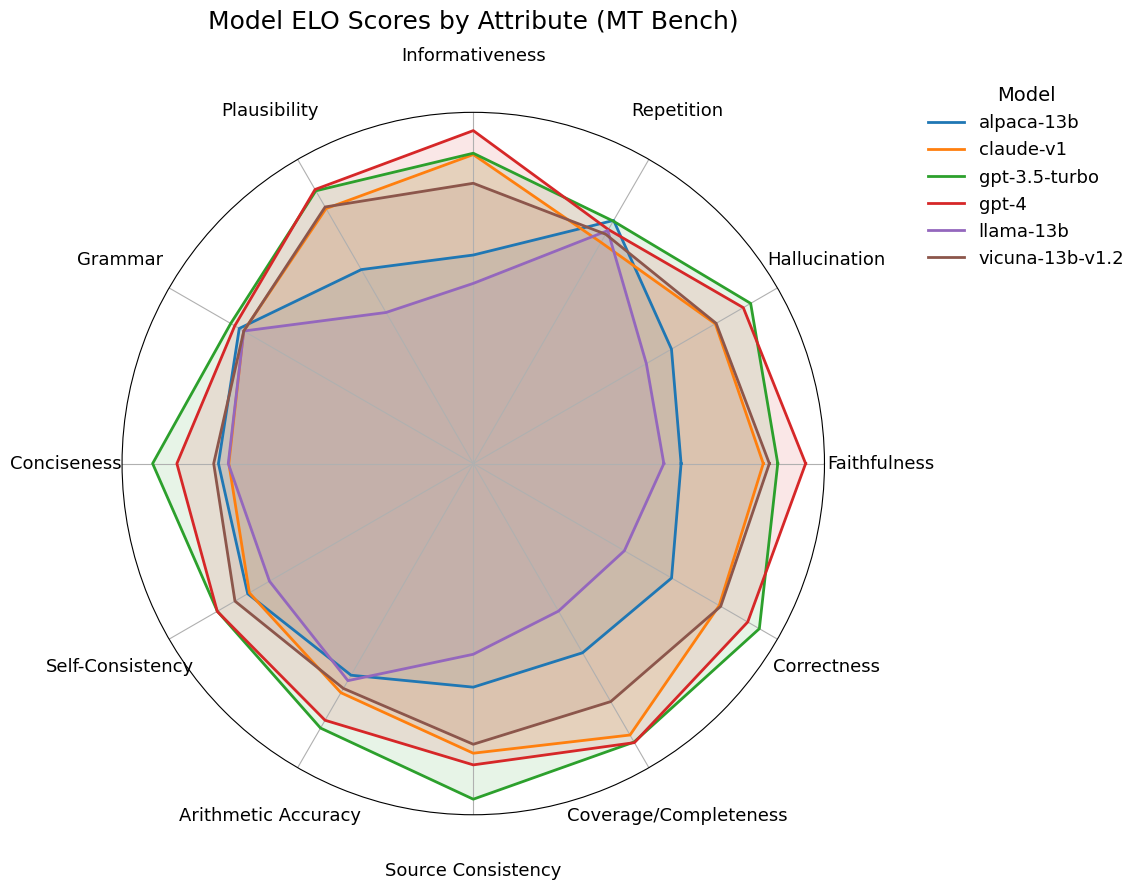}
    \caption{
        Radar chart showing model ELO scores by attribute on MT Bench. Each axis represents an evaluation attribute, and each polygon represents a model. (Higher is better)
    }
    \label{fig:mtbench-elo-score-radar}
\end{figure}
\end{enumerate}
\newpage
\subsection{Examples} \label{gpt4o-mistakes-example}
An example illustrating a mistake made by the GPT-4o model on the Correctness attribute: In this case, both the chosen and rejected rationales provide incorrect final answers. However, GPT-4o assigns a correctness score of 1.0 to the chosen rationale, failing to recognize the error in the solution.

\begin{figure}[H]
    \centering
    \includegraphics[width=0.95\linewidth]{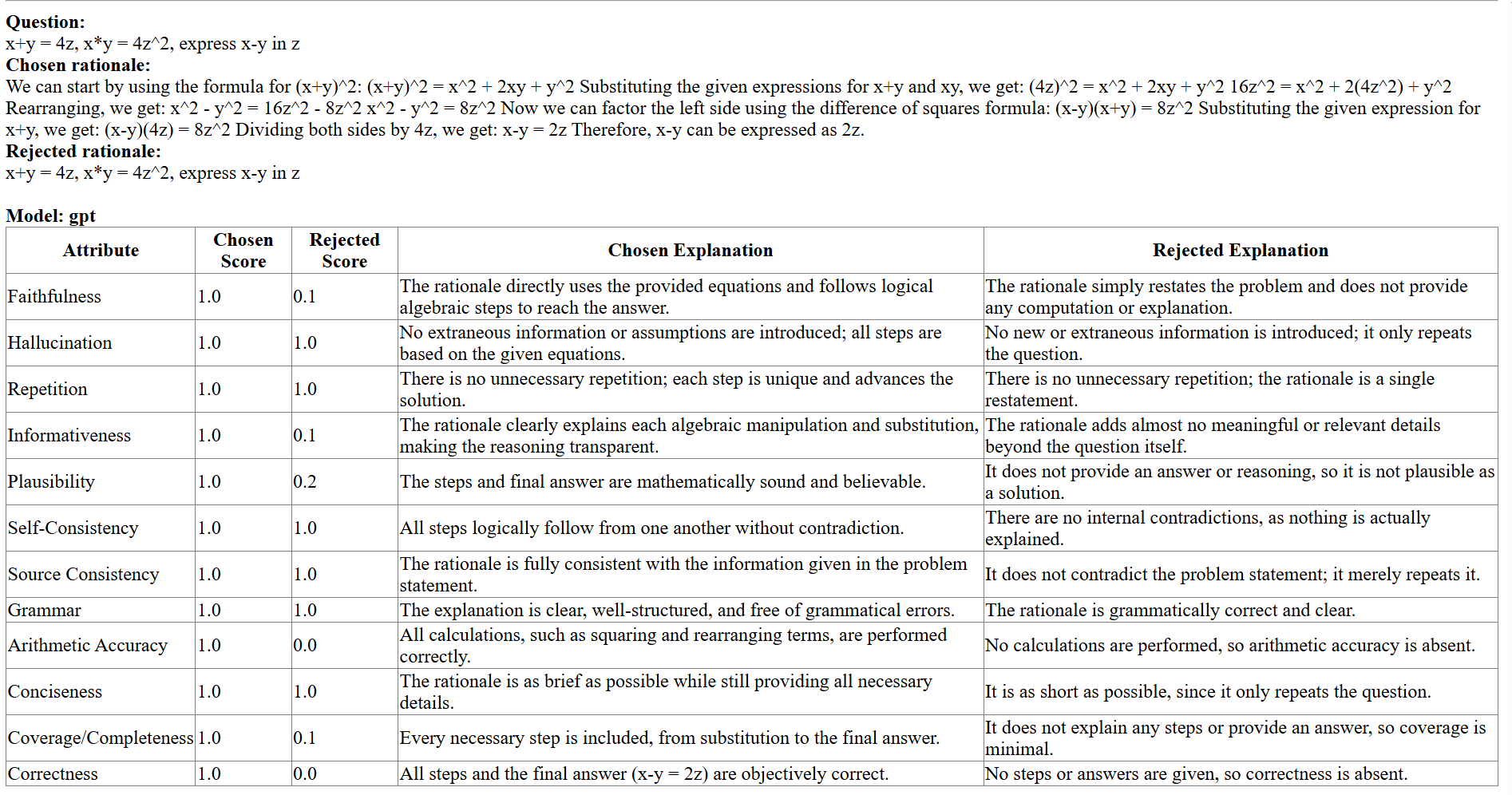}
    \caption{An example of a scoring error made by GPT-4o on the \texttt{Correctness} attribute. The 'Chosen' rationale contains an algebraic error but is incorrectly assigned a perfect score of 1.0, while the 'Rejected' rationale, which only restates the prompt, is correctly scored as 0.0.}
    \label{fig:enter-label}
\end{figure}

\end{document}